\documentclass[twoside,letterpaper]{article}
\usepackage[T1]{fontenc}
\usepackage{lmodern}
\usepackage[preprint]{aistats2027}
\usepackage{amsmath,amssymb,amsthm,mathtools,booktabs,graphicx,microtype,array,longtable}
\usepackage[round,authoryear]{natbib}

\usepackage{algorithm,algpseudocode}
\usepackage{tikz}\usetikzlibrary{arrows.meta,decorations.pathreplacing}
\usepackage[hidelinks]{hyperref}
\hypersetup{pdftitle={Directional Bias in Quantile SGD and Score-Inversion Inference under Dependence},
  pdfauthor={Xiaoli Li, Qianqian Lei, Wei Biao Wu}}
\newtheorem{theorem}{Theorem}
\newtheorem{proposition}[theorem]{Proposition}
\newtheorem{corollary}[theorem]{Corollary}
\newtheorem{lemma}[theorem]{Lemma}
\theoremstyle{remark}\newtheorem{remark}[theorem]{Remark}
\theoremstyle{definition}\newtheorem{assumption}{Assumption}

\newcommand{\E}{\mathbb E}\newcommand{\Pp}{\mathbb P}
\newcommand{\R}{\mathbb R}\newcommand{\ind}{\mathbf 1}
\newcommand{\Var}{\operatorname{Var}}\newcommand{\Cov}{\operatorname{Cov}}
\newcommand{\norm}[1]{\left\lVert#1\right\rVert}

\newcommand{\op}{o_{\Pp}}\newcommand{\Op}{O_{\Pp}}

\newcommand{\dd}{\mathrm d}
\newcommand{\AMem}{\ref{as:memory}}\newcommand{\AMarg}{\ref{as:marginal}}
\newcommand{\ACond}{\ref{as:conditional}}\newcommand{\AScore}{\ref{as:score}}
\graphicspath{{figures/}}
\begin{document}
\runningtitle{Directional Bias in Quantile SGD}
\runningauthor{Xiaoli Li, Qianqian Lei, and Wei Biao Wu}
\setlength{\footskip}{20pt}
\fancyfoot[C]{\thepage}
\raggedbottom
\twocolumn[

\aistatstitle{Directional Bias in Quantile SGD\\and Score-Inversion Inference under Dependence}

\aistatsauthor{Xiaoli Li \And Qianqian Lei \And Wei Biao Wu}

\aistatsaddress{
University of Chicago\\
\href{mailto:xiaolil@uchicago.edu}{\texttt{xiaolil@uchicago.edu}}
\And
University of Chicago\\
\href{mailto:qianqian@uchicago.edu}{\texttt{qianqian@uchicago.edu}}
\And
University of Chicago\\
\href{mailto:wbwu@uchicago.edu}{\texttt{wbwu@uchicago.edu}}
}
]
\begin{abstract}
Quantile stochastic gradient descent (SGD) tracks a quantile by comparing each observation with an adaptive threshold, and closely related updates calibrate online conformal sets and drive quantile temporal-difference learning. Their inputs are serially dependent, yet existing guarantees largely assume independence. We show that dependence creates a directional bias that score autocovariances cannot detect: reversing a stationary stream leaves every score autocovariance unchanged but can change the leading bias threefold. For finite memory and decreasing stepsizes, the bias splits into lag-specific interactions between the threshold and the current observation and a curvature term; delaying the threshold removes the interactions one lag at a time while keeping every observation and the leading variance. At stepsizes near $t^{-1/2}$, as used for online inference, the bias is comparable to the standard error and makes an SGD-based interval miss mostly on one side. Testing fixed candidates instead keeps the score centered, so calibration depends only on score autocovariances. Extending local-dependence self-normalized approximations to the implemented acceptance hull, we show that inverting these tests gives a density-free, moment-free interval for the population quantile with $O(n^{-1/2})$ coverage error for a fixed sufficient lag window, a rate that is sharp for this interval, and a guarantee for data-selected windows; an exact scan takes $O(n\log n)$ time. Experiments confirm the lag-by-lag predictions and show calibration competitive with block bootstrap at lower cost.
\end{abstract}

\section{Introduction}
Quantile stochastic gradient descent (SGD) tracks a quantile with constant memory by comparing each new observation with an adaptive threshold \citep{robbins1951,tierney1983}. Closely related updates calibrate online conformal prediction sets \citep{gibbs2021aci,angelopoulos2024decaying} and drive quantile temporal-difference learning \citep{dabney2018qr,rowland2024qtd}. Their inputs, such as conformity scores, returns or rewards, are serially dependent, whereas guarantees for recursive quantiles largely assume independent inputs. We study the decreasing-step recursion for a fixed population quantile of a stationary stream, as in the decaying-stepsize calibration of \citet{angelopoulos2024decaying}.

\begin{figure}[t]
\centering
\resizebox{.98\columnwidth}{!}{%
\begin{tikzpicture}[x=1cm,y=1cm,font=\footnotesize,
 obs/.style={draw,rounded corners=1pt,minimum width=.94cm,minimum height=.4cm,inner sep=1pt,fill=gray!10},
 thr/.style={draw,rounded corners=1pt,minimum height=.4cm,inner sep=2pt},
 cmp/.style={-{Latex[length=1.5mm]},semithick}]
\node[obs] (old) at (0,0) {$X_{t-r}$};
\node[obs] (first) at (1.25,0) {$X_{t-r+1}$};
\node at (2.45,0) {$\cdots$};
\node[obs] (prev) at (3.6,0) {$X_{t-1}$};
\node[obs,fill=gray!30] (now) at (4.95,0) {$X_t$};
\draw[decorate,decoration={brace,amplitude=4pt,raise=2pt}] (first.north west)--(now.north east)
 node[midway,above=6pt]{recent observations can depend on $X_t$};
\draw[gray,dashed] (now.south)--(4.95,-1.75);
\node[anchor=east] at (-.6,-.8) {SGD};
\node[thr] (q) at (2.8,-.8) {$q_{t-1}$};
\draw[gray,dotted,->] (prev)--(q);
\draw[cmp] (q)--node[above]{interaction}(4.95,-.8);
\node[anchor=east] at (-.6,-1.6) {delay $r$};
\node[thr] (qr) at (0,-1.6) {$q_{t-r}$};
\draw[gray,dotted,->] (old)--(qr);
\draw[cmp] (qr)--node[above]{threshold independent of $X_t$}(4.95,-1.6);
\node[anchor=east] at (-.6,-2.35) {inversion};
\node[thr,anchor=west] at (-.5,-2.35) {fixed $q$: $\E[\ind(X_s\le q)-\tau]=0$ at $q=q^\star$};
\end{tikzpicture}}
\caption{Threshold choice under innovation memory $r$. Ordinary SGD uses a threshold moved by recent data; delay $r$ makes it independent of the current observation; inversion tests fixed candidates whose scores are centered at the population quantile.}\label{fig:schematic}
\end{figure}
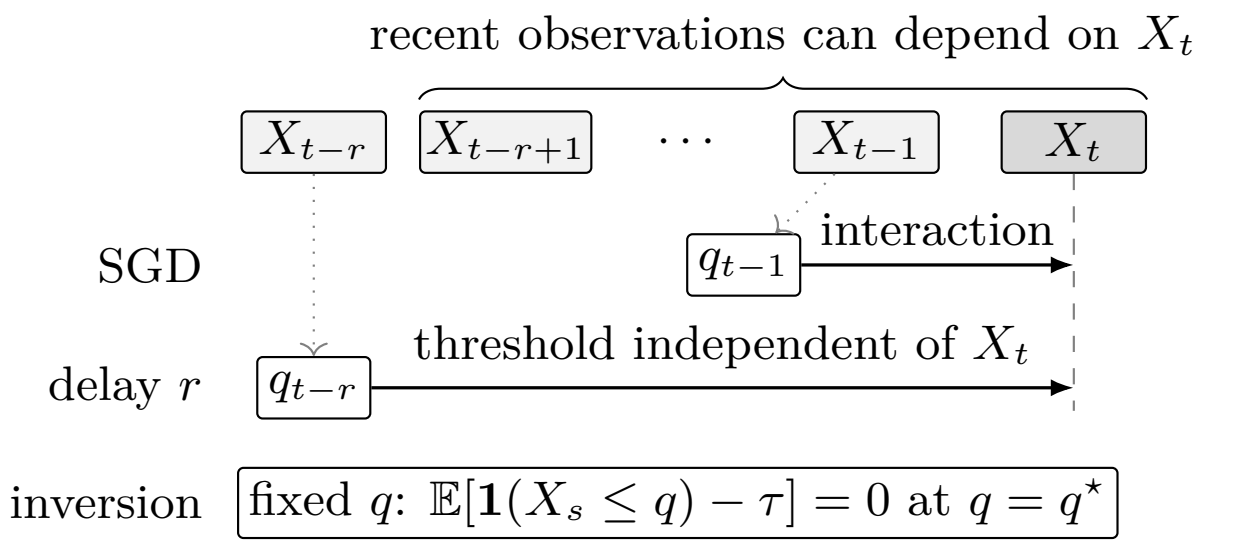

Under dependence, recent observations move the threshold and predict the next input (Figure~\ref{fig:schematic}). Time reversal exposes the consequence: reversing a stationary stream preserves its marginal quantiles and every autocovariance of the scalar quantile score, so every covariance-based quantity is unchanged; nevertheless, the leading bias of quantile SGD changes. In the example of Section~\ref{sec:bias}, its coefficient is $-.113$ forward and $-.038$ backward. Our results separate two orders of behavior. To first order, the calibration of score tests depends only on score autocovariances: these are invariant under reversal, can be estimated without a density, and yield an equal-weight interval that is exactly reversal invariant. To second order, the mean of the adaptive recursion depends on boundary derivatives of bivariate laws, which are directional and are not determined by the score autocovariances.

For averaged SGD with stepsizes proportional to $t^{-\gamma}$, the bias is $O(n^{-\gamma})=o(n^{-1/2})$, but it is systematic. When estimates from $R$ independent streams are averaged, for example across devices, the variance decreases as $1/R$ while the bias does not, so the bias-to-standard-deviation ratio, $4.5\%$ per stream in the example of Section~\ref{sec:bias} at $n=48{,}000$, reaches one at $R\approx500$. With stepsizes close to $t^{-1/2}$, as used for online inference, the bias is of the order of the standard error and makes an SGD-based online interval miss mostly on one side (Section~\ref{sec:experiments}).

\paragraph{Contributions.}
\begin{itemize}\setlength{\itemsep}{1pt}\setlength{\parskip}{0pt}\setlength{\leftmargin}{1em}
\item \emph{Directional bias.} A signed expansion for raw-indicator quantile SGD under finite memory shows that the leading bias depends on boundary derivatives of bivariate laws; time reversal gives covariance-identical streams with different biases (Theorem~\ref{thm:bias}, Corollary~\ref{cor:reversal}).
\item \emph{Lag-resolved intervention.} Each additional threshold delay removes one lag interaction while keeping every observation and the leading second moment; sufficient delays need no conditional-density assumption (Theorem~\ref{thm:delay}).
\item \emph{Inference for the implemented interval.} Fixed-candidate score inversion needs no density or moments. We transfer local-dependence normal approximations to the implemented interval, with its candidate-dependent denominator, $1/n$ floor and closed hull, and obtain root-$n$ coverage for a fixed sufficient window and explicit bounds for omitted covariance and data-selected windows (Proposition~\ref{prop:hullsandwich}, Theorem~\ref{cor:sharpcoverage}, Corollary~\ref{cor:datawindow}).
\item \emph{Computation and evidence.} An exact $O(n\log n)$ scan with $O(n)$ storage (Proposition~\ref{prop:fenwick}). Experiments verify the mechanism lag by lag and the finite-sample bias, explain the one-sided misses of an SGD-based online interval, tie coverage loss to omitted covariance, and show calibration competitive with block bootstrap.
\end{itemize}

\section{Related work}\label{sec:related}
\paragraph{Stochastic approximation under dependence.}
Stochastic approximation (SA) with Markovian noise is analyzed through the ODE method and Poisson equations \citep{benveniste1990,borkar2008}, and SGD on Markovian data can converge more slowly than on independent data \citep{nagaraj2020markov,even2023markov}. For constant stepsizes the bias is of the order of the stepsize, for independent noise \citep{dieuleveut2020} and for Markov noise \citep{huo2024nonlinear,huo2023linear}, and Richardson--Romberg (RR) extrapolation removes its leading term; \citet{lauand2025} give the averaged bias of decreasing-step linear SA with Markovian multiplicative noise. These expansions require smooth updates: \citet{huo2024nonlinear} identify a bias term from the interaction of memory and nonlinearity under three bounded derivatives, whereas the quantile indicator jumps and smoothing makes such bounds diverge. We instead expand the conditional distribution function across the moving threshold. Reverse experience replay shows that processing order matters for linear system identification \citep{kowshik2021rer}; for quantile SGD we identify the boundary functional responsible. Delays usually arise from asynchrony \citep{agarwal2011delay}; here a delay is a deliberate intervention.

\paragraph{Recursive quantiles and their uses.}
Recursive quantile estimators match the sample quantile asymptotically with constant memory \citep{tierney1983}; recent work gives non-asymptotic confidence bounds, constant-step limits and smoothed simultaneous inference for independent inputs \citep{chen2023quantile,costa2021,wei2026constant,chen2026smooth}. Adaptive conformal inference updates the calibration level by quantile-SGD steps with constant or adaptive stepsizes, targeting long-run coverage \citep{gibbs2021aci,gibbs2024jmlr,zaffran2022}; decaying stepsizes also estimate the population quantile \citep{angelopoulos2024decaying}. Quantile temporal-difference learning applies quantile updates to bootstrapped targets along Markovian trajectories \citep{dabney2018qr,rowland2024qtd}. Streaming sketches answer approximate rank queries in sublinear memory \citep{greenwald2001} but do not calibrate inference for a population quantile under dependence.

\paragraph{Online inference for SGD.}
Polyak--Ruppert and weighted averages are asymptotically normal \citep{polyak1992,wei2026weighted}, and their covariance can be assessed online by plug-in or batch-means estimators, the bootstrap or random scaling \citep{fang2018,chen2020sgd,zhu2023,lee2022random,wei2026covariance}, also for quantile regression \citep{lee2025qr} and for Markovian or mixing data \citep{li2023markov,liu2023mixing}. These methods calibrate the covariance of the averaged iterate; the mean term studied here is of smaller order, but at stepsizes close to $t^{-1/2}$ it is comparable to the standard error.

\paragraph{Quantile inference under dependence.}
For i.i.d.\ data, quantile intervals invert the distribution function \citep{woodruff1952,sitter2001}, whereas studentized quantiles require a density estimate \citep{hall1988}. The Bahadur representation of the sample quantile \citep{bahadur1966} extends to dependent data \citep{wu2005bahadur}, with Berry--Esseen rates under mixing and local dependence \citep{lahiri2009quantile,dey2025quantile}. Intervals then estimate a long-run variance with a selected bandwidth \citep{newey1987,andrews1991,politis2003adaptive}, resample blocks \citep{kunsch1989,politis1992circular,lahiri2003book,sun2006quantile}, use blockwise empirical likelihood \citep{kitamura1997,chen2009sbel}, or self-normalize with non-Gaussian critical values \citep{lobato2001,shao2010self,shao2015review}. Our interval inverts chronological score tests with Gaussian critical values and an explicit window, is invariant under increasing transformations, and has a coverage guarantee for the implemented hull. It builds on normal approximation for locally dependent and self-normalized sums \citep{rinott1994,chen2004local,zhang2024self,cai2026refined}. Appendix~\ref{app:related} discusses further related work and compares assumptions and statistical objects.

\section{Setting and assumptions}\label{sec:setup}
Let $\mathcal F_t=\sigma(\varepsilon_s:s\le t)$ for i.i.d.\ innovations $\varepsilon_t$, let $F$ be the marginal distribution function of the stationary observations $X_t$, fix $\tau\in(0,1)$ and $q^\star$ with $F(q^\star)=\tau$, and write $\psi_t=\ind(X_t\le q^\star)-\tau$ for the target score. Quantile SGD with projection $\Pi_I$ onto a compact interval $I$ is
\begin{equation}\label{eq:sgd}
q_t=\Pi_I\{q_{t-1}-\eta_tg_t(q_{t-1})\},\qquad g_t(q)=\ind(X_t\le q)-\tau,
\end{equation}
with $\eta_t=c(t+t_0)^{-\gamma}$, $c>0$, $t_0\ge0$, $1/2<\gamma<1$ and deterministic $q_0\in I$. Put $\bar q_n=n^{-1}\sum_{t=1}^nq_t$ and $\bar\eta_n=n^{-1}\sum_{t=1}^n\eta_t$. We write $\Phi$ and $\phi$ for the standard normal distribution function and density, $z_p=\Phi^{-1}(p)$, and $x\vee y=\max(x,y)$.

\begin{assumption}[Finite innovation memory]\label{as:memory}
$X_t=G(\varepsilon_{t-r+1},\ldots,\varepsilon_t)$ for a fixed $r\ge1$, so observations $r$ or more steps apart are independent.
\end{assumption}
\begin{assumption}[Marginal regularity]\label{as:marginal}
$I=[\underline q,\overline q]$ contains $q^\star$ in its interior; $F$ has a bounded density $f$ on a neighborhood of $I$; $(q-q^\star)\{F(q)-\tau\}\ge\mu(q-q^\star)^2$ on $I$ for some $\mu>0$; and near $q^\star$, $h=f(q^\star)>0$ and $f'$ is continuous.
\end{assumption}
\begin{assumption}[Conditional density]\label{as:conditional}
$F_t(q)=\Pp(X_t\le q\mid\mathcal F_{t-1})$ has a density $f_t$, and $f_t,f_t'$ are bounded uniformly in $t$ on a fixed neighborhood of $q^\star$.
\end{assumption}
\begin{assumption}[Score calibration]\label{as:score}
The law of $X_t$ has no atom at $q^\star$; deterministic weights satisfy $w_t\ge0$, $\sum_tw_t^2=1$ and $w_*:=\max_tw_t\le C_wn^{-1/2}$; and $V_n:=\Var(\sum_tw_t\psi_t)\ge v_*>0$.
\end{assumption}
Assumption~\AMarg{} makes the population score pull the recursion toward $q^\star$. Assumption~\ACond{} is needed only when the threshold can depend on the recent past of $X_t$, and Assumption~\AScore{} concerns bounded scores only. No result requires moments of $X_t$, and the coverage results require no density.

\section{Directional bias and delayed thresholds}\label{sec:bias}
For $d\ge1$ define
\begin{equation}\label{eq:omega}
\begin{gathered}
\Gamma_k=\E(\psi_t\psi_{t-k}),\qquad \Omega=\Gamma_0+2\sum_{k=1}^{r-1}\Gamma_k,\\
\kappa_{r,d}=\sum_{k=d}^{r-1}\E\{\psi_{t-k}f_t(q^\star)\},\qquad \kappa_r=\kappa_{r,1},
\end{gathered}
\end{equation}
with empty sums equal to zero. Here $\Omega$ is the long-run variance of the target score, and $\kappa_r$ measures how recent scores predict the current conditional density at the target.

\begin{theorem}[Signed bias expansion]\label{thm:bias}
Under Assumptions~\AMem--\ACond,
\begin{equation}\label{eq:signed}
\frac{\E(q_t-q^\star)}{\eta_t}\longrightarrow B_\tau
:=\underbrace{\frac{\kappa_r}{h}}_{\text{interaction}}
-\underbrace{\frac{f'(q^\star)\Omega}{4h^2}}_{\text{curvature}},
\end{equation}
$\E(\bar q_n-q^\star)=B_\tau\bar\eta_n+o(n^{-\gamma})$, and the last iterate satisfies $\E(q_t-q^\star)^2/\eta_t\to\Omega/(2h)$.
\end{theorem}
The interaction vanishes for independent inputs, and the curvature term is the independent-input expansion with the score variance replaced by $\Omega$.

\paragraph{Mechanism.}
The threshold $q_{t-1}$ has just been pushed by the $r-1$ most recent observations. If those observations also predict whether $X_t$ falls near the threshold, that is, if they shift the conditional density of $X_t$ at $q^\star$, the update is biased in a fixed direction. Formally, $q_{t-r}$ is independent of $X_t$, and the recent displacement $\Delta_t=q_{t-1}-q_{t-r}$ is $O(\eta_t)$. With $\delta_t(q)=F_t(q)-F(q)$, an expansion across $\Delta_t$ gives
\begin{equation}\label{eq:conditionalbias}
\begin{split}
\E\delta_t(q_{t-1})&=\E[\{f_t(q_{t-r})-f(q_{t-r})\}\Delta_t]+o(\eta_t)\\
&=-\eta_t\kappa_r+o(\eta_t).
\end{split}
\end{equation}
This extra drift balances the restoring drift $h\eta_t\E(q_{t-1}-q^\star)$, and marginal curvature supplies the second term in \eqref{eq:signed} (Appendix~\ref{app:bias}). For $J_k(x,y)=\Pp(X_{t-k}\le x,X_t\le y)$, the interaction has the observable form
\begin{equation}\label{eq:observablekappa}
\kappa_{r,d}=\sum_{k=d}^{r-1}\{\partial_2J_k(q^\star,q^\star)-\tau h\},
\end{equation}
which depends on boundary derivatives of bivariate laws rather than on score covariances.

\begin{corollary}[Covariance-identical time reversal]\label{cor:reversal}
Suppose a stationary process and its time reversal both satisfy Assumptions~\AMem--\ACond, and mark their quantities by $\to$ and $\leftarrow$. They share the marginal law and all score autocovariances, hence $\Omega$ and $V_n$ for every deterministic weight vector, yet
\begin{equation}\label{eq:reversal}
B_\tau^{\to}-B_\tau^{\leftarrow}
=\frac1h\sum_{k=1}^{r-1}(\partial_2J_k-\partial_1J_k)(q^\star,q^\star).
\end{equation}
\end{corollary}
For example, let $U_t,Z_t$ be independent standard Gaussian sequences and
\begin{equation}\label{eq:nonreversible}
X_t=U_t+0.8U_{t-1}^2+0.7Z_t.
\end{equation}
The reversed process has the law of $U_{t-1}+0.8U_t^2+0.7Z_t$, and Gaussian smoothing verifies Assumption~\ACond{} in both directions. At $\tau=.25$, quadrature gives $B_\tau^{\to}=-.113107$ and $B_\tau^{\leftarrow}=-.038089$. Stationary Gaussian processes are time reversible \citep{weiss1975}, so such directional effects require non-Gaussian dependence \citep{lawrance1991}; a reversible process can still have nonzero $\kappa_r$. The leading bias-to-standard-deviation ratio of $\bar q_n$, $|B_\tau|\bar\eta_nh\sqrt n/\sqrt\Omega$, is of order $n^{1/2-\gamma}$; for \eqref{eq:nonreversible} with $c=3$, $t_0=20$ and $\gamma=.75$ it is $7.4\%$ at $n=3{,}000$ and $4.5\%$ at $n=48{,}000$, and it exceeds $21\%$ for $3{,}000\le n\le10^6$ when $\gamma=.55$ (Appendix~\ref{app:biasscale}).

\subsection{Delaying the threshold}
The expansion suggests changing which state enters the indicator, without discarding observations. For a fixed integer $d\ge1$, let $q_s^{(d)}=q_0$ for $s\le0$ and
\begin{equation}\label{eq:delayrecursion}
q_t^{(d)}=\Pi_I\{q_{t-1}^{(d)}-\eta_tg_t(q_{t-d}^{(d)})\}.
\end{equation}
Every observation is used on arrival with an older threshold, at $O(d)$ storage; $d=1$ is ordinary quantile SGD.

\begin{theorem}[Delay-specific interaction bias]\label{thm:delay}
Under Assumptions~\AMem--\AMarg, for every fixed $d\ge r$, the recursion \eqref{eq:delayrecursion} satisfies
\begin{align}
\frac{\E(q_t^{(d)}-q^\star)^2}{\eta_t}&\longrightarrow\frac{\Omega}{2h},\label{eq:delayvariance}\\
\frac{\E(q_t^{(d)}-q^\star)}{\eta_t}&\longrightarrow B_\tau^{\rm del}
:=-\frac{f'(q^\star)\Omega}{4h^2},\label{eq:delaybias}
\end{align}
and $\E(\bar q_n^{(d)}-q^\star)=B_\tau^{\rm del}\bar\eta_n+o(n^{-\gamma})$. Under Assumptions~\AMem--\ACond, every fixed $d\ge1$ has the same second-moment limit and mean coefficient
\begin{equation}\label{eq:delayprofile}
B_{\tau,d}=\frac{\kappa_{r,d}}h-\frac{f'(q^\star)\Omega}{4h^2},
\end{equation}
with $\E(\bar q_n^{(d)}-q^\star)=B_{\tau,d}\bar\eta_n+o(n^{-\gamma})$.
\end{theorem}
For $d\ge r$ the threshold is independent of the current observation, so the expected update uses only the marginal drift; for $d<r$ the lags $d,\ldots,r-1$ remain. The squared recursion involves $q_{t-1}^{(d)}$, whose recent updates recover the full $\Omega$ (Appendix~\ref{app:delay}). Increasing the delay by one isolates a single lag,
\begin{equation}\label{eq:adjacentdelay}
\frac{\E(\bar q_n^{(d)}-\bar q_n^{(d+1)})}{\bar\eta_n}
\longrightarrow \frac{\E\{\psi_{t-d}f_t(q^\star)\}}h,\quad d<r,
\end{equation}
and, for the directional contrast $D_d=\bar q_n^{(d),\to}-\bar q_n^{(d),\leftarrow}$,
\begin{equation}\label{eq:directionaldelay}
\frac{\E(D_d-D_{d+1})}{\bar\eta_n}
\longrightarrow\frac{(\partial_2J_d-\partial_1J_d)(q^\star,q^\star)}h,
\quad d<r.
\end{equation}

Delay changes the composition of the bias, not necessarily its size. Lag contributions can have different signs, and the interaction can partly cancel the curvature: in the Gaussian-copula Cauchy example of Appendix~\ref{app:cauchyexample}, a sufficient delay increases $|B_\tau|$ by $26\%$ and $53\%$ for memories two and four, and in the memory-four experiment of Section~\ref{sec:experiments} absolute bias and MSE vary non-monotonically with $d$ (Appendix Table~\ref{tab:profilemoments}). The delay is therefore an intervention that isolates and controls the interaction. Shared-stream two-gain extrapolation cancels the leading mean term, and with a sufficient delay it does so under Assumptions~\AMem--\AMarg{} alone (Appendix~\ref{app:rr}).

\section{Score inversion for the population quantile}\label{sec:method}
The adaptive threshold of Section~\ref{sec:bias} depends on recent data, so its mean requires directional information about joint laws. A fixed candidate keeps the score centered at the true quantile, so its first-order calibration requires only score covariances. Testing fixed candidates and inverting the tests therefore gives a separate route to inference for the population quantile. Exact inversion stores the $n$ observations, and its center is first-order equivalent to averaged SGD (Proposition~\ref{prop:equivalence}).

\subsection{Candidate scores and the interval}
For a candidate $q$ and weights as in Assumption~\AScore, write $s_t(q)=\ind(X_t\le q)-\tau$ and
\begin{align}
S_n(q)&=\sum_t w_ts_t(q),\label{eq:score}\\
\widehat V_b(q)&=\sum_tw_t^2s_t(q)^2\notag\\
&\quad+2\sum_{k=1}^{b}\sum_{t=k+1}^nw_tw_{t-k}s_t(q)s_{t-k}(q).
\label{eq:candvar}
\end{align}
Centering at the hypothesized $\tau$ makes $S_n(q^\star)$ mean zero, and
\begin{equation}\label{eq:tail}
\E\widehat V_b(q^\star)=V_n-T_{n,b},\quad T_{n,b}=2\sum_{k=b+1}^{r-1}\Gamma_k\sum_{t=k+1}^nw_tw_{t-k},
\end{equation}
so the estimate is unbiased when $b\ge r-1$, and $T_{n,b}$ is the omitted covariance otherwise. The rectangular lag kernel need not be positive semidefinite, unlike a Bartlett kernel \citep{newey1987}, so we floor the quadratic form. For nominal level $1-\alpha$, let $z=z_{1-\alpha/2}$ and $v_{\min,n}=1/n$, and define
\begin{equation}\label{eq:hullset}
\begin{split}
\mathcal A_n&=\{q:|S_n(q)|\le z\sqrt{\widehat V_b(q)\vee v_{\min,n}}\},\\
C^{\rm H}_{n,\alpha}&=\overline{\operatorname{conv}}(\mathcal A_n),
\end{split}
\end{equation}
written $C^{\rm H}_{n,\alpha}(b)$ when the window matters. Empty and unbounded outcomes are retained. A floor of order $b^2/n$ makes $\mathcal A_n$ connected (Appendix~\ref{app:inversion}) but can inflate intervals: for i.i.d.\ Cauchy data at $n=3{,}000$ it covers $99.4\%$, whereas the hull with the $1/n$ floor covers $95.4\%$ with $29\%$ shorter intervals (Appendix~\ref{app:floor}). Equal weights $w_t=n^{-1/2}$ give the default interval, E-hull, which is exactly invariant under reversing the data; finite-filter weights that mimic averaged SGD (S-hull, Appendix~\ref{app:hull}) perform similarly (Appendix~\ref{app:revisionexperiments}).

\subsection{Hull enlargement and local exactness}
Taking the hull can turn a rejected true value into a covered one. The following deterministic bound excludes this enlargement when the variance at the true candidate is not too small.

\begin{proposition}[Hull enlargement in score space]\label{prop:hullsandwich}
Put $\ell_{n,b}=(4b+1)w_*$. For $q\le q'$,
\begin{equation}\label{eq:scorelipmain}
|\widehat V_b(q')-\widehat V_b(q)|
\le\ell_{n,b}\{S_n(q')-S_n(q)\}.
\end{equation}
At any $q^\dagger\notin\{X_1,\ldots,X_n\}$ with $Q_0=\widehat V_b(q^\dagger)>v_{\min,n}$,
\begin{align}
\{|S_n(q^\dagger)|\le z\sqrt{Q_0}\}
&\subseteq\{q^\dagger\in C^{\rm H}_{n,\alpha}\}\notag\\
&\subseteq\{|S_n(q^\dagger)|\le z\sqrt{Q_0}+z^2\ell_{n,b}\}.
\label{eq:hullsandwich}
\end{align}
If moreover $Q_0\ge z^2\ell_{n,b}^2/4$, raw acceptance and hull membership at $q^\dagger$ coincide.
\end{proposition}
An indicator flip changes $S_n$ by $w_i$ and $\widehat V_b$ by at most $(4b+1)w_*w_i$, which gives \eqref{eq:scorelipmain}; a quadratic inequality then rules out accepted points beyond a rejected candidate. Consequently, for a fixed sufficient window, taking the hull adds an exponentially small probability to coverage, even when the acceptance set is disconnected away from the target (Appendix~\ref{app:sharp}).

\subsection{Coverage}
\begin{theorem}[Root-$n$ coverage of the implemented interval]\label{cor:sharpcoverage}
Under Assumptions~\AMem{} and \AScore, for fixed $r$, fixed $b\ge r-1$ and fixed $\alpha\in(0,1)$, the interval \eqref{eq:hullset} satisfies
\[
\big|\Pp(q^\star\in C^{\rm H}_{n,\alpha})-(1-\alpha)\big|
\le C_{r,b,\alpha,C_w,v_*}n^{-1/2}.
\]
\end{theorem}
Write $S=S_n(q^\star)$, $Q=\widehat V_b(q^\star)$ and $Y_i=\sum_{|j-i|\le b}w_j\psi_j$. Existing local-dependence theory gives this order for a self-normalized sum with the centered denominator $Q_c=Q-(S/n)\sum_iY_i$ \citep{zhang2024self,cai2026refined}. We transfer this order to the implemented interval. For $n>2b$ and the boundary term $E_\partial=\sum_j(2b+1-|\{i:|i-j|\le b\}|)w_j\psi_j$, the exact identity
\begin{equation}\label{eq:denominatoridentity}
Q_c=Q-\frac{(2b+1)S^2}{n}+\frac{S E_\partial}{n},\qquad |E_\partial|\le b(b+1)w_*,
\end{equation}
relates the two denominators. On either acceptance event $|S|$ is bounded, so the studentized statistics differ by $O(n^{-1})$; exponential concentration controls nonpositive denominators and keeps the $1/n$ floor inactive; and local exactness transfers raw acceptance to hull membership (Appendix~\ref{app:sharp}).

\begin{remark}[The order cannot be improved]\label{rem:lattice}
For i.i.d.\ data, $b=0$ and equal weights, $q^\star\in C^{\rm H}_{n,\alpha}$ exactly when a $\mathrm{Bin}(n,\tau)$ count lies in an integer interval with endpoints $n\tau\pm z\{n\tau(1-\tau)\}^{1/2}+O(1)$. The lattice term of the Edgeworth expansion then oscillates at exact order $n^{-1/2}$ \citep{bhattacharya1976,brown2002}, so $\limsup_nn^{1/2}|\Pp(q^\star\in C^{\rm H}_{n,\alpha})-(1-\alpha)|>0$. Exact binomial sums confirm the oscillation (Appendix~\ref{app:lattice}).
\end{remark}

\paragraph{Omitted dependence and data-selected windows.}
For any window, Theorem~\ref{thm:hullcoverage} in Appendix~\ref{app:sharp} bounds the coverage error by a Gaussian-approximation term, a variance-concentration term, $|T_{n,b}|/V_n$ and the hull term $\ell_{n,b}/\sqrt{V_n}$. Positive omitted covariance lowers the expected denominator and shortens intervals; negative omitted covariance lengthens them. The window may also be chosen from the data. Let $r_\psi=1+\max\{k\ge0:\Gamma_k\ne0\}\le r$, so that $r_\psi-1$ is the score memory.

\begin{corollary}[Data-selected windows]\label{cor:datawindow}
Under Assumptions~\AMem{} and \AScore, let $r_\psi-1\le b_{\max}\le n-1$, put $\mathcal W=\{r_\psi-1,\ldots,b_{\max}\}$, and let $\hat b$ be any window computed from $X_1,\ldots,X_n$. For $n\ge2$, with $C$ depending only on $r,\alpha,C_w,v_*$,
\begin{align*}
&\big|\Pp\{q^\star\in C^{\rm H}_{n,\alpha}(\hat b)\}-(1-\alpha)\big|\\
&\qquad\le C(b_{\max}+1)\sqrt{\log n/n}+\Pp(\hat b\notin\mathcal W).
\end{align*}
\end{corollary}
On the event that the denominators at $q^\star$ concentrate for all windows in $\mathcal W$ at once, Proposition~\ref{prop:hullsandwich} brackets hull membership between two score events that do not depend on $\hat b$ (Appendix~\ref{app:datawindow}); a fixed $b_{\max}$ gives the rate $\sqrt{\log n/n}$. Our rule is $\hat b=\min(2\hat m,\lceil n^{1/3}\rceil)$, where $\hat m$ is the correlogram rule of \citet{politis2003adaptive} applied to the indicators $\ind(X_t\le\widehat q_\tau)$ at the sample $\tau$-quantile (Appendix~\ref{app:v3intervals}). With $b_{\max}=\lceil n^{1/3}\rceil$, its coverage error is $O(n^{-1/6}\sqrt{\log n})+\Pp(\hat b<r_\psi-1)$; Appendix Table~\ref{tab:v3auto} reports the selected windows. The guarded rule $\max(\hat b,\lceil n^{1/4}\rceil)$ removes the last term once $\lceil n^{1/4}\rceil\ge r_\psi-1$, giving $O(n^{-1/6}\sqrt{\log n})$ for every fixed memory. The deterministic default $b_n=\lceil n^{1/4}\rceil$ eventually covers any fixed memory, with coverage error $O_r(n^{-1/4}\sqrt{\log n})$, and a known memory permits $b=r-1$.

\subsection{Exact computation}\label{sec:computation}
Let $x_{(1)}<\cdots<x_{(K)}$ be the distinct observed values. Scores are constant on $(-\infty,x_{(1)})$, $[x_{(j)},x_{(j+1)})$ and $[x_{(K)},\infty)$, so inversion is a scan over sorted values (Algorithm~\ref{alg:mainhull}). Each time index flips once, from $-\tau$ to $1-\tau$, and changes $\widehat V_b$ by $w_i^2(1-2\tau)+2w_i\sum_{0<|j-i|\le b}w_js_j$.

\begin{proposition}[Exact scan]\label{prop:fenwick}
Maintaining $a_j=w_js_j$ in a Fenwick tree indexed by time \citep{fenwick1994}, Algorithm~\ref{alg:mainhull} computes $C^{\rm H}_{n,\alpha}$, including ties and empty or unbounded outcomes, in $O(n\log n)$ time and $O(n)$ storage for every $b$. The output is equivariant under continuous strictly increasing transformations of the data.
\end{proposition}
Each neighbor sum costs two prefix queries and each flip one point update, so the scan has the same $O(n\log n)$ order as sorting. A direct $O(b)$ neighbor loop is comparably fast at the default windows, and the tree removes the computational constraint on large or data-selected windows (Appendix~\ref{app:fenwick}). Since $\ind\{\chi(X_t)\le\chi(q)\}=\ind(X_t\le q)$ for continuous strictly increasing $\chi$, scores and lag products are unchanged and endpoints map by $\chi$, so coverage is invariant under such transformations.

\begin{algorithm}[t]
\caption{Acceptance-hull inversion}\label{alg:mainhull}
\begin{algorithmic}[1]
\Require Chronological $X_{1:n}$, weights $w$, $\tau$, $b$, $z$.
\State Sort indices by $X_i$, keeping time locations; set $s_i=-\tau$, $S=-\tau\sum_iw_i$, $Q$ from \eqref{eq:candvar} via prefix sums of $w$, and a Fenwick tree over $a_i=w_is_i$.
\State Test $(-\infty,x_{(1)})$ using $|S|\le z\sqrt{Q\vee n^{-1}}$.
\For{each group of tied observations, in increasing order}
\For{each time index $i$ in the group}
\State $N\gets\textsc{Sum}(i-b,i+b)-a_i$ \Comment{prefix sums}
\State $Q\gets Q+w_i^2(1-2\tau)+2w_iN$; $S\gets S+w_i$; $s_i\gets s_i+1$; $\textsc{Add}(i,w_i)$.
\EndFor
\State Test $[x_{(j)},x_{(j+1)})$, or $[x_{(K)},\infty)$ for the last group; keep accepted endpoints.
\EndFor
\State Return the extreme accepted endpoints; retain empty or unbounded outcomes.
\end{algorithmic}
\end{algorithm}

\begin{figure*}[t]
\centering\includegraphics[width=.94\textwidth]{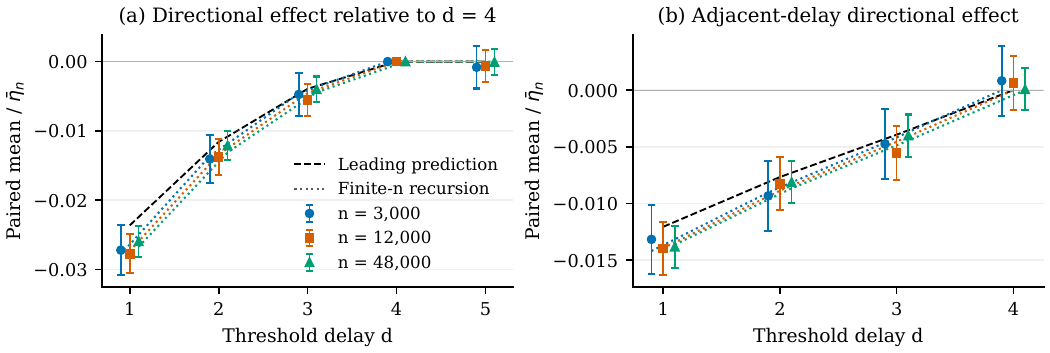}
\caption{Lag-specific directional effects in the memory-four model, 40,000 paired streams per sample size; $D_d$ is the forward-minus-reverse contrast for delay $d$. (a) Means of $(D_d-D_4)/\bar\eta_n$, zero at $d=4$ by construction. (b) Means of $(D_d-D_{d+1})/\bar\eta_n$, predicted by \eqref{eq:directionaldelay}. Bars: $\pm1.96$ paired MCSE; dashed: leading prediction; dotted: finite-$n$ recursion.}\label{fig:profile}
\end{figure*}
\subsection{Interval precision and connection to averaged SGD}
Let $W_n=\sum_tw_t$ and $\widehat q_{w,\tau}=\inf\{q:\sum_tw_t\ind(X_t\le q)\ge\tau W_n\}$; Assumption~\AScore{} implies $W_n\asymp\sqrt n$.

\begin{corollary}[Precision with a sufficient window]\label{cor:fixedprecision}
Under Assumptions~\AMem{} and \AScore{} with fixed $b\ge r-1$, suppose $F$ is differentiable at $q^\star$ with $h=F'(q^\star)>0$. With probability tending to one the hull is nonempty and bounded, with endpoints
\begin{equation}\label{eq:endpointsexpansion}
\widehat q_{w,\tau}\mp\frac{z\sqrt{V_n}}{hW_n}+\op(n^{-1/2}).
\end{equation}
\end{corollary}
The derivative controls width, but the algorithm does not estimate it. Theorem~\ref{thm:endpoints} in Appendix~\ref{app:precision} also covers growing windows and local alternatives. For equal and finite-filter weights, $W_n/\sqrt n\to1$ and $V_n\to\Omega$, so the first-order length is $2z\sqrt{\Omega}/(h\sqrt n)$, the length of an oracle interval that knows $h$ and $\Omega$.

\begin{proposition}[First-order equivalence]\label{prop:equivalence}
Under Assumptions~\AMem--\AMarg, for equal, finite-filter or symmetric finite-filter weights, $\sqrt n(\widehat q_{w,\tau}-\bar q_n)\to0$ in probability.
\end{proposition}
The proof combines weighted local score increments with the Bahadur representation of averaged SGD (Appendix~\ref{app:equivalence}).

\section{Numerical studies}\label{sec:experiments}
\begin{table*}[t]
\centering\small
\caption{Coverage (\%) / median length of nominal 95\% intervals, 1,600 common streams per row (coverage MCSE about .55 point). E-hull: equal weights, floor $1/n$, and $b=\lceil n^{1/4}\rceil$ or the data-selected $\hat b$ (median in parentheses). SN: self-normalization; CBB$_4$: circular block bootstrap. Last row: median single-thread time at $n=12{,}000$. Further results are in Appendix~\ref{app:v3intervals}.}\label{tab:block}
\vspace{2pt}
\begin{tabular}{llcccc}
\toprule
$n$ & Process & E-hull, $\lceil n^{1/4}\rceil$ & E-hull, $\hat b$ & SN & CBB$_4$\\
\midrule
3k & i.i.d. & 95.00 / 0.1962 & 95.38 / 0.1947 \,(0) & 95.50 / 0.2402 & 93.75 / 0.1906\\
3k & MA(1) & 95.06 / 0.2495 & 95.12 / 0.2489 \,(2) & 94.94 / 0.2999 & 94.19 / 0.2401\\
3k & MA(11) & 94.25 / 0.5455 & 95.06 / 0.5690 \,(15) & 95.38 / 0.6800 & 92.94 / 0.5310\\
12k & MA(11) & 96.12 / 0.2826 & 96.06 / 0.2834 \,(22) & 95.44 / 0.3431 & 95.00 / 0.2722\\
3k & AR(.8) & 94.38 / 0.4474 & 96.12 / 0.4734 \,(15) & 95.12 / 0.5586 & 93.56 / 0.4479\\
12k & AR(.8) & 94.19 / 0.2304 & 95.00 / 0.2383 \,(23) & 94.38 / 0.2887 & 93.19 / 0.2288\\
\midrule
\multicolumn{2}{l}{Time at 12k (ms)} & 1.26 & 1.50 & 0.51 & 9.66\\
\bottomrule
\end{tabular}
\end{table*}
We test the mechanism of Section~\ref{sec:bias} lag by lag, the size of the bias at practical sample sizes, and interval calibration. Methods share observations within each independent stream; Monte Carlo standard errors (MCSEs) use streams as units, with paired differences for method comparisons. Bias studies use $c=3$, $\gamma=.75$, $t_0=20$ and projection to $[-6,6]$. Every configuration was fixed before sampling (Appendix~\ref{app:experiments}).

\paragraph{Finite-sample predictions.}
The leading term $B_\tau\bar\eta_n$ omits two finite-sample effects: the recursion lags behind its decreasing stepsizes, by a relative amount $\gamma/(hc\,t^{1-\gamma})$ that is large for the early iterates, where the bias is largest, and the initialization adds an $O(n^{-1})$ transient. A finite-$n$ recursion retains both, using the drift terms of the proofs, the exact schedule and only model constants (Appendix~\ref{app:finite}).

\begin{figure}[!tb]
\centering\includegraphics[width=\columnwidth]{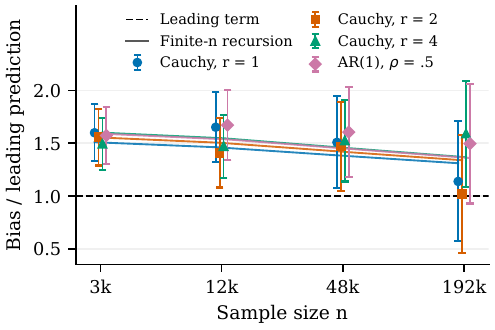}
\caption{Bias of averaged quantile SGD divided by the leading prediction $B_\tau\bar\eta_n$; 4,000 streams per point, bars $\pm1.96$ MCSE. Gaussian-copula Cauchy margins with memory $r$, and an AR(1) Gaussian copula outside finite memory.}\label{fig:biasscale}
\end{figure}
\subsection{Lag-specific mechanism}
For the memory-two model \eqref{eq:nonreversible}, 4,000 paired streams per sample size give ordinary-SGD directional contrasts within 1.2 MCSE of the leading prediction at all three sample sizes (Appendix Table~\ref{tab:reversal}). On 2,500 further streams per sample size, a delay $d=2$ removes the directional interaction: at $n=48{,}000$, the paired removal $D_1-D_2$ equals $-.0828\bar\eta_n$ (MCSE $.0042\bar\eta_n$), against $-.0870\bar\eta_n$ from the finite-$n$ recursion (Appendix Figure~\ref{fig:delay}). To separate individual lags we use $X_t=V_t+.6V_{t-1}^2+.4V_{t-2}^2+.2V_{t-3}^2+.7Z_t$, with $V_t$ uniform on $\{-1,0,1\}$ and $Z_t$ standard normal, and process each of 40,000 streams per sample size in both directions with delays $1,\ldots,5$. Figure~\ref{fig:profile} compares the paired contrasts with \eqref{eq:directionaldelay}. Across the twelve adjacent contrasts, the finite-$n$ recursion lies within 0.82 MCSE of the observations, and at $n=48{,}000$ the two sufficient delays $d=4,5$ differ by $.00007\bar\eta_n$ (MCSE $.00095\bar\eta_n$).

\subsection{Bias at practical sample sizes}
Figure~\ref{fig:biasscale} uses Gaussian-copula Cauchy margins with memories $r=1,2,4$ and an AR(1) Gaussian copula with $\rho=.5$, for which the coefficients are convergent series (Appendix~\ref{app:ar1}). Up to $n=48{,}000$, observed biases exceed the leading term by $41$--$67\%$, and the finite-$n$ recursion reproduces this gap: across the 16 cells it lies within 1.12 MCSE of the observations, whereas the leading term is up to 4.3 MCSE away. It also accounts for the Cauchy, reversal-contrast and initialization experiments (Appendix~\ref{app:finite}). Delay leaves the variance unchanged (ratios $1.00$--$1.03$ to ordinary SGD), whereas extrapolation raises it by $19$--$59\%$ (Appendix~\ref{app:rr}).

\subsection{Coverage and length}
Latent Gaussian processes are i.i.d., equal-coefficient moving averages with $r=2$ or $12$ innovations, or AR(1) with coefficient $.8$; $X_t=\tan\{\pi(\Phi(Y_t)-1/2)\}$ gives Cauchy margins and $q^\star=-1$ at $\tau=.25$, without clipping. Table~\ref{tab:block} compares E-hull with the power window and with $\hat b$ against the self-normalized interval of \citet{shao2010self}, with critical value 45.4 \citep{lobato2001}, and circular block bootstrap \citep[CBB$_4$: blocks of $4\lceil n^{1/3}\rceil$ observations, 399 resamples;][]{politis1992circular} on 1,600 new streams per cell.

With $\hat b$, E-hull covers $95.0$--$96.1\%$ in all six cells, including AR(.8), where the power window covers $94.2$--$94.4\%$; the selected window costs at most $6\%$ in median length. Under strong dependence $\hat b$ attains the coverage of the deterministic cap $\lceil n^{1/3}\rceil$, and under weak dependence it stays short (Appendix~\ref{app:v3intervals}). Self-normalization attains comparable coverage with intervals $20$--$25\%$ longer, the price of its non-Gaussian critical value. CBB$_4$ is within $4\%$ of E-hull with the power window in length but covers $0.8$--$1.3$ points less in paired comparisons and falls below the nominal level in five of six cells, at about eight times the computation; shorter blocks undercover more (Appendix~\ref{app:block600}). A random-scaling pivot for averaged SGD \citep{lee2022random} uses constant memory but covers $90.3$--$91.6\%$ here, mostly missing below the target. With stepsizes $3t^{-.505}$ its second-order bias is $0.47$--$0.82$ standard deviations, within $6\%$ of the leading term, and explains these one-sided misses (Appendix~\ref{app:v3intervals}). Appendix~\ref{app:wald} adds Wald, online-bootstrap \citep{chen2026smooth} and simultaneous intervals.

\begin{table}[t]
\centering\small
\caption{MA(11) window control: coverage (\%) / mean length, 1,600 paired streams per $n$, floor $1/n$. Omit is $100T_{n,b}/V_n$ for symmetric weights; at 12k the power and sufficient windows coincide.}\label{tab:fixedwindow}
\vspace{2pt}
{\setlength{\tabcolsep}{2.5pt}
\begin{tabular}{rrrcc}
\toprule
$n$ & $b$ & Omit (\%) & E-hull & S-hull\\
\midrule
3k & 3 & 44.0 & 85.31 / 0.4289 & 85.19 / 0.4310\\
3k & 8 & 6.7 & 94.44 / 0.5577 & 94.13 / 0.5608\\
3k & 11 & 0.0 & 95.19 / 0.5795 & 95.13 / 0.5829\\
12k & 4 & 33.5 & 89.56 / 0.2308 & 89.75 / 0.2315\\
12k & 11 & 0.0 & 94.44 / 0.2837 & 94.31 / 0.2846\\
\bottomrule
\end{tabular}}
\end{table}

\paragraph{Window choice and omitted covariance.}
Table~\ref{tab:fixedwindow} isolates $T_{n,b}$ on common MA(11) streams. It is shown that the logarithmic window $\lceil(\log n)^{1/2}\rceil$ omits 44\% of the score variance at $n=3{,}000$ and covers about 85\%; the power window omits 6.7\% and covers 94.4\%, and the sufficient window $b=11$ covers 95.2\%. Coverage tracks the omitted covariance, consistent with the term $|T_{n,b}|/V_n$ of Theorem~\ref{thm:hullcoverage} in Appendix~\ref{app:sharp}. On 11,727 daily JPY/USD log returns from 1980 to 2026 \citep{fred_dexjpus}, accounting for dependence widens the interval for the 5\% quantile by 31--72\% relative to i.i.d.\ calibration ($b=0$), increasingly with the window, consistent with persistent clustering of tail exceedances (Appendix~\ref{app:data}).

\section{Discussion}
Time reversal separates the directional mean of quantile SGD, governed by boundary derivatives, from the first-order calibration of score tests and the second moment, governed by score autocovariances. Threshold delays isolate the interaction lag by lag; with fixed candidates, scores stay centered and the implemented hull attains root-$n$ coverage for sufficient windows at $O(n\log n)$ cost.

For decaying-stepsize online calibration, the expansion identifies which lags of dependent scores bias the tracked threshold, and a delayed threshold, averaged over many streams or devices, gives a variance-neutral diagnostic of those lags. Extending the expansion to constant stepsizes and bootstrapped targets would cover adaptive conformal inference and quantile temporal-difference learning directly. Beyond finite innovation memory, the AR(1) experiments show that the bias coefficients remain predictive under geometric dependence, and coverage extends to a specified indicator-coupling envelope (Appendix~\ref{app:geometric}). Constant-memory calibration with comparable accuracy is a natural next step; a fixed candidate grid already admits streaming updates (Appendix~\ref{app:inversion}).

\clearpage
\subsection*{Code and Data Availability}
The code, experimental protocols, saved simulation records, and data used in the numerical studies are provided in the ancillary archive \texttt{replication.zip} accompanying this preprint.

\subsection*{AI Use Statement}
We used generative AI tools for literature search, drafting and checking mathematical arguments, manuscript editing, experimental design, code development, and running and checking simulations.

We have reviewed all AI-assisted work. Experimental code and numerical results were executed and verified by the authors.

We take responsibility for the final content of this work, including all text, claims, proofs, code, experiments, and other artifacts produced with the aid of generative AI.
\bibliographystyle{plainnat}
\bibliography{references}
\clearpage
\onecolumn
\raggedbottom
\appendix
\renewcommand{\theHtable}{appendix.\arabic{table}}
\renewcommand{\theHfigure}{appendix.\arabic{figure}}
\phantomsection\pdfbookmark[1]{Roadmap and notation}{app-roadmap}
\section*{Roadmap and notation}\label{app:roadmap}
The appendices are self-contained given Sections~\ref{sec:setup}--\ref{sec:method}. Appendices~\ref{app:sgd}--\ref{app:rr} concern quantile SGD, Appendices~\ref{app:inversion}--\ref{app:precision} concern score inversion, and Appendices~\ref{app:experiments}--\ref{app:wald} document the numerical studies. Within the theory, Appendix~\ref{app:bias} uses the moment bounds of Appendix~\ref{app:sgd}, Appendix~\ref{app:delay} reuses the expansion of Appendix~\ref{app:bias}, Appendix~\ref{app:rr} combines both with the Bahadur representation, and Appendix~\ref{app:sharp} relies on the deterministic geometry of Appendix~\ref{app:inversion}. Throughout, $C$ denotes a constant that may change from line to line; subscripts, as in $O_r$ or $C_{r,b}$, list the quantities on which it may depend, and it never depends on $n$ or $t$. Weights are written $w_t$ or $w_{nt}$.

\paragraph{Roadmap.}
\begin{center}
\small
\renewcommand{\arraystretch}{1.12}
\begin{tabular}{@{}>{\raggedright\arraybackslash}p{.06\textwidth}>{\raggedright\arraybackslash}p{.60\textwidth}>{\raggedright\arraybackslash}p{.28\textwidth}@{}}
\toprule
& Content & Main-text results\\
\midrule
\ref{app:sgd} & Stable products, finite-filter weights, high moments without conditional unbiasedness, and a quantitative Bahadur representation of averaged SGD (Lemmas~\ref{lem:products}--\ref{lem:sgdmoments}, Proposition~\ref{prop:bahadur}) & Theorems~\ref{thm:bias}, \ref{thm:delay}; Proposition~\ref{prop:equivalence}\\
\ref{app:bias} & Conditional drift across the moving threshold, moment and mean recursions, the observable form \eqref{eq:observablekappa}, verification of the examples, geometric memory, and the finite-$n$ moment recursion used for all predictions & Theorem~\ref{thm:bias}, Corollary~\ref{cor:reversal}; Section~\ref{sec:experiments}\\
\ref{app:delay} & Sufficient delays under Assumptions~\ref{as:memory}--\ref{as:marginal}; partial delays and the lag-resolved profile & Theorem~\ref{thm:delay}, \eqref{eq:adjacentdelay}--\eqref{eq:directionaldelay}\\
\ref{app:rr} & Shared-stream extrapolation, cross-gain covariance, second-order variance and MSE, a delayed version under Assumptions~\ref{as:memory}--\ref{as:marginal}, and the finite-sample cost (Corollaries~\ref{cor:rr}, \ref{cor:rrcost}, \ref{cor:delayedrr}) & Sections~\ref{sec:bias}, \ref{sec:experiments}\\
\ref{app:inversion} & Connected-floor interval, score-space Lipschitz bound and hull sandwich, connectedness, the Fenwick scan and its verification, finite-filter and symmetric weights & Propositions~\ref{prop:hullsandwich}, \ref{prop:fenwick}\\
\ref{app:sharp} & Dependency-graph and concentration tools; connected-floor and quantitative hull coverage (Theorems~\ref{thm:coverage}, \ref{thm:hullcoverage}); infinite memory; data-selected windows; local exactness; the lattice lower bound & Theorem~\ref{cor:sharpcoverage}, Remark~\ref{rem:lattice}, Corollary~\ref{cor:datawindow}\\
\ref{app:precision} & Endpoint expansion and local alternatives (Theorem~\ref{thm:endpoints}), weighted-quantile center, several quantiles, geometric indicator coupling & Corollary~\ref{cor:fixedprecision}, Proposition~\ref{prop:equivalence}\\
\ref{app:experiments} & Interval configurations, observation models, sampling budgets, pairing and implementation checks & Section~\ref{sec:experiments}\\
\ref{app:biasexp} & Cauchy, reversal, two-delay, initialization and delay-profile experiments & Figures~\ref{fig:profile}, \ref{fig:biasscale}\\
\ref{app:revisionexperiments} & Block lengths, variance floor, hull validation, window sensitivity, covariance tails, negative dependence, self-normalized and online comparisons, the bias of an SGD-based pivot, computing time, JPY/USD returns & Tables~\ref{tab:block}, \ref{tab:fixedwindow}\\
\ref{app:wald} & Wald intervals around averaged SGD, an online multiplier bootstrap and simultaneous intervals & Section~\ref{sec:experiments}\\
\ref{app:related} & Broader context and assumption-level comparison with the closest work & Section~\ref{sec:related}\\
\bottomrule
\end{tabular}
\end{center}

\paragraph{Notation.}
{\small
\renewcommand{\arraystretch}{1.12}
\begin{longtable}{@{}>{\raggedright\arraybackslash}p{.17\textwidth}>{\raggedright\arraybackslash}p{.57\textwidth}>{\raggedright\arraybackslash}p{.20\textwidth}@{}}
\toprule
Symbol & Meaning & Defined in\\
\midrule
\endfirsthead
\toprule
Symbol & Meaning & Defined in\\
\midrule
\endhead
\bottomrule
\endfoot
\multicolumn{3}{@{}l}{\emph{Data and recursions}}\\
$\varepsilon_t$, $\mathcal F_t$ & i.i.d.\ innovations and $\mathcal F_t=\sigma(\varepsilon_s:s\le t)$ & Section~\ref{sec:setup}\\
$X_t$, $r$ & stationary observations $X_t=G(\varepsilon_{t-r+1},\ldots,\varepsilon_t)$ with innovation memory $r$ & Assumption~\ref{as:memory}\\
$\tau$, $q^\star$ & quantile level and target quantile, $F(q^\star)=\tau$ & Section~\ref{sec:setup}\\
$F$, $f$, $h$ & marginal distribution function, its density, and $h=f(q^\star)$ & Assumption~\ref{as:marginal}\\
$F_t$, $f_t$, $\delta_t$ & conditional distribution function and density of $X_t$ given $\mathcal F_{t-1}$; $\delta_t(q)=F_t(q)-F(q)$ & Assumption~\ref{as:conditional}, \eqref{eq:conditionalbias}\\
$\psi_t$ & target score $\ind(X_t\le q^\star)-\tau$ & Section~\ref{sec:setup}\\
$g_t(q)$, $s_t(q)$ & score $\ind(X_t\le q)-\tau$ at an SGD threshold and at an inversion candidate & \eqref{eq:sgd}, Section~\ref{sec:method}\\
$\eta_t$, $\bar\eta_n$ & stepsize $c(t+t_0)^{-\gamma}$ with $1/2<\gamma<1$, and its average $n^{-1}\sum_{t\le n}\eta_t$ & \eqref{eq:sgd}\\
$I$, $\Pi_I$, $\mu$ & compact interval containing $q^\star$, projection onto it, and drift constant & \eqref{eq:sgd}, Assumption~\ref{as:marginal}\\
$q_t$, $\bar q_n$ & quantile-SGD iterate and its average & \eqref{eq:sgd}\\
$q_t^{(d)}$, $\bar q_n^{(d)}$ & recursion whose indicator uses the threshold $q^{(d)}_{t-d}$; $d=1$ is ordinary SGD & \eqref{eq:delayrecursion}\\
$e_t$, $m_t$ & error $q_t-q^\star$ (or $q^{(d)}_t-q^\star$) and its mean & Appendices~\ref{app:sgd}--\ref{app:delay}\\
$\widetilde m_t$, $\widetilde M_{1,t}$ & finite-$n$ moment recursion for $\E e_t$ and $\E e_t^2$ & Appendix~\ref{app:finite}\\
$\Delta_t$ & recent displacement $q_{t-1}-q_{t-r}$ of the threshold & \eqref{eq:conditionalbias}\\[3pt]
\multicolumn{3}{@{}l}{\emph{Dependence and bias coefficients}}\\
$\Gamma_k$, $\Omega$ & score autocovariance $\E(\psi_t\psi_{t-k})$ and long-run variance $\Gamma_0+2\sum_{k=1}^{r-1}\Gamma_k$ & \eqref{eq:omega}\\
$\kappa_{r,d}$, $\kappa_r$ & interaction $\sum_{k=d}^{r-1}\E\{\psi_{t-k}f_t(q^\star)\}$, and $\kappa_r=\kappa_{r,1}$ & \eqref{eq:omega}\\
$J_k$, $\partial_1J_k$, $\partial_2J_k$ & lag-$k$ bivariate distribution function $J_k(x,y)=\Pp(X_{t-k}\le x,X_t\le y)$ and its partial derivatives & \eqref{eq:observablekappa}\\
$B_\tau$, $B_{\tau,d}$, $B_\tau^{\rm del}$ & bias coefficients of ordinary, delay-$d$ and sufficiently delayed SGD & \eqref{eq:signed}, \eqref{eq:delaybias}, \eqref{eq:delayprofile}\\
$\to$, $\leftarrow$, $D_d$ & forward and time-reversed process; directional contrast $\bar q^{(d),\to}_n-\bar q^{(d),\leftarrow}_n$ & Corollary~\ref{cor:reversal}, \eqref{eq:directionaldelay}\\
$\widehat q_n^{\rm RR}$ & shared-stream extrapolation $2\bar q_n(c/2)-\bar q_n(c)$ & Corollary~\ref{cor:rr}\\
$r_\psi$ & $1+\max\{k\ge0:\Gamma_k\ne0\}$, so that $r_\psi-1$ is the score memory & Section~\ref{sec:method}\\[3pt]
\multicolumn{3}{@{}l}{\emph{Score inversion}}\\
$w_t$, $w_*$, $W_n$ & deterministic weights with $\sum_tw_t^2=1$, their maximum, and their sum & Assumption~\ref{as:score}, Section~\ref{sec:method}\\
$V_n$ & variance $\Var(\sum_tw_t\psi_t)$ of the target score sum & Assumption~\ref{as:score}\\
$A_{nt}(u_0)$, $w_t^{\rm sym}$ & finite-filter coefficients with nominal slope $u_0$, and their reversal symmetrization & \eqref{eq:filter}, \eqref{eq:symweights}\\
$S_n(q)$, $\widehat V_b(q)$ & weighted candidate score and lag-window variance; $S$ and $Q$ denote their values at $q^\star$ & \eqref{eq:score}, \eqref{eq:candvar}\\
$b$, $\hat b$, $b_{\max}$, $\mathcal W$ & lag window, data-selected window, its upper bound, and the set of sufficient windows & Section~\ref{sec:method}, Corollary~\ref{cor:datawindow}\\
$T_{n,b}$ & covariance omitted by the window $b$ & \eqref{eq:tail}\\
$\alpha$, $z$, $z_p$ & nominal level $1-\alpha$, $z=z_{1-\alpha/2}$, and $z_p=\Phi^{-1}(p)$ & Sections~\ref{sec:setup}, \ref{sec:method}\\
$v_{\min,n}$, $v_{\rm conn,n}$ & variance floors: $1/n$ for the hull, $(\lambda z(4b+1)w_*)^2$ for the connected-floor interval & \eqref{eq:hullset}, \eqref{eq:floor}\\
$\mathcal A_n$, $C^{\rm H}_{n,\alpha}$, $C_{n,\alpha}$ & acceptance set, its closed convex hull (the implemented interval), and the connected-floor interval & \eqref{eq:hullset}, \eqref{eq:floor}\\
$\ell_{n,b}$, $L_b$ & score-space Lipschitz constant $\ell_{n,b}=L_bw_*$, with $L_b=4b+1$ & Proposition~\ref{prop:hullsandwich}\\
$\beta_n$, $\epsilon_n(u)$ & Gaussian-approximation and variance-concentration terms & \eqref{eq:be}, \eqref{eq:dn}\\
$Q_c$, $E_\partial$ & centered self-normalizing denominator and boundary term & \eqref{eq:denominatoridentity}\\
$\widehat q_{w,\tau}$ & weighted $\tau$-quantile, the first-order center of the interval & Section~\ref{sec:method}\\[3pt]
\multicolumn{3}{@{}l}{\emph{General}}\\
$\Phi$, $\phi$, $\Phi_2(x,y;\rho)$ & standard normal distribution function and density; bivariate standard normal distribution function with correlation $\rho$ & Section~\ref{sec:setup}, Appendix~\ref{app:cauchyexample}\\
$\ind$, $x\vee y$ & indicator function; $\max(x,y)$ & Section~\ref{sec:setup}\\
$O_r$, $\Op$, $\op$ & order symbols, with subscripts listing what constants may depend on & ---\\
MCSE & Monte Carlo standard error & Section~\ref{sec:experiments}\\
E-hull, S-hull, Filter-hull, S-floor & hull interval with equal, symmetric finite-filter or original finite-filter weights; S-floor uses the connected floor & Section~\ref{sec:method}, Appendix~\ref{app:experiments}\\
SN, CBB$_j$, RS-SGD & self-normalized interval; circular block bootstrap with blocks of length $j\lceil n^{1/3}\rceil$; random-scaling interval for averaged SGD & Section~\ref{sec:experiments}, Appendix~\ref{app:v3intervals}\\
\end{longtable}}

\clearpage
\section{Moment bounds and the Bahadur representation}\label{app:sgd}
This appendix proves the moment bounds and the Bahadur representation used in Sections~\ref{sec:bias} and \ref{sec:method}, under Assumptions~\ref{as:memory}--\ref{as:marginal}. Constants may depend on the innovation memory $r$, the projection interval, the stepsize parameters and the density bounds, but not on $n$ or $t$.

\subsection{Two deterministic estimates}
For $\eta_t=c(t+t_0)^{-\gamma}$ with $0<\gamma<1$ and $a>0$, define $\pi_{s,t}(a)=\prod_{j=s+1}^t(1-a\eta_j)$. Finite early factors can be absorbed into constants; below, products are taken after a fixed index at which $a\eta_j\le1/2$.
\begin{lemma}[Stable products and recursion comparison]\label{lem:products}
For every $a>0$, $|\pi_{s,t}(a)|\le C\exp(-c_a\sum_{j=s+1}^t\eta_j)$. Consequently, for fixed $\iota>0$,
\[
\sum_{s=1}^t\eta_s^{1+\iota}|\pi_{s,t}(a)|\le C_\iota\eta_t^\iota.
\]
If a nonnegative sequence satisfies
$u_t\le(1-a\eta_t)u_{t-1}+C\eta_t^{1+\iota}$ for all sufficiently large $t$, then $u_t=O(\eta_t^\iota)$. If instead
$u_t=(1-a\eta_t)u_{t-1}+\varpi\eta_t^2+o(\eta_t^2)$,
where $u_t$ may be signed, then $u_t/\eta_t\to \varpi/a$.
\end{lemma}
\begin{proof}
The product bound follows from $1-x\le e^{-x}$. Split the sum at $t/2$. Its early part is exponentially small in $t^{1-\gamma}$, while on the late part $\eta_s\asymp\eta_t$ and
$\sum_{s>t/2}\eta_s |\pi_{s,t}(a)|\le C$, by telescoping or comparison with a geometric series. Iterating the recursion proves the upper bound. For the limit, write $y_t=u_t/\eta_t$ and use
$\eta_{t-1}/\eta_t=1+O(t^{-1})=1+o(\eta_t)$. The resulting recursion is
$y_t=(1-a\eta_t+o(\eta_t))y_{t-1}+\varpi\eta_t+o(\eta_t)$.
Subtract $\varpi/a$, bound the absolute value by a contracting recursion with forcing $\epsilon\eta_t$ after an arbitrarily late index, and let first $t\to\infty$ and then $\epsilon\downarrow0$.
\end{proof}

\begin{lemma}[Finite-filter weights]\label{lem:filters}
For fixed $u_0$ in a compact positive interval, with $\eta_1u_0<1$, the weights in \eqref{eq:filter} (Appendix~\ref{app:hull}) obey
\[
0\le A_{nt}(u_0)\le Cn^{-1/2},\qquad
\sum_{t=1}^n A_{nt}(u_0)^2\longrightarrow u_0^{-2}.
\]
For every $\epsilon\in(0,1/2)$,
$\sup_{\epsilon n\le t\le(1-\epsilon)n}|\sqrt n A_{nt}(u_0)-u_0^{-1}|\to0$.
If $p_{nt}=A_{nt}(u_0)/\sum_s A_{ns}(u_0)$, then
\begin{equation}\label{eq:filterl2}
n\sum_{t=1}^n(p_{nt}-n^{-1})^2\longrightarrow0.
\end{equation}
The absolute weight bound and bulk limit also hold with any fixed positive slope in place of $u_0$, after absorbing a finite early prefix; positivity is only needed for the inference weights. The exact linear identity for $e_t=(1-h\eta_t)e_{t-1}-\eta_t\psi_t$ is
\begin{equation}\label{eq:linearidentity}
\sqrt n\bar e_n=I_n-\sum_{t=1}^n A_{nt}(h)\psi_t,
\quad
I_n=\frac{e_0}{\sqrt n}\sum_{k=1}^n\prod_{j=1}^k(1-h\eta_j),
\quad |I_n|\le C|e_0|n^{-1/2}.
\end{equation}
\end{lemma}
\begin{proof}
Unrolling the linear recursion and interchanging its two finite sums gives \eqref{eq:linearidentity}. The stable-product bound makes the series defining $I_n\sqrt n/e_0$ summable. For the weights, the effective range of $k-t$ in \eqref{eq:filter} is of order $\eta_t^{-1}$. On $t\ge\epsilon n$, a window of length $O(\eta_t^{-1}\log n)$ is $o(t)$, so its learning rates are uniformly $\eta_t(1+o(1))$. The geometric sum multiplied by $\eta_t$ converges to $1/u_0$; for $t\le(1-\epsilon)n$ its tail beyond $n$ is negligible.
For a bound uniform down to the first index, divide $k\ge t$ into $[t,2t]$ and dyadic intervals $[2^jt,2^{j+1}t]$. The first contribution is at most $C/\eta_t$; subsequent contributions are bounded by $C2^jt\exp\{-c(2^jt)^{1-\gamma}\}$, whose sum multiplied by $\eta_t$ is bounded. This gives $\sqrt n A_{nt}\le C$.
The bulk convergence and this bound imply $n^{-1}\sum_t(\sqrt n A_{nt}-u_0^{-1})^2\to0$ by taking $n\to\infty$ then $\epsilon\downarrow0$. The norm and \eqref{eq:filterl2} follow. In particular, normalizing $A_n$ yields $w_*\le C_w/\sqrt n$.
\end{proof}

\subsection{High moments without assuming conditional unbiasedness}
\begin{lemma}[Moment control for the recursion]\label{lem:sgdmoments}
Under Assumptions~\ref{as:memory}--\ref{as:marginal}, for every fixed integer $p\ge1$,
$\E|q_t-q^\star|^{2p}\le C_{p,r}\eta_t^p$.
\end{lemma}

Write $e_t=q_t-q^\star$, $\nu=q_{t-r}-q^\star$, and $g_t(q)=\ind(X_t\le q)-\tau$. Nonexpansiveness of projection and $|g_t|\le1$ imply
\begin{equation}\label{eq:delaymovement}
|e_{t-1}-\nu|\le\sum_{j=t-r+1}^{t-1}\eta_j\le C_r\eta_t
\end{equation}
for all sufficiently large $t$. The variable $q_{t-r}$ is measurable with respect to $\mathcal F_{t-r}$ and is independent of $X_t$. Thus, for any $\mathcal F_{t-r}$-measurable threshold $q\in I$,
\begin{equation}\label{eq:frozenprob}
\E\{g_t(q)\mid\mathcal F_{t-r}\}=F(q)-\tau,
\quad
\Pp(|X_t-q|\le C_r\eta_t\mid\mathcal F_{t-r})\le C\eta_t.
\end{equation}
The finite-innovation representation gives independence at the frozen threshold $q_{t-r}$.

For $p\ge1$, polynomial expansion and projection give
\begin{align*}
\E|e_t|^{2p}
\le{}&\E|e_{t-1}|^{2p}
-2p\eta_t\E[e_{t-1}^{2p-1}g_t(q_{t-1})]\\
&+C_p\eta_t^2\E|e_{t-1}|^{2p-2}+C_p\eta_t^{2p}.
\end{align*}
Replacing the polynomial factor by $\nu^{2p-1}$ costs at most
$C\eta_t\E|\nu|^{2p-2}+C\eta_t^{2p-1}$. Replacing the indicator is bounded by the short-window event in \eqref{eq:frozenprob}, with the $\nu$-dependent polynomial factor held fixed. Since $I$ is bounded, that error obeys the same bound. The drift condition therefore yields
\[
\E[e_{t-1}^{2p-1}g_t(q_{t-1})]
\ge \mu\E|\nu|^{2p}-C\eta_t\E|\nu|^{2p-2}-C\eta_t^{2p-1}.
\]
For any fixed small $\delta>0$, \eqref{eq:delaymovement} and Young's inequality imply
$\E|e_{t-1}|^{2p}\le(1+\delta)\E|\nu|^{2p}+C_\delta\eta_t^{2p}$,
and $\E|\nu|^{2p-2}\le C\E|e_{t-1}|^{2p-2}+C\eta_t^{2p-2}$. Hence, with $M_{p,t}=\E|e_t|^{2p}$ and $M_{0,t}=1$,
\begin{equation}\label{eq:momentrecursion}
M_{p,t}\le(1-c_p\eta_t)M_{p,t-1}
+C_p\eta_t^2M_{p-1,t-1}+C_p\eta_t^{2p}.
\end{equation}
Induct on $p$ and apply Lemma~\ref{lem:products}: the forcing is $O(\eta_t^{p+1})$, so $M_{p,t}=O(\eta_t^p)$. Finite initial indices are absorbed into the constants. This proves the moment bounds used in Theorems~\ref{thm:bias} and \ref{thm:delay} without any observation-moment assumption.

\paragraph{Projection terms.}
Let $\delta_I=\min(q^\star-\underline q,\overline q-q^\star)>0$, and let $P_t$ be the difference between the projected and unprojected updates. For sufficiently large $t$,
\[
|P_t|\le\eta_t\ind\{|e_{t-1}|\ge\delta_I/2\},\qquad
\E|P_t|\le C_s\eta_t^{s+1}
\]
for every fixed $s$, by the $2s$-moment bound. Similar estimates hold in each fixed $L^p$ norm by using a higher moment. Thus projection contributes $o(\eta_t^2)$ to the mean and second-moment recursions below.

\subsection{A quantitative Bahadur representation}
\begin{proposition}[Raw averaged SGD]\label{prop:bahadur}
Under Assumptions~\ref{as:memory}--\ref{as:marginal}, for every fixed $p\ge2$,
\begin{align}
\sqrt n(\bar q_n-q^\star)&=-\sum_{t=1}^nA_{nt}(h)\psi_t+R_n,\label{eq:bahadur}\\
\norm{R_n}_p&\le C_{p,r}\{n^{1/2-\gamma}+n^{-\gamma/4}+n^{-1/2}\}.\label{eq:remainder}
\end{align}
Let $\mathrm d_{\rm K}(Y,Z)=\sup_x|\Pp(Y\le x)-\Pp(Z\le x)|$ denote Kolmogorov distance, also allowing a distribution in place of $Z$. If the variance of the linear sum is bounded away from zero, then its Gaussian approximation gives
\[
\mathrm d_{\rm K}\!\left(\sqrt n(\bar q_n-q^\star),N(0,\Var(\sum_tA_{nt}(h)\psi_t))\right)
\le C_{p,r}\{n^{-1/2}+n^{-\vartheta_{p,\gamma}}\},
\quad
\vartheta_{p,\gamma}=\frac{p}{p+1}\min(\gamma-1/2,\gamma/4).
\]
The remainder gives a quantitative upper bound on the Gaussian approximation error.
\end{proposition}
\begin{proof}
Let $H_t(q)=\ind(X_t\le q)-F(q)$. The score decomposition is
\[
g_t(q_{t-1})=\psi_t+h e_{t-1}+
\underbrace{F(q_{t-1})-F(q^\star)-h e_{t-1}}_{\Lambda_t}
+\underbrace{H_t(q_{t-1})-H_t(q^\star)}_{\Xi_t}.
\]
On a neighborhood of $q^\star$, $|\Lambda_t|\le C e_{t-1}^2$. On its complement, compactness and arbitrary high moments yield the same $L^p$ order $\norm{\Lambda_t}_p\le C_p\eta_t$.
Unrolling the stable linear recursion gives \eqref{eq:bahadur} with terms $I_n$, $-\sum_tA_{nt}\Lambda_t$, $-\sum_tA_{nt}\Xi_t$, and the projection correction $\sum_tA_{nt}P_t/\eta_t$.
By Lemma~\ref{lem:filters},
\[
\left\lVert\sum_tA_{nt}\Lambda_t\right\rVert_p
\le Cn^{-1/2}\sum_t\eta_t\le Cn^{1/2-\gamma}.
\]
The initial term is $O_{L^p}(n^{-1/2})$. For the projection term, use a sufficiently high moment so that $\norm{P_t/\eta_t}_p\le C\eta_t^a$ with $a\gamma>1$; its sum is $O(n^{-1/2})$.

For $t\ge r$, freeze the stochastic increment at $q_{t-r}$:
\[
\mathcal M_t=H_t(q_{t-r})-H_t(q^\star),\qquad
\mathcal N_t=H_t(q_{t-1})-H_t(q_{t-r}),\qquad \Xi_t=\mathcal M_t+\mathcal N_t.
\]
Then $\E(\mathcal M_t\mid\mathcal F_{t-r})=0$ and
$\E(\mathcal M_t^2\mid\mathcal F_{t-r})\le C|e_{t-r}|$.
For each residue class modulo $r$, $A_{nt}\mathcal M_t$ is a martingale difference sequence. The martingale moment inequality gives
\begin{align*}
\left\lVert\sum_tA_{nt}\mathcal M_t\right\rVert_p
&\le C_{p,r}\left\lVert\sum_t A_{nt}^2|e_{t-r}|\right\rVert_{p/2}^{1/2}
+C_{p,r}\left(\sum_t|A_{nt}|^p\E|\mathcal M_t|^p\right)^{1/p}\\
&\le C_{p,r}\left\{\left(n^{-1}\sum_t\eta_t^{1/2}\right)^{1/2}
+n^{-1/2}\left(\sum_t\eta_t^{1/2}\right)^{1/p}\right\}
\le C_{p,r}n^{-\gamma/4}.
\end{align*}
The last inequality holds for $p\ge2$. For $\mathcal N_t$, \eqref{eq:delaymovement}--\eqref{eq:frozenprob} give $\E(|\mathcal N_t|\mid\mathcal F_{t-r})\le C\eta_t$ and the same bound for its conditional second moment. Its conditional mean contributes $Cn^{-1/2}\sum_t\eta_t$; its centered residue-class martingale contributes no more than the preceding bound. The first $r-1$ terms are $O(n^{-1/2})$.

Apply the dependency-graph Gaussian bound to the bounded target-score linear sum. For any $u>0$, the difference between its distribution and that of the perturbed sum is bounded by $\Pp(|R_n|>u)$ plus a Gaussian interval probability of length $2u$ and the Gaussian approximation error. Markov's inequality gives $\norm{R_n}_p^p/u^p$; choosing $u\asymp\norm{R_n}_p^{p/(p+1)}$ proves the stated rate.
\end{proof}

\section{Signed bias expansion, reversal and examples}\label{app:bias}
This appendix proves Theorem~\ref{thm:bias} and Corollary~\ref{cor:reversal}, verifies the assumptions for the examples of Sections~\ref{sec:bias} and \ref{sec:experiments}, and derives the finite-$n$ recursion behind the predictions in Section~\ref{sec:experiments}.

\subsection{Conditional drift and cross-moment}
Let $\delta_t(q)=F_t(q)-F(q)$ and let $a_t=q_{t-r}$. By Assumption~\ref{as:conditional}, $f_t$ and $f_t'$ are uniformly bounded on a neighborhood of $q^\star$. Outside that neighborhood, high moments allow localization with an error $o(\eta_t)$ in the following equations. Set $\Delta_t=q_{t-1}-a_t$. Apart from projection terms already controlled,
\[
\Delta_t=-\sum_{k=1}^{r-1}\eta_{t-k}g_{t-k}(q_{t-k-1}),\qquad |\Delta_t|\le C_r\eta_t.
\]
Since $a_t$ is $\mathcal F_{t-r}$-measurable and $X_t$ is independent of $\mathcal F_{t-r}$,
\begin{equation}\label{eq:frozentower}
\E[\delta_t(a_t)\mid\mathcal F_{t-r}]=0.
\end{equation}
Taylor expansion across the displaced threshold gives
\[
\delta_t(a_t+\Delta_t)=\delta_t(a_t)+\{f_t(a_t)-f(a_t)\}\Delta_t+O(\Delta_t^2).
\]
Because $q_j\to q^\star$ in probability and the marginal distribution has no atom at $q^\star$, $g_j(q_{j-1})-\psi_j\to0$ in $L^1$. The bounded, locally Lipschitz density factors can likewise be replaced at $q^\star$ in expectation. Since $\eta_{t-k}/\eta_t\to1$ for fixed $k$, \eqref{eq:frozentower} proves
\begin{equation}\label{eq:driftdelta}
\E\delta_t(q_{t-1})
=-\eta_t\sum_{k=1}^{r-1}\E[\psi_{t-k}\{f_t(q^\star)-h\}]+o(\eta_t)
=-\eta_t\kappa_r+o(\eta_t).
\end{equation}
The same expansion multiplied by $e_{t-1}=a_t-q^\star+\Delta_t$ gives
\begin{equation}\label{eq:crossdelta}
\E[e_{t-1}\delta_t(q_{t-1})]
=-\eta_t\sum_{k=1}^{r-1}\Gamma_k+o(\eta_t).
\end{equation}
Indeed, the term $(a_t-q^\star)\delta_t(a_t)$ has expectation zero by \eqref{eq:frozentower}. The leading remaining term is $\Delta_t\delta_t(a_t)$. Its expectation equals
$-\eta_t\sum_k\E[\psi_{t-k}\delta_t(q^\star)]+o(\eta_t)$.
By conditional expectation, $\E[\psi_{t-k}\delta_t(q^\star)]=\E(\psi_{t-k}\psi_t)=\Gamma_k$.
Terms involving $(a_t-q^\star)\Delta_t$ times a bounded density derivative are $O(\eta_t^{3/2})$; terms with $\Delta_t^2$ are $O(\eta_t^2)$.

\subsection{Moment recursion and mean recursion}
Let $M_{1,t}=\E e_t^2$ and $m_t=\E e_t$. Expand the squared recursion. The marginal population drift contributes
$\E[e_{t-1}\{F(q_{t-1})-\tau\}]=h M_{1,t-1}+o(\eta_t)$.
The cross-moment \eqref{eq:crossdelta} supplies the lag-covariance terms, and bounded convergence gives
$\E g_t(q_{t-1})^2\to\tau(1-\tau)=\Gamma_0$.
Thus
\begin{equation}\label{eq:secondrec}
M_{1,t}=(1-2h\eta_t)M_{1,t-1}+\eta_t^2\Omega+o(\eta_t^2).
\end{equation}
Lemma~\ref{lem:products} yields $M_{1,t}/\eta_t\to\Omega/(2h)$. The Taylor expansion of the mean recursion, together with \eqref{eq:driftdelta}, is
\begin{align}
m_t&=(1-h\eta_t)m_{t-1}
-\tfrac12\eta_t f'(q^\star)M_{1,t-1}+\eta_t^2\kappa_r+o(\eta_t^2)\notag\\
&=(1-h\eta_t)m_{t-1}
+\eta_t^2\left\{\kappa_r-\frac{f'(q^\star)\Omega}{4h}\right\}+o(\eta_t^2).\label{eq:meanrec}
\end{align}
The same recursion lemma gives \eqref{eq:signed}. Ces\`aro summation with weights $\eta_t$ gives the averaged expansion because $\sum_{t=1}^n\eta_t\asymp n^{1-\gamma}$ diverges. This completes Theorem~\ref{thm:bias}.

\subsection{An observable form and the reversal corollary}
We prove Corollary~\ref{cor:reversal}.

For a lag $k<r$, differentiation under the expectation is valid by the uniform conditional-density bound. Therefore
\[
\E[\psi_{t-k}f_t(q^\star)]
=\left.\frac{\partial}{\partial y}\E[\psi_{t-k}\ind(X_t\le y)]\right|_{y=q^\star}
=\partial_2J_k(q^\star,q^\star)-\tau h.
\]
This proves \eqref{eq:observablekappa}. A stationary scalar process has $\Cov(\psi_t,\psi_{t-k})=\Cov(\psi_t,\psi_{t+k})$. Its reverse therefore has the same full score-covariance function. Its lagged bivariate CDF is $J_k^{\leftarrow}(x,y)=J_k^{\to}(y,x)$, so the difference of the two boundary coefficients is the right-hand side of \eqref{eq:reversal}. The marginal $h,f'$ and $\Omega$ agree, proving Corollary~\ref{cor:reversal}. A nonzero contrast thus requires a time-irreversible process, a property that can be tested without moment restrictions \citep{chen2000reversal}.

For equal weights, both $S_n(q)$ and the rectangular lag sum \eqref{eq:candvar} are exactly invariant under reversing a finite data sequence. Consequently, equal-weight score-inversion intervals are exactly reversal-invariant, even though the raw adaptive SGD estimator is not. With general finite-filter weights the two finite-data intervals need not coincide; the distributional covariance comparison in Corollary~\ref{cor:reversal} still holds.

\subsection{Verification of the smooth nonreversible example}\label{app:biasscale}
For $X_t=U_t+aU_{t-1}^2+sZ_t$ with $a=.8$, $s=.7$, write $v_0=(1+s^2)^{1/2}$ and let $U,V$ be independent standard normals. Define
\begin{align*}
A_c(u;q)&=\Phi((q-au^2)/v_0),& A_f(u;q)&=\phi((q-au^2)/v_0)/v_0,\\
B_c(u;q)&=\E_V\Phi((q-u-aV^2)/s),&B_f(u;q)&=\E_V\phi((q-u-aV^2)/s)/s.
\end{align*}
The common marginal CDF is $F(q)=\E A_c(U;q)$, $h=\E A_f(U;q^\star)$, and
\begin{align*}
\Gamma_1&=\E[A_c(U;q^\star)B_c(U;q^\star)]-\tau^2,\\
\kappa^{\to}&=\E[(B_c(U;q^\star)-\tau)A_f(U;q^\star)],\\
\kappa^{\leftarrow}&=\E[(A_c(U;q^\star)-\tau)B_f(U;q^\star)].
\end{align*}
At $\tau=.25$, quadrature yields $q^\star=-.302355668$, $h=.231437583$, $f'(q^\star)=.101619432$, $\Omega=.165742632$, $\kappa^{\to}=-.007983732$, and $\kappa^{\leftarrow}=.009378296$. The Gaussian $sZ_t$ convolution bounds every fixed conditional-density derivative uniformly, in both causal representations. The population density is positive on compact intervals, so the drift condition on $[-6,6]$ holds. The quadrature implementation checks 160 against 220 Hermite nodes and records its numerical discrepancy.

For $c=3$, $t_0=20$ and $\gamma=.55$, the leading bias-to-standard-deviation ratio $|B_\tau|\bar\eta_n h\sqrt n/\sqrt\Omega$ is $25.8\%$ at $n=3{,}000$ and $21.3\%$ at $n=10^6$ in the forward direction. Replacing $|B_\tau|$ by $|B_\tau^{\to}-B_\tau^{\leftarrow}|$ gives directional-difference ratios of $17.1\%$ and $14.1\%$, respectively. These calculations evaluate the leading expansion at $\gamma=.55$; the SGD simulations use $\gamma=.75$.

\subsection{Gaussian-copula Cauchy example}\label{app:cauchyexample}
Let $Z_t=r^{-1/2}\sum_{j=0}^{r-1}\varepsilon_{t-j}$ and $X_t=\tan\{\pi(\Phi(Z_t)-1/2)\}$. Write $\Phi_2(x,y;\rho)$ for the bivariate standard normal CDF with correlation $\rho$. With $z_\tau=\Phi^{-1}(\tau)$ and $\rho_k=(r-k)/r$,
\[
\Gamma_k=\Phi_2(z_\tau,z_\tau;\rho_k)-\tau^2,\qquad
\kappa_r=h\sum_{k=1}^{r-1}\left[\Phi\!\left(z_\tau\sqrt{\frac{1-\rho_k}{1+\rho_k}}\right)-\tau\right].
\]
At $\tau=.25$, $q^\star=-1$, $h=f(q^\star)=1/(2\pi)$ and $f'(q^\star)=1/(2\pi)$. For $r=1,2,4$, the bias coefficients are $-.294524311$, $-.377546255$, and $-.565889270$. For $r=2$ and $4$ the interaction coefficients $\kappa_r/h$ are $.0985$ and $.2985$ and the curvature coefficients are $-.4760$ and $-.8644$, so the interaction partly cancels the curvature and a sufficient delay increases $|B_\tau|$ by $26\%$ and $53\%$. Conditional densities and their derivatives are bounded on a compact quantile neighborhood because the latent conditional Gaussian variance is positive and the marginal transformation is smooth there.

\subsection{Geometric memory: an AR(1) Gaussian copula}\label{app:ar1}
The representation \eqref{eq:observablekappa} depends only on bivariate laws. For a Gaussian copula with latent correlation $\rho_k$ at lag $k$, margin $F$ and $z_\tau=\Phi^{-1}(\tau)$, one has $\partial_2J_k(q^\star,q^\star)=h\,\Phi\big(z_\tau\{(1-\rho_k)/(1+\rho_k)\}^{1/2}\big)$ and $\Gamma_k=\Phi_2(z_\tau,z_\tau;\rho_k)-\tau^2$; the formulas of Appendix~\ref{app:cauchyexample} are the case $\rho_k=(r-k)/r$. For the stationary latent AR(1) process $Y_t=\rho Y_{t-1}+(1-\rho^2)^{1/2}\varepsilon_t$, $\rho_k=\rho^k$, so
\[
\kappa_\infty=h\sum_{k\ge1}\Big[\Phi\Big(z_\tau\sqrt{\tfrac{1-\rho^k}{1+\rho^k}}\Big)-\tau\Big],\qquad
\Omega_\infty=\Gamma_0+2\sum_{k\ge1}\Gamma_k
\]
are series whose terms are $O(\rho^k)$. With Cauchy margins, $\tau=.25$ and $\rho=.5$, truncation at $\rho^k<10^{-13}$ gives $B_\infty=\kappa_\infty/h-f'(q^\star)\Omega_\infty/(4h^2)=-.4395$, and $-.6407$ after removing the lags $k<10$ for the delayed recursion with $d=10$. This process has infinite memory and lies outside Theorems~\ref{thm:bias}--\ref{thm:delay}. Nevertheless the coefficients, used in the finite-$n$ recursion with the same truncation, predict the observed bias at every sample size (Table~\ref{tab:biasscale}). Gaussian copulas are time reversible \citep{weiss1975}, so this model has no directional contrast.

\subsection{A finite-sample moment recursion}\label{app:finite}
The proofs of Theorems~\ref{thm:bias} and \ref{thm:delay} obtain the limits from recursions for $\E e_t$ and $\E e_t^2$ whose remainders are $o(\eta_t^2)$. Write $\kappa^{(k)}=\E\{\psi_{t-k}f_t(q^\star)\}$ for the lag-$k$ interaction. Retaining their drift terms with the actual schedule, the lag stepsizes and the initial error $e_0=q_0-q^\star$, rather than their limits, defines the finite-$n$ moment recursion
\begin{align*}
\widetilde M_{1,t}&=(1-2h\eta_t)\widetilde M_{1,t-1}+\eta_t^2\Gamma_0+2\eta_t\sum_{k=1}^{(r-1)\wedge(t-1)}\eta_{t-k}\Gamma_k,\\
\widetilde m_t&=\widetilde m_{t-1}-\eta_t\big\{h\widetilde m_{t-d}+\tfrac12f'(q^\star)\widetilde M_{1,t-d}\big\}+\eta_t\sum_{k=d}^{(r-1)\wedge(t-1)}\eta_{t-k}\kappa^{(k)},
\end{align*}
with $\widetilde m_s=e_0$ and $\widetilde M_{1,s}=e_0^2$ for $s\le0$. We predict $\E(\bar q^{(d)}_n-q^\star)$ by $n^{-1}\sum_{t=1}^n\widetilde m_t$ and $\E e_n^2/\eta_n$ by $\widetilde M_{1,n}/\eta_n$; extrapolated estimators combine the predictions for the two gains. By Lemma~\ref{lem:products}, $\widetilde m_t/\eta_t\to B_{\tau,d}$ and $\widetilde M_{1,t}/\eta_t\to\Omega/(2h)$, so the recursion has the leading limits. It retains two effects that the limits discard. First, writing $\widetilde m_t=a_t\eta_t$ and using $\eta_{t-1}/\eta_t=1+\gamma/t+O(t^{-2})$ shows that $a_t$ exceeds its limit in magnitude by a relative amount of order $\gamma/(hc\,t^{1-\gamma})$, which is large for the early iterates, where the bias is largest. Second, a nonzero $e_0$ adds a transient of order $n^{-1}$; it equals $2.16$, $.54$ and $.135$ ($\times10^{-3}$) for the ordinary-initialization experiment of Appendix~\ref{app:readability} at $n=3{,}000$, $12{,}000$ and $48{,}000$. Like the exact linear filter of Appendix~\ref{app:rrfinite}, the recursion is computed from model constants alone and uses no Monte Carlo output.

Table~\ref{tab:finiteagreement} summarizes the agreement across all bias experiments; the individual predictions appear in the corresponding tables. The leading term is systematically too small in magnitude at these sample sizes, whereas the finite-$n$ recursion is within $2.2$ MCSEs for every bias, second moment and directional contrast except in one shared-stream sample. Two discrepancies remain. First, the shared-stream reversal sample at $n=48{,}000$ has forward and reverse biases $2.7$ and $3.2$ MCSEs more negative than predicted. The two directions share streams, so this is essentially one draw of a common component; pooling this sample with the independent ordinary-initialization sample, after removing the predicted transient, gives $-.609$ (MCSE $.102$) forward and $-.349$ ($.102$) reverse ($\times10^{-3}$), against finite-$n$ values $-.437$ and $-.161$. Second, single-direction differences between consecutive delays carry a positive delay-specific term that neither prediction contains. It reaches $.005\bar\eta_n$ at $n=3{,}000$, appears even between the two sufficient delays $d=4,5$ ($.0023$--$.0025\bar\eta_n$ forward and $.0016$--$.0017\bar\eta_n$ reverse for $n\le12{,}000$), falls to about $.0008\bar\eta_n$ at $n=48{,}000$, and cancels in directional contrasts, which Figure~\ref{fig:profile} therefore uses.

\begin{table}[H]
\centering\small
\caption{Range of standardized discrepancies (observed minus prediction, divided by MCSE) for the leading term $B_{\tau,d}\bar\eta_n$ and for the finite-$n$ recursion.}\label{tab:finiteagreement}
\begin{tabular}{lrrr}
\toprule
Experiment & Cells & Leading & Finite-$n$\\
\midrule
Cauchy bias, Table~\ref{tab:bias} & 9 & -4.12 to -1.27 & -0.41 to 1.58\\
Cauchy RR bias, Table~\ref{tab:bias} & 9 & -- & -0.45 to 1.49\\
Cauchy second moment, Table~\ref{tab:bias} & 9 & 1.87 to 6.24 & -0.38 to 2.12\\
Reversal contrast, Table~\ref{tab:reversal} & 3 & -1.16 to 0.36 & -0.35 to 1.01\\
Reversal single-direction bias, Table~\ref{tab:reversal} & 6 & -3.48 to 0.46 & -3.17 to 1.66\\
Ordinary initialization, Table~\ref{tab:nonoraclefull} & 12 & 0.68 to 3.77 & -0.23 to 1.14\\
Initialization contrasts, Table~\ref{tab:nonoraclecontrasts} & 9 & -3.51 to 0.53 & -1.47 to 0.75\\
Delay profile, adjacent, Figure~\ref{fig:profile} & 12 & -1.87 to 0.52 & -0.82 to 0.66\\
Delay profile, single direction, Table~\ref{tab:profilesingle} & 24 & -0.49 to 5.46 & 0.02 to 4.45\\
Delay profile, bias, Table~\ref{tab:profilemoments} & 30 & -4.88 to -0.06 & -2.11 to 1.97\\
Bias scale, ordinary, Table~\ref{tab:biasscale} & 16 & -4.35 to -0.07 & -1.12 to 1.12\\
Bias scale, delayed, Table~\ref{tab:biasscale} & 12 & -6.80 to -0.29 & -1.74 to 1.08\\
Bias scale, RR and delayed RR, Table~\ref{tab:biasscale} & 28 & -- & -0.89 to 1.04\\
\bottomrule
\end{tabular}
\end{table}

The recursion also explains why shared-stream extrapolation leaves a residual bias at these sample sizes. The lag correction contributes to $\widetilde m_t$ a term of order $t^{-1}$ whose coefficient does not depend on the gain, because $\eta_tt^{\gamma-1}/c$ is free of $c$; any affine combination of gains with weights summing to one preserves it. For the Cauchy example with $r=2$, $n$ times the predicted RR bias divided by $\log n$ equals $-3.47$, $-3.65$, $-3.70$, $-3.70$ at $n=48{,}000$, $192{,}000$, $768{,}000$ and $3{,}072{,}000$, consistent with a residual of order $(\log n)/n$, while the ratio of the ordinary bias to $B_\tau\bar\eta_n$ is $1.42$, $1.34$, $1.27$, $1.21$.

\begin{table}[H]
\centering\footnotesize
\caption{Bias scale (prespecified; 4,000 streams per row, $q_0=q^\star$, gains 3 and 1.5). Biases and MCSEs $\times10^3$. Delay $d=r$ for moving averages and $d=10$ for AR(1); for $r=1$ the selected delay $d=r=1$ is ordinary SGD. Variance ratios are relative to ordinary SGD with gain 3.}\label{tab:biasscale}
\resizebox{\textwidth}{!}{%
\begin{tabular}{lrrrrrrrrrrrrr}
\toprule
 & & \multicolumn{3}{c}{Ordinary} & \multicolumn{2}{c}{Delayed} & \multicolumn{2}{c}{RR} & \multicolumn{2}{c}{Delayed RR} & \multicolumn{3}{c}{Variance ratio}\\
Model & $n$ & Leading & Finite-$n$ & Observed & Observed & Finite-$n$ & Observed & Finite-$n$ & Observed & Finite-$n$ & RR & Delayed RR & Delayed\\
\midrule
Cauchy, $r=1$ & 3k & -6.227 & -9.403 & -9.962 (0.859) & -- & -- & -4.055 (1.084) & -3.093 & -- & -- & 1.59 & -- & --\\
Cauchy, $r=1$ & 12k & -2.457 & -3.594 & -4.061 (0.417) & -- & -- & -2.064 (0.512) & -1.630 & -- & -- & 1.50 & -- & --\\
Cauchy, $r=1$ & 48k & -0.933 & -1.293 & -1.408 (0.207) & -- & -- & -0.572 (0.240) & -0.554 & -- & -- & 1.34 & -- & --\\
Cauchy, $r=1$ & 192k & -0.346 & -0.454 & -0.394 (0.101) & -- & -- & -0.077 (0.111) & -0.164 & -- & -- & 1.22 & -- & --\\
Cauchy, $r=2$ & 3k & -7.982 & -12.418 & -12.433 (1.092) & -15.287 (1.091) & -15.182 & -4.200 (1.357) & -4.276 & -5.328 (1.366) & -4.959 & 1.54 & 1.56 & 0.999\\
Cauchy, $r=2$ & 12k & -3.150 & -4.741 & -4.441 (0.529) & -5.582 (0.530) & -5.805 & -2.196 (0.646) & -2.288 & -2.582 (0.647) & -2.623 & 1.49 & 1.50 & 1.005\\
Cauchy, $r=2$ & 48k & -1.196 & -1.700 & -1.758 (0.258) & -2.147 (0.258) & -2.089 & -0.755 (0.298) & -0.780 & -0.850 (0.298) & -0.893 & 1.34 & 1.34 & 1.002\\
Cauchy, $r=2$ & 192k & -0.444 & -0.594 & -0.453 (0.127) & -0.596 (0.127) & -0.733 & -0.085 (0.140) & -0.231 & -0.123 (0.140) & -0.265 & 1.22 & 1.22 & 1.000\\
Cauchy, $r=4$ & 3k & -11.964 & -19.155 & -17.887 (1.494) & -26.621 (1.502) & -27.517 & -5.147 (1.844) & -6.849 & -7.077 (1.868) & -8.875 & 1.52 & 1.56 & 1.011\\
Cauchy, $r=4$ & 12k & -4.721 & -7.306 & -6.950 (0.719) & -10.430 (0.722) & -10.529 & -3.137 (0.872) & -3.726 & -4.021 (0.875) & -4.727 & 1.47 & 1.48 & 1.008\\
Cauchy, $r=4$ & 48k & -1.793 & -2.612 & -2.734 (0.354) & -3.954 (0.354) & -3.791 & -1.449 (0.410) & -1.274 & -1.845 (0.410) & -1.612 & 1.34 & 1.34 & 1.001\\
Cauchy, $r=4$ & 192k & -0.665 & -0.910 & -1.058 (0.170) & -1.482 (0.171) & -1.330 & -0.539 (0.187) & -0.378 & -0.640 (0.187) & -0.478 & 1.21 & 1.21 & 1.001\\
AR(1), $\rho=.5$ & 3k & -9.292 & -14.766 & -14.623 (1.294) & -22.489 (1.315) & -20.369 & -5.936 (1.593) & -5.227 & -7.163 (1.634) & -6.495 & 1.51 & 1.59 & 1.032\\
AR(1), $\rho=.5$ & 12k & -3.667 & -5.634 & -6.130 (0.622) & -8.879 (0.622) & -7.796 & -3.350 (0.769) & -2.833 & -4.077 (0.778) & -3.480 & 1.53 & 1.57 & 1.002\\
AR(1), $\rho=.5$ & 48k & -1.393 & -2.016 & -2.237 (0.301) & -3.197 (0.301) & -2.808 & -1.151 (0.346) & -0.968 & -1.366 (0.347) & -1.188 & 1.32 & 1.33 & 1.003\\
AR(1), $\rho=.5$ & 192k & -0.517 & -0.703 & -0.773 (0.150) & -1.102 (0.150) & -0.986 & -0.344 (0.164) & -0.287 & -0.425 (0.164) & -0.353 & 1.19 & 1.20 & 1.000\\
\bottomrule
\end{tabular}}
\end{table}

\section{Delayed thresholds}\label{app:delay}
This appendix proves Theorem~\ref{thm:delay}, first for sufficient delays $d\ge r$, which use only Assumptions~\ref{as:memory}--\ref{as:marginal}, and then for partial delays $1\le d<r$, which also use Assumption~\ref{as:conditional}.

\subsection{Sufficient delays}
\begin{proof}[Proof of Theorem~\ref{thm:delay}, sufficient delays]
Write $e_t=q_t^{(d)}-q^\star$, $\nu_t=e_{t-d}$, and $g_t(q)=\ind(X_t\le q)-\tau$. All indices at most zero refer to the fixed initialization. The movement bound is $|e_{t-1}-\nu_t|\le C_d\eta_t$. Because $d\ge r$, $q_{t-d}^{(d)}$ is measurable with respect to $\mathcal F_{t-d}$ and independent of $X_t$. In particular,
\begin{equation}\label{eq:delayexact}
\E g_t(q_{t-d}^{(d)})=\E\{F(q_{t-d}^{(d)})-\tau\}.
\end{equation}
This equality follows by conditioning on $\mathcal F_{t-d}$; successive update scores remain dependent.

For the $2p$-th moment, expand the projected update as in Appendix~\ref{app:sgd}. Replacing the polynomial factor $e_{t-1}^{2p-1}$ by $\nu_t^{2p-1}$ costs at most $C_d\eta_t\E|\nu_t|^{2p-2}+C_d\eta_t^{2p-1}$. There is no threshold-replacement error, since the threshold already equals $q_{t-d}^{(d)}$. Conditional independence and the drift condition give $\E[\nu_t^{2p-1}g_t(q_{t-d}^{(d)})]\ge\mu\E|\nu_t|^{2p}$. Young's inequality then yields the same contracting moment recursion \eqref{eq:momentrecursion}, with constants depending on $d$. Induction gives $\E|e_t|^{2p}=O(\eta_t^p)$ for every fixed $p$. As before, projection corrections have arbitrarily high polynomial order.

For the second moment put $\Delta_t=e_{t-1}-\nu_t$. Up to negligible projection terms,
\[
\Delta_t=-\sum_{k=1}^{d-1}\eta_{t-k}g_{t-k}(q_{t-k-d}^{(d)}).
\]
The moment bound implies convergence of each adapted score to its target score in every fixed $L^p$: boundedness and the event $|X_t-q^\star|\le |q_{t-d}^{(d)}-q^\star|$ suffice. Thus, using $\eta_{t-k}/\eta_t\to1$,
\[
\E[\Delta_t g_t(q_{t-d}^{(d)})]
=-\eta_t\sum_{k=1}^{d-1}\Gamma_k+o(\eta_t)
=-\eta_t\sum_{k=1}^{r-1}\Gamma_k+o(\eta_t).
\]
The other term equals $\E[\nu_t\{F(q^\star+\nu_t)-\tau\}]=h\E \nu_t^2+o(\eta_t)$. Moreover,
$|\E e_{t-1}^2-\E \nu_t^2|\le C_d\eta_t\E|\nu_t|+C_d\eta_t^2=O(\eta_t^{3/2})$.
Consequently the squared recursion is
\[
\E e_t^2=(1-2h\eta_t)\E e_{t-1}^2+\eta_t^2\Omega+o(\eta_t^2),
\]
which gives the claimed limit by Lemma~\ref{lem:products}. The coefficient retains every target-score covariance in $\Omega$.

Let $m_t=\E e_t$. Equation~\eqref{eq:delayexact}, Taylor expansion, and the moment bounds give
\[
m_t=m_{t-1}-\eta_t\{h m_{t-d}+\tfrac12f'(q^\star)\E e_{t-d}^2+o(\eta_t)\}.
\]
Each mean increment has magnitude $O(\eta_t^{3/2})$, by $\E|e_{t-d}|=O(\sqrt{\eta_t})$ and localization of the marginal drift. Hence $m_{t-d}-m_{t-1}=O_d(\eta_t^{3/2})$, so substituting the second-moment limit yields
\[
m_t=(1-h\eta_t)m_{t-1}-\eta_t^2\frac{f'(q^\star)\Omega}{4h}+o(\eta_t^2).
\]
The signed recursion comparison and weighted Ces\`aro summation prove both mean assertions. The sufficient-delay proof uses marginal regularity and finite innovation memory, because the frozen threshold is independent of the current observation.
\end{proof}

\subsection{Partial delays and lag-specific interactions}\label{app:partialdelay}
\begin{proof}[Proof of Theorem~\ref{thm:delay}, partial delays]
Assumptions~\ref{as:memory}--\ref{as:conditional} are in force. For $d\ge r$ the sum defining $\kappa_{r,d}$ is empty and the sufficient-delay proof applies; if $r=1$ every delay is sufficient. Fix $1\le d<r$, write $e_t=q_t^{(d)}-q^\star$, and freeze at $a_t=q^{(d)}_{t-r}$, which is $\mathcal F_{t-r}$-measurable and independent of $X_t$. Bounded updates give $|q^{(d)}_{t-1}-a_t|+|q^{(d)}_{t-d}-a_t|\le C\eta_t$; constants may depend on $r$, $d$ and the moment order.

\emph{Moments.} The argument of Lemma~\ref{lem:sgdmoments} applies with $\nu=a_t-q^\star$. Replacing $e_{t-1}^{2p-1}$ by $\nu^{2p-1}$ costs $C\eta_t\E|\nu|^{2p-2}+C\eta_t^{2p-1}$, and replacing the threshold $q^{(d)}_{t-d}$ by $a_t$ costs at most $\E[|\nu|^{2p-1}\Pp\{|X_t-a_t|\le C\eta_t\mid\mathcal F_{t-r}\}]\le C\eta_t\E|\nu|^{2p-2}$ by the bounded marginal density and compactness of $I$. Independence and the drift condition give $\E\{\nu^{2p-1}g_t(a_t)\}\ge\mu\E|\nu|^{2p}$, so \eqref{eq:momentrecursion} holds for the delayed states and $\E|e_t|^{2p}=O(\eta_t^p)$ for every $p$. As in Appendix~\ref{app:sgd}, projection corrections and contributions from outside a fixed neighborhood of $q^\star$ have arbitrarily high polynomial order, so the expansions below may be localized.

\emph{Conditional drift.} Up to projection terms, $\Delta_{t,j}=q^{(d)}_{t-j}-a_t=-\sum_{k=j}^{r-1}\eta_{t-k}g_{t-k}(q^{(d)}_{t-k-d})$ for $j\in\{1,d\}$. With $\delta_t=F_t-F$, frozen independence gives $\E\{\delta_t(a_t)\mid\mathcal F_{t-r}\}=0$, and the expansion of Appendix~\ref{app:bias} with $\Delta_{t,d}$ in place of $\Delta_t$ yields
\begin{equation}\label{eq:partialdrift}
\E\delta_t(q_{t-d}^{(d)})
=-\eta_t\sum_{k=d}^{r-1}\E\{\psi_{t-k}(f_t(q^\star)-h)\}+o(\eta_t)
=-\eta_t\kappa_{r,d}+o(\eta_t),
\end{equation}
where the second equality uses $\E\psi_{t-k}=0$. Here each adapted score converges in $L^1$ to its target score, because its mismatch probability is at most $\Pp(|X_s-q^\star|\le a)+\Pp(|q^{(d)}_{s-d}-q^\star|>a)$ for every $a>0$, and $\eta_{t-k}/\eta_t\to1$. The sum starts at $k=d$ because the threshold is $q^{(d)}_{t-d}$.

\emph{Second moment.} The squared update involves the current state $e_{t-1}=\nu+\Delta_{t,1}$, whose displacement sums from $k=1$. Frozen independence gives $\E\{\nu g_t(a_t)\}=\E[\nu\{F(a_t)-\tau\}]=h\E\nu^2+o(\eta_t)$, and, as in the sufficient-delay proof, $\E\{\Delta_{t,1}g_t(a_t)\}=-\eta_t\sum_{k=1}^{r-1}\Gamma_k+o(\eta_t)$, while replacing $a_t$ by $a_t+\Delta_{t,d}$ inside $g_t$ costs $O(\eta_t^{3/2})$ by the same crossing bound. With $|\E e_{t-1}^2-\E\nu^2|=O(\eta_t^{3/2})$ and $\E g_t(q^{(d)}_{t-d})^2\to\Gamma_0$,
\begin{equation}\label{eq:partialsecond}
\E e_t^2=(1-2h\eta_t)\E e_{t-1}^2+\eta_t^2\Omega+o(\eta_t^2),
\end{equation}
and Lemma~\ref{lem:products} gives $\E e_t^2/\eta_t\to\Omega/(2h)$.

\emph{Mean.} By \eqref{eq:partialdrift} and the marginal Taylor expansion, $m_t=\E e_t$ satisfies $m_t=m_{t-1}-h\eta_tm_{t-d}-\frac12f'(q^\star)\eta_t\E e_{t-d}^2+\eta_t^2\kappa_{r,d}+o(\eta_t^2)$. Mean increments are $O(\eta_t^{3/2})$, because the expected marginal drift is $O(\sqrt{\eta_t})$ and \eqref{eq:partialdrift} is $O(\eta_t)$; hence $m_{t-d}-m_{t-1}=O(\eta_t^{3/2})$, and the second-moment limit gives
\[
m_t=(1-h\eta_t)m_{t-1}+\eta_t^2\Big\{\kappa_{r,d}-\frac{f'(q^\star)\Omega}{4h}\Big\}+o(\eta_t^2).
\]
Lemma~\ref{lem:products} and weighted Ces\`aro summation give \eqref{eq:delayprofile} and the averaged expansion. Subtracting consecutive delays gives \eqref{eq:adjacentdelay}, and $d=1$ recovers Theorem~\ref{thm:bias}.
\end{proof}

The bivariate representation is
$\kappa_{r,d}=\sum_{k=d}^{r-1}\{\partial_2J_k(q^\star,q^\star)-\tau h\}$.
For a process and its reverse satisfying the respective assumptions, their coefficients differ by
\begin{equation}\label{eq:profilereversal}
B_{\tau,d}^{\to}-B_{\tau,d}^{\leftarrow}
=h^{-1}\sum_{k=d}^{r-1}(\partial_2J_k-\partial_1J_k)(q^\star,q^\star).
\end{equation}
Differencing \eqref{eq:profilereversal} in $d$ gives \eqref{eq:directionaldelay}.

\section{Shared-stream extrapolation}\label{app:rr}
This appendix proves that two-gain extrapolation on a shared stream removes the leading bias without changing the first-order variance, derives its second-order variance and MSE, gives a delayed version under Assumptions~\ref{as:memory}--\ref{as:marginal}, and quantifies the finite-sample cost.

\begin{corollary}[Shared-stream extrapolation]\label{cor:rr}
Under Assumptions~\ref{as:memory}--\ref{as:conditional}, run gains $c$ and $c/2$ on the same observations, with the same $\gamma,t_0$, and set $\widehat q_n^{\rm RR}=2\bar q_n(c/2)-\bar q_n(c)$. Then $\E(\widehat q_n^{\rm RR}-q^\star)=o(n^{-\gamma})$. If $\Omega>0$,
\[
\sqrt n(\widehat q_n^{\rm RR}-q^\star)\Rightarrow N(0,\Omega/h^2),\qquad
n\Var(\widehat q_n^{\rm RR})\to\Omega/h^2.
\]
\end{corollary}
Shared-stream extrapolation preserves the leading variance in constant-step smooth SGD and Markovian linear SA \citep{huo2024nonlinear,sheshukova2025rr,levin2026}, and multi-step Richardson--Romberg extrapolation also reduces discretization bias in SA, including diffusion-quantile estimation \citep{frikha2015}. Here the decreasing-step boundary expansion and a common $L^2$ score term establish the corresponding bias cancellation and variance limit for raw quantile SGD.

\begin{proof}[Proof of Corollary~\ref{cor:rr}]
The coefficient $B_\tau$ in Theorem~\ref{thm:bias} is independent of the fixed gain. Since $\bar\eta_n(c/2)=\bar\eta_n(c)/2$, its two mean expansions cancel, giving the asserted $o(n^{-\gamma})$ bias. For either gain $a\in\{c,c/2\}$, Proposition~\ref{prop:bahadur} with $p=2$ and Lemma~\ref{lem:filters} give
\[
\sqrt n\{\bar q_n(a)-q^\star\}=-\frac1{h\sqrt n}\sum_{t=1}^n\psi_t+o_{L^2}(1).
\]
To see the filter replacement in $L^2$, finite dependence bounds the variance of a score sum with coefficients $v_t$ by $C_r\sum_tv_t^2$; apply \eqref{eq:filterl2} at the true slope $h$. The leading random variable is identical for both gains because they share the observations. Its coefficient in the combination is $2-1=1$. The finite-dependence central limit theorem and $n^{-1}\Var(\sum_t\psi_t)\to\Omega$ prove the distributional limit. The $L^2$ remainder also proves the variance limit.
\end{proof}
\subsection{Cross-gain covariance of the average filter}\label{app:crossgain}
The finite filter identifies the variance contribution that remains after leading-order bias cancellation. We display its gain dependence explicitly:
\[
A_{nt}(h;c)=\frac{c(t+t_0)^{-\gamma}}{\sqrt n}
\sum_{k=t}^n\prod_{j=t+1}^k\{1-hc(j+t_0)^{-\gamma}\},\qquad
\mathcal L_n(c)=-\sum_{t=1}^n A_{nt}(h;c)\psi_t.
\]
These coefficients correspond to the full average of iterates $1,\ldots,n$.

\begin{proposition}[Cross-gain covariance]\label{prop:crossgain}
Let $(\psi_t)_{t\in\mathbb Z}$ be centered and second-order stationary. Suppose
\[
\Gamma_k=\Cov(\psi_0,\psi_k),\qquad
\sum_{k\in\mathbb Z}(1+|k|)|\Gamma_k|<\infty,\qquad
\Omega=\sum_{k\in\mathbb Z}\Gamma_k>0.
\]
For fixed $h>0$, $t_0\ge0$, $1/2<\gamma<1$ and fixed gains $c,c'>0$,
\begin{equation}\label{eq:crossgain}
\Cov\{\mathcal L_n(c),\mathcal L_n(c')\}
=\frac{\Omega}{h^2}+\frac{\Omega}{h^3(c+c')}n^{\gamma-1}+o(n^{\gamma-1}).
\end{equation}
For the linear recursion $e_t(c)=\{1-h\eta_t(c)\}e_{t-1}(c)-\eta_t(c)\psi_t$ with deterministic initialization, the left side equals $n\Cov\{\bar e_n(c),\bar e_n(c')\}$.
\end{proposition}

\begin{proof}
For one gain, put $a_t=hc(t+t_0)^{-\gamma}$ and
\[
\pi_{t,k}=\prod_{j=t+1}^k(1-a_j),\qquad
K_{n,t}=h\sqrt n A_{nt}(h;c)=a_t\sum_{k=t}^n\pi_{t,k}.
\]
We first assume $0<a_1<1$. The product bound
$\pi_{t,k}\le\exp(-\sum_{j=t+1}^ka_j)$ makes
$K_{\infty,t}=a_t\sum_{k=t}^\infty \pi_{t,k}$ finite. Telescoping gives
\begin{equation}\label{eq:crossbulk}
\sum_{k=t}^\infty a_{k+1}\pi_{t,k}=1,\qquad
K_{\infty,t}=1+\varsigma_t,\qquad
\varsigma_t=\sum_{k=t}^\infty(a_t-a_{k+1})\pi_{t,k}.
\end{equation}
For $t\le k\le2t$, the mean-value theorem bounds the summand's step difference by $Ca_t(k+1-t)/t$. The product is at most $\exp\{-C_1(k-t)t^{-\gamma}\}$. Summing this bound yields $Ct^{\gamma-1}$. On successive dyadic intervals beyond $2t$, the product contributes $\exp\{-C_2(2^j t)^{1-\gamma}\}$, which dominates their polynomial lengths. Thus
\begin{equation}\label{eq:crossbulkbound}
0\le \varsigma_t\le C(t+t_0)^{\gamma-1},\qquad
\sup_{n,t\le n}|K_{n,t}|+\sup_t|K_{\infty,t}|\le C,\qquad
\sum_{t=1}^n \varsigma_t^2=O(n^{2\gamma-1}).
\end{equation}
Finite initial indices are absorbed into $C$.

The exact truncation identity is
\begin{equation}\label{eq:crossterminal}
K_{n,t}=K_{\infty,t}-E_{n,t},\qquad
E_{n,t}=\frac{a_t}{a_{n+1}}K_{\infty,n+1}\pi_{t,n+1}.
\end{equation}
For $t=n-\lfloor xn^\gamma\rfloor$ with fixed $x\ge0$, the gain ratio tends to one and
$\log \pi_{t,n+1}\to-hcx$. Indeed, the gains over this terminal range sum to $hcx+o(1)$, while their squares sum to $O(n^{-\gamma})$. It follows that $E_{n,n-\lfloor xn^\gamma\rfloor}(c)\to e^{-hcx}$.
For $t\ge n/2$, the bound $|E_{n,t}(c)|\le C\exp\{-C_3(n-t)n^{-\gamma}\}$ gives an integrable envelope. For $t<n/2$, the sum is bounded by a polynomial in $n$ times $\exp(-C_4n^{1-\gamma})$. Riemann sums therefore give
\begin{equation}\label{eq:crossintegral}
n^{-\gamma}\sum_{t=1}^nE_{n,t}(c)E_{n,t}(c')
\longrightarrow\int_0^\infty e^{-h(c+c')x}\,\dd x=\frac1{h(c+c')}.
\end{equation}
With $u_{n,t}=K_{n,t}-1=\varsigma_t-E_{n,t}$, the product of the two $\varsigma_t$ terms is $O(n^{2\gamma-1})=o(n^\gamma)$. Cauchy--Schwarz bounds each mixed term by $O(n^{(3\gamma-1)/2})=o(n^\gamma)$. Hence
\[
n^{-\gamma}\sum_tu_{n,t}(c)u_{n,t}(c')\longrightarrow\frac1{h(c+c')}.
\]

The linear terms in $u_{n,t}$ cancel to smaller order through the exact mass identity
\begin{equation}\label{eq:crossmass}
\sum_{t=1}^nK_{n,t}(c)
=\sum_{k=1}^n\left\{1-\prod_{j=1}^k(1-a_j(c))\right\}=n+O(1).
\end{equation}
The first equality follows by interchanging sums and telescoping; the remaining product series is absolutely summable. Combining the last two displays yields
\begin{equation}\label{eq:crossgram}
\frac1n\sum_{t=1}^nK_{n,t}(c)K_{n,t}(c')
=1+\frac{n^{\gamma-1}}{h(c+c')}+o(n^{\gamma-1}).
\end{equation}

To incorporate dependence, extend $K_n$ by zero outside $1,\ldots,n$. Its backward recurrence is
\[
K_{n,t}=a_t+\frac{a_t}{a_{t+1}}(1-a_{t+1})K_{n,t+1}.
\]
Boundedness and $a_t/a_{t+1}=1+O(t^{-1})$ give
$|K_{n,t}-K_{n,t+1}|\le C(t^{-\gamma}+t^{-1})$. Since $2\gamma>1$ and the two boundary jumps are bounded, $\|\Delta K_n\|_{\ell_2}\le C$, while $\|K_n\|_{\ell_2}\le C\sqrt n$. For every integer $k$,
\[
\left|\sum_t K_{n,t}(c)K_{n,t-k}(c')-\sum_tK_{n,t}(c)K_{n,t}(c')\right|
\le C|k|\sqrt n.
\]
Absolute covariance summability with one lag moment now gives
\[
\Cov\{\mathcal L_n(c),\mathcal L_n(c')\}
=\frac{\Omega}{h^2n}\sum_tK_{n,t}(c)K_{n,t}(c')+O(n^{-1/2}).
\]
The remainder is $o(n^{\gamma-1})$, proving \eqref{eq:crossgain} under the initial positivity restriction.

For arbitrary fixed positive gains, choose a fixed index after which all $a_t<1/2$. The same bounds apply on that tail. Each earlier coefficient remains bounded because its product tail decays at a stretched-exponential rate. These finitely many terms contribute $O(n^{-1})$ to normalized inner products. The exact mass identity and the bounded difference norm also remain valid, including any early negative or zero product factors. This proves the general case. Finally, unrolling the linear recursion identifies its averaged stochastic term as $\mathcal L_n(c)/\sqrt n$; deterministic initialization contributes only to the mean.
\end{proof}

\subsection{Second-order variance and MSE of quantile extrapolation}
Finite-memory target scores satisfy the covariance condition in Proposition~\ref{prop:crossgain}. The quantitative remainder in Proposition~\ref{prop:bahadur} then transfers the expansion to the nonlinear recursion.

\begin{corollary}[Second-order extrapolation cost]\label{cor:rrcost}
Under Assumptions~\ref{as:memory}--\ref{as:marginal}, let $\Omega>0$, $4/5<\gamma<1$, and fix positive gains $c,c'$. Use deterministic initial values in $I$ and the same observations, offset $t_0$, and projection interval. Then
\begin{equation}\label{eq:rawcrossgain}
n\Cov\{\bar q_n(c),\bar q_n(c')\}
=\frac{\Omega}{h^2}+\frac{\Omega}{h^3(c+c')}n^{\gamma-1}+o(n^{\gamma-1}).
\end{equation}
For $\widehat q_n^{\rm RR}=2\bar q_n(c/2)-\bar q_n(c)$,
\begin{equation}\label{eq:rrcostratio}
\frac{\Var(\widehat q_n^{\rm RR})}{\Var(\bar q_n(c))}
=1+\frac4{3hc}n^{\gamma-1}+o(n^{\gamma-1}).
\end{equation}
Writing $\operatorname{MSE}(\widehat q)=\E(\widehat q-q^\star)^2$, we also have
\begin{equation}\label{eq:rrmsecost}
\operatorname{MSE}(\widehat q_n^{\rm RR})-\operatorname{MSE}(\bar q_n(c))
=\frac{4\Omega}{3h^3c}n^{\gamma-2}+o(n^{\gamma-2}).
\end{equation}
\end{corollary}

\begin{proof}
By Proposition~\ref{prop:bahadur}, $X_n(c):=\sqrt n\{\bar q_n(c)-q^\star\}=\mathcal L_n(c)+\mathcal R_n(c)$, where
\[
\|\mathcal R_n(c)\|_2\le C_c \Upsilon_n,\qquad
\Upsilon_n=n^{1/2-\gamma}+n^{-\gamma/4}+n^{-1/2}.
\]
The linear terms have bounded $L^2$ norms. Expanding covariances and applying Cauchy--Schwarz bounds the change from the linear covariance by $C(\Upsilon_n+\Upsilon_n^2)$. Relative to $n^{\gamma-1}$, the three remainder terms are
\[
n^{3/2-2\gamma},\qquad n^{1-5\gamma/4},\qquad n^{1/2-\gamma}.
\]
They all vanish for $\gamma>4/5$, proving \eqref{eq:rawcrossgain}.

The covariance expansion gives
\[
n\Var(\widehat q_n^{\rm RR})
=\frac\Omega{h^2}\left\{1+\frac{11}{6hc}n^{\gamma-1}+o(n^{\gamma-1})\right\},
\quad
n\Var(\bar q_n(c))
=\frac\Omega{h^2}\left\{1+\frac1{2hc}n^{\gamma-1}+o(n^{\gamma-1})\right\}.
\]
Taking the ratio proves \eqref{eq:rrcostratio}. Since $\E\mathcal L_n(c)=0$, $|\E X_n(c)|\le C_c\Upsilon_n$, and the same bound holds for the fixed RR combination. Set $a_n=n^{\gamma-1}$. Here $\Upsilon_n=o(a_n)$ and $a_n\to0$, so both squared means are $O(\Upsilon_n^2)=o(a_n)$. Their contribution to MSE is therefore $o(n^{\gamma-2})$. Subtracting the variance expansions proves \eqref{eq:rrmsecost}.
\end{proof}

The sufficient threshold $4/5$ comes from the available $L^2$ remainder. The explicit leading-mean cancellation in Corollary~\ref{cor:rr} additionally uses the conditional-density assumptions of Theorem~\ref{thm:bias}. Together, the results show how bias reduction and an eventual second-order MSE cost coexist for fixed-gain, full-trajectory averaging.

\begin{remark}[Fixed affine combinations]
For fixed gains $c_1,\ldots,c_m$ and fixed coefficients with $\sum_j\lambda_j=1$, the second-order coefficient of the linear average is
\[
n\Var\left(\sum_j\lambda_j\bar e_n(c_j)\right)
=\frac\Omega{h^2}\left\{1+\frac{n^{\gamma-1}}h
\sum_{j,k}\frac{\lambda_j\lambda_k}{c_j+c_k}+o(n^{\gamma-1})\right\}.
\]
The correction kernel is positive semidefinite because
$\sum_{j,k}\lambda_j\lambda_k/(c_j+c_k)=\int_0^\infty(\sum_j\lambda_je^{-c_jx})^2\,\dd x$.
For a fixed $s\in(0,1)$, the extrapolation $\{\bar e_n(sc)-s\bar e_n(c)\}/(1-s)$ has variance-ratio correction
$\{(2s+1)/(2hcs(1+s))\}n^{\gamma-1}$ relative to $\bar e_n(c)$. The same formulas transfer under Corollary~\ref{cor:rrcost}. Their asymptotics hold with the gains and combination coefficients fixed.
\end{remark}

For finite samples, let $\boldsymbol\Gamma_n=(\Gamma_{s-t})_{s,t=1}^n$ and $A_n(c)=(A_{nt}(h;c))_{t=1}^n$. The exact linear-filter variance ratio is
\begin{equation}\label{eq:rrexactfilter}
\frac{\{2A_n(c/2)-A_n(c)\}^{\top}\boldsymbol\Gamma_n\{2A_n(c/2)-A_n(c)\}}
{A_n(c)^{\top}\boldsymbol\Gamma_n A_n(c)}.
\end{equation}
Appendix~\ref{app:rrfinite} evaluates this quantity at $\gamma=.75$ and Appendix~\ref{app:rr085} at $\gamma=.85$, using the true density and target-score covariances.

\subsection{Delayed extrapolation under marginal assumptions}\label{app:delayedrr}
Corollary~\ref{cor:rr} uses the conditional-density Assumption~\ref{as:conditional} through Theorem~\ref{thm:bias}. A sufficient delay removes this requirement.

\begin{lemma}[Delayed Bahadur representation]\label{lem:delayedbahadur}
Under Assumptions~\ref{as:memory}--\ref{as:marginal}, for fixed $d\ge r$ and fixed gain $c>0$,
\[
\sqrt n(\bar q_n^{(d)}-q^\star)=-\frac1{h\sqrt n}\sum_{t=1}^n\psi_t+o_{L^2}(1).
\]
\end{lemma}
\begin{proof}
Write $e_t=q_t^{(d)}-q^\star$, $H_t(q)=\ind(X_t\le q)-F(q)$ and let $P_t$ be the projection correction. The recursion gives
\[
he_{t-d}=\frac{e_{t-1}-e_t}{\eta_t}+\frac{P_t}{\eta_t}-\psi_t-\Lambda_t-\Xi_t,
\]
with $\Lambda_t=F(q^{(d)}_{t-d})-\tau-he_{t-d}$ and $\Xi_t=H_t(q^{(d)}_{t-d})-H_t(q^\star)$. Sum over $t\le n$ and divide by $\sqrt n$. Abel summation gives $\sum_t(e_{t-1}-e_t)/\eta_t=e_0/\eta_1-e_n/\eta_n+\sum_{t<n}e_t(\eta_{t+1}^{-1}-\eta_t^{-1})$; the moment bounds in the proof of Theorem~\ref{thm:delay} (Appendix~\ref{app:delay}) give $\|e_t\|_{2p}\le C_p\eta_t^{1/2}$ for every fixed $p$ and $|\eta_{t+1}^{-1}-\eta_t^{-1}|\le Ct^{\gamma-1}$, so this term is $O(n^{\gamma/2})$ in $L^2$ and $o(\sqrt n)$. Next, $\|\Lambda_t\|_2\le C\eta_t$ by the Lipschitz density and the fourth-moment bound, so $n^{-1/2}\sum_t\|\Lambda_t\|_2=O(n^{1/2-\gamma})$. Because $d\ge r$, $X_t$ is independent of $\mathcal F_{t-d}$ and $\E(\Xi_t\mid\mathcal F_{t-d})=0$; within each residue class modulo $d$, $(\Xi_t)$ is a martingale difference sequence with $\E\Xi_t^2\le C\E|e_{t-d}|\le C\eta_t^{1/2}$, so $\|\sum_t\Xi_t\|_2=O(n^{(1-\gamma/2)/2})=o(\sqrt n)$. Projection corrections have arbitrarily high polynomial order (Appendix~\ref{app:delay}). Finally $n^{-1/2}\sum_{t\le n}e_{t-d}$ differs from $\sqrt n\bar e_n$ by $O(d/\sqrt n)$ in $L^2$. The remainder is therefore $O(n^{(\gamma-1)/2}+n^{1/2-\gamma}+n^{-\gamma/4})$ in $L^2$.
\end{proof}

\begin{corollary}[Delayed shared-stream extrapolation]\label{cor:delayedrr}
Under Assumptions~\ref{as:memory}--\ref{as:marginal}, for fixed $d\ge r$, let $\widehat q^{\,{\rm RR},(d)}_n=2\bar q^{(d)}_n(c/2)-\bar q^{(d)}_n(c)$ on common data. Then $\E(\widehat q^{\,{\rm RR},(d)}_n-q^\star)=o(n^{-\gamma})$, $\sqrt n(\widehat q^{\,{\rm RR},(d)}_n-q^\star)\Rightarrow N(0,\Omega/h^2)$ and $n\Var(\widehat q^{\,{\rm RR},(d)}_n)\to\Omega/h^2$.
\end{corollary}
\begin{proof}
The coefficient $B_\tau^{\rm del}=-f'(q^\star)\Omega/(4h^2)$ of Theorem~\ref{thm:delay} does not depend on the gain, and $\bar\eta_n(c/2)=\bar\eta_n(c)/2$, so the leading mean terms cancel. By Lemma~\ref{lem:delayedbahadur} both gains share the leading linear term, whose coefficient in the combination is $2-1=1$; the finite-dependence central limit theorem and the $L^2$ remainder give the distributional and variance limits.
\end{proof}
The lemma's remainder is too coarse for second-order variance comparisons, but empirically delay does not reduce the variance cost of extrapolation: in Table~\ref{tab:biasscale}, ordinary and delayed extrapolation raise the variance by $19$--$59\%$ and $20$--$59\%$, while the delayed recursion alone has variance ratio $1.00$--$1.03$ to ordinary SGD. Both extrapolations leave the gain-free residual described in Appendix~\ref{app:finite}.

\subsection{Finite-sample cost of shared-stream extrapolation}\label{app:rrfinite}
The nine Cauchy bias settings use 4,000 independent streams per setting, paired across gains. Table~\ref{tab:rrfinite} compares the ordinary average with $2\bar q_n(c/2)-\bar q_n(c)$. Write $v_\infty=\Omega/h^2$ for their common first-order variance coefficient. Sample variances use denominator $R-1$; MSE averages squared errors from the true quantile. Ratio MCSEs use the paired delta method across streams. For a ratio $A/B$, the influence value is $(a_i-(A/B)b_i)/B$, using centered squared deviations for variance and squared errors for MSE.

\begin{table}[H]
\centering\small
\caption{Shared-stream Richardson--Romberg extrapolation on 4,000 streams per setting. Empirical variances are normalized by $v_\infty/n$, where $v_\infty=\Omega/h^2$. All ratios compare RR with ordinary averaged SGD. The exact linear ratio uses \eqref{eq:rrexactfilter}; parentheses on empirical ratios report paired MCSEs.}\label{tab:rrfinite}
{\setlength{\tabcolsep}{4pt}
\begin{tabular}{rrrrrrr}
\toprule
$r$ & $n$ & \multicolumn{2}{c}{Normalized variance} & \multicolumn{2}{c}{Variance ratio} & MSE ratio\\
& & Ordinary & RR & Exact linear & Empirical & Empirical\\
\midrule
1 & 3k & 1.204 & 1.845 & 1.480 & 1.532 (0.029) & 1.500 (0.029)\\
1 & 12k & 1.122 & 1.723 & 1.453 & 1.535 (0.024) & 1.509 (0.024)\\
1 & 48k & 1.115 & 1.464 & 1.302 & 1.314 (0.017) & 1.302 (0.017)\\
2 & 3k & 1.199 & 1.822 & 1.478 & 1.519 (0.029) & 1.474 (0.029)\\
2 & 12k & 1.129 & 1.690 & 1.452 & 1.498 (0.024) & 1.473 (0.024)\\
2 & 48k & 1.126 & 1.491 & 1.302 & 1.324 (0.017) & 1.313 (0.017)\\
4 & 3k & 1.213 & 1.807 & 1.475 & 1.490 (0.027) & 1.441 (0.026)\\
4 & 12k & 1.070 & 1.658 & 1.451 & 1.550 (0.024) & 1.521 (0.024)\\
4 & 48k & 1.116 & 1.462 & 1.302 & 1.310 (0.016) & 1.299 (0.016)\\
\bottomrule
\end{tabular}}
\end{table}

For the exact linear filter, we use $h=1/(2\pi)$, $c=3$, $t_0=20$, $\gamma=.75$, and the exact finite-memory target-score covariance in \eqref{eq:rrexactfilter}. At $n=48{,}000$, the exact variance ratios are 1.3021, 1.3019 and 1.3016 for $r=1,2,4$. The nonlinear estimates are 1.314, 1.324 and 1.310. At $n=12{,}000$, the empirical ratios exceed the linear values by .045--.099.

The leading expansion $1+4n^{\gamma-1}/(3hc)$ gives 1.1887 at $n=48{,}000$, below the exact filter ratio. The terminal-layer asymptotics therefore converge slowly at these gains and sample sizes. Exact finite filters account for much of the observed variance increase. Corollary~\ref{cor:rrcost} establishes the nonlinear second-order coefficient for $\gamma>4/5$, and Appendix~\ref{app:rr085} repeats the experiment at $\gamma=.85$.

\subsection{Extrapolation at \texorpdfstring{$\gamma=.85$}{gamma = .85}}\label{app:rr085}
Corollary~\ref{cor:rrcost} requires $\gamma>4/5$, whereas the main experiments use $\gamma=.75$. A separate prespecified study repeats the Cauchy extrapolation experiment with $\gamma=.85$, gains 3 and 1.5, $t_0=20$, $q_0=q^\star$, memories $r=1,2,4$, $n\in\{3{,}000,12{,}000,48{,}000,192{,}000\}$ and 4,000 streams per cell (Table~\ref{tab:rr085}). For $n\ge12{,}000$ the empirical variance ratio lies within 2.9 MCSE of the exact linear-filter ratio \eqref{eq:rrexactfilter}. The second-order expansion $1+4n^{\gamma-1}/(3hc)$ has the right form but, for $n\ge12{,}000$, is far below both, because $n^{\gamma-1}$ decays slowly and higher-order terms of the terminal layer are not negligible at these sample sizes. At this $\gamma$ the leading bias $B_\tau\bar\eta_n$ is of order $n^{-.85}$ and is not larger than the gain-free residual, so extrapolation reduces the absolute bias at $n\le12{,}000$ but not at larger $n$, apart from a tie within MCSE at $r=1$, $n=192{,}000$, while more than doubling the variance from $n=12{,}000$ on. The second-order theory thus describes an eventual regime; the exact finite filter predicts the variance cost at practical sample sizes.

\begin{table}[H]
\centering\footnotesize
\caption{Shared-stream extrapolation at $\gamma=.85$ (4,000 streams per row). Normalized variances divide $n\Var$ by $\Omega/h^2$. Ratios compare RR with ordinary averaged SGD; parentheses give paired delta-method MCSEs. Biases $\times10^3$.}\label{tab:rr085}
{\setlength{\tabcolsep}{3pt}
\begin{tabular}{rrrrrrrrrr}
\toprule
 & & \multicolumn{2}{c}{Normalized variance} & \multicolumn{4}{c}{Variance ratio and MSE ratio} & \multicolumn{2}{c}{Bias}\\
$r$ & $n$ & Ordinary & RR & Exact linear & Expansion & Empirical & MSE ratio & Ordinary & RR\\
\midrule
1 & 3k & 1.490 & 2.666 & 1.651 & 1.840 & 1.789 (0.037) & 1.755 (0.037) & -8.850 (0.959) & -3.341 (1.282)\\
1 & 12k & 1.423 & 3.611 & 2.469 & 1.683 & 2.539 (0.054) & 2.510 (0.054) & -3.706 (0.468) & -3.116 (0.746)\\
1 & 48k & 1.355 & 3.721 & 2.660 & 1.554 & 2.747 (0.058) & 2.740 (0.058) & -1.122 (0.229) & -1.410 (0.379)\\
1 & 192k & 1.231 & 2.865 & 2.320 & 1.450 & 2.326 (0.046) & 2.325 (0.047) & -0.246 (0.109) & -0.239 (0.166)\\
2 & 3k & 1.500 & 2.495 & 1.646 & 1.840 & 1.663 (0.034) & 1.632 (0.034) & -11.039 (1.223) & -3.320 (1.577)\\
2 & 12k & 1.404 & 3.499 & 2.464 & 1.683 & 2.492 (0.053) & 2.470 (0.053) & -4.254 (0.592) & -3.772 (0.934)\\
2 & 48k & 1.305 & 3.541 & 2.657 & 1.554 & 2.713 (0.056) & 2.711 (0.056) & -1.493 (0.285) & -2.289 (0.470)\\
2 & 192k & 1.220 & 2.903 & 2.318 & 1.450 & 2.380 (0.047) & 2.386 (0.047) & -0.327 (0.138) & -0.872 (0.213)\\
4 & 3k & 1.558 & 2.505 & 1.637 & 1.840 & 1.608 (0.029) & 1.575 (0.030) & -15.383 (1.679) & -1.900 (2.130)\\
4 & 12k & 1.441 & 3.605 & 2.456 & 1.683 & 2.503 (0.052) & 2.470 (0.052) & -6.361 (0.807) & -3.964 (1.277)\\
4 & 48k & 1.366 & 3.678 & 2.651 & 1.554 & 2.692 (0.058) & 2.680 (0.058) & -2.557 (0.393) & -3.190 (0.645)\\
4 & 192k & 1.213 & 2.980 & 2.316 & 1.450 & 2.456 (0.048) & 2.459 (0.049) & -0.868 (0.185) & -1.474 (0.290)\\
\bottomrule
\end{tabular}}
\end{table}

\section{Geometry and computation of the inversion}\label{app:inversion}
This appendix establishes the deterministic properties of the acceptance set: the score-space Lipschitz bound and the hull sandwich of Proposition~\ref{prop:hullsandwich}, connectedness under a larger floor, the exact scan of Proposition~\ref{prop:fenwick}, and the finite-filter weights used by S-hull and Filter-hull.

\subsection{The connected-floor interval}
For comparison with the hull procedure, fix $\lambda\ge1/2$ and define
\begin{equation}\label{eq:floor}
v_{\rm conn,n}=(\lambda z(4b+1)w_*)^2,\qquad
C_{n,\alpha}=\overline{\{q:|S_n(q)|\le z\sqrt{\widehat V_b(q)\vee v_{\rm conn,n}}\}}.
\end{equation}
\begin{proposition}[Connected-floor geometry]\label{prop:scan}
For every data set the acceptance set in \eqref{eq:floor} is an interval, possibly empty or unbounded. It is computed by the scan of Proposition~\ref{prop:fenwick} with the floor $v_{\rm conn,n}$ and is equivariant under increasing transformations.
\end{proposition}
\begin{theorem}[Connected-floor coverage]\label{thm:coverage}
Under Assumptions~\ref{as:memory} and \ref{as:score}, with $b\ge r-1$, if $v_{\rm conn,n}\le V_n/2$ and $\epsilon_n(u)\le V_n/2$, then
\begin{equation}\label{eq:coveragerate}
\left|\Pp(q^\star\in C_{n,\alpha})-(1-\alpha)\right|
\le2\beta_n+C_\alpha \epsilon_n(u)/V_n+2e^{-u}.
\end{equation}
The quantities $\beta_n$ and $\epsilon_n(u)$ are defined in \eqref{eq:be}--\eqref{eq:dn}.
\end{theorem}
Theorem~\ref{thm:coverage} is proved in Appendix~\ref{app:sharp}. Compared with the hull, the larger floor requires $v_{\rm conn,n}\le V_n/2$ and can be active in finite samples even when $\widehat V_b(q^\star)$ is accurate.
\subsection{A global Lipschitz bound in score space}\label{app:lipschitz}
Write $Q(q)=\widehat V_b(q)$ and $S(q)=S_n(q)$. Flipping the indicator at index $i$ changes the quadratic form by
\[
\Delta Q=w_i^2(1-2\tau)+2w_i\sum_{0<|j-i|\le b}w_js_j,
\qquad |\Delta Q|\le(4b+1)w_*w_i.
\]
Summing this bound over the indicators that change between two thresholds proves
\begin{equation}\label{eq:globalqlip}
|Q(q')-Q(q)|\le\ell_{n,b}\{S(q')-S(q)\},\quad q\le q',\qquad
\ell_{n,b}=(4b+1)w_*.
\end{equation}
This is deterministic, holds with ties, and does not require a positive semidefinite covariance kernel.

\begin{proof}[Proof of Proposition~\ref{prop:hullsandwich}]
Fix a value $q^\dagger$ that is not an observed value, and suppose $Q_0=Q(q^\dagger)>v_{\min,n}$. Raw acceptance at $q^\dagger$ implies membership in its closed hull. Conversely, suppose $q^\dagger$ belongs to the hull. If $S_0=S(q^\dagger)\ge0$, there is an accepted threshold $q\ge q^\dagger$; a limiting endpoint is handled by the constant-score interval that contains $q^\dagger$. Set $s=S(q)\ge S_0$. If acceptance uses the floor, then $s\le z\sqrt{v_{\min,n}}\le z\sqrt{Q_0}$. Otherwise \eqref{eq:globalqlip} implies
\[
s^2\le z^2 Q(q)\le z^2\{Q_0+\ell_{n,b}(s-S_0)\}
\le z^2(Q_0+\ell_{n,b}s).
\]
The positive root of this quadratic is at most $z\sqrt{Q_0}+z^2\ell_{n,b}$. Therefore $S_0$ satisfies the same bound. If $S_0<0$, choose an accepted $q\le q^\dagger$, put $s=-S(q)\ge -S_0$, and use the same argument. This proves the sandwich. If the acceptance set is empty, its hull is defined to be empty and the assertions remain valid.

Algorithm~\ref{alg:mainhull} records the smallest and largest accepted endpoints across every component, at the cost given in Proposition~\ref{prop:fenwick}. Increasing continuous transformations preserve threshold order and every chronological lag product, so all raw components and their hulls transform equivariantly.
\end{proof}

\subsection{Proof of connectedness and equivariance}
The map $x\mapsto\sqrt{x\vee v_{\rm conn,n}}$ is $(2\sqrt{v_{\rm conn,n}})^{-1}$-Lipschitz. As a threshold passes $X_i$, the single-flip bound $|\Delta Q|\le\ell_{n,b}w_{ni}$ of Appendix~\ref{app:lipschitz} and $\lambda\ge1/2$ imply
\[
z|\Delta\sqrt{\widehat V_b\vee v_{\rm conn,n}}|
\le\frac{z\ell_{n,b}w_{ni}}{2\sqrt{v_{\rm conn,n}}}
=\frac{w_{ni}}{2\lambda}\le\Delta S_n.
\]
Both $S_n-z\sqrt{\widehat V_b\vee v_{\rm conn,n}}$ and $S_n+z\sqrt{\widehat V_b\vee v_{\rm conn,n}}$ are nondecreasing. The first inequality in the acceptance condition specifies a lower ray and the second an upper ray; their intersection is an interval. Ties are handled by summing the single-flip inequalities. Continuous strictly increasing bijections of $\R$ preserve every indicator and every chronological lag pair, and map order-statistic endpoints to transformed endpoints. This proves Proposition~\ref{prop:scan}.

For a tied group $\mathcal I=\{i:X_i=x\}$, let $s_j$ denote the scores just below $x$. Updating the entire group changes the quadratic form by
\begin{equation}\label{eq:batchflip}
\begin{split}
\Delta Q={}&(1-2\tau)\sum_{i\in \mathcal I}w_i^2
+2\sum_{i\in \mathcal I}\sum_{\substack{j\notin \mathcal I\\0<|j-i|\le b}}w_iw_js_j\\
&+2(1-2\tau)\sum_{\substack{i<j,\ i,j\in \mathcal I\\j-i\le b}}w_iw_j.
\end{split}
\end{equation}
For a pair within $\mathcal I$, both scores change from $-\tau$ to $1-\tau$, giving the last term. Sequential single-index updates sum to this identity in any within-group order. Acceptance is evaluated only after the whole group has changed. Zero-weight indices leave both sums unchanged. Testing all constant-score intervals and closing their union supplies the endpoint convention in Algorithm~\ref{alg:mainhull}.

Algorithm~\ref{alg:mainhull} with the floor $v_{\rm conn,n}$ computes $C_{n,\alpha}$. With a direct neighbor loop the initialization costs $O(nb)$ and each update $O(b)$; Appendix~\ref{app:fenwick} reduces the total to $O(n\log n)$. Scores and time indices require $O(n)$ storage. For a fixed grid of candidate values and a known terminal $n$, the same score and quadratic sums can instead be accumulated as observations arrive: each grid point stores a scalar score, a scalar diagonal sum, its lag-product sum, and a buffer of the previous $b$ scores. This streaming construction determines acceptance at the specified grid points.

\subsection{Fenwick-tree implementation and verification}\label{app:fenwick}
A Fenwick tree over $a_1,\ldots,a_n$ stores partial sums over dyadic index blocks, so a prefix sum $\sum_{j\le i}a_j$ and a point update $a_i\leftarrow a_i+x$ each take $O(\log n)$ operations; the tree is built in $O(n)$ \citep{fenwick1994}. In Algorithm~\ref{alg:mainhull}, $a_j=w_js_j$. The neighbor sum $\sum_{0<|j-i|\le b}w_js_j$ is a difference of two prefix sums minus $a_i$, and flipping index $i$ adds $w_i$ to $a_i$. The initial quadratic form uses $\sum_{k=1}^b\sum_{t>k}w_tw_{t-k}=\sum_tw_t(A_{t-1}-A_{t-b-1})$ with $A_t=\sum_{j\le t}w_j$, an $O(n)$ computation. Each index flips once, so after an $O(n\log n)$ sort the scan costs $O(n\log n)$ for every $b$. Tied groups are fully updated before their interval is tested, so \eqref{eq:batchflip} holds; the full-window rule $Q=S^2$, the floor, and the reporting of empty or unbounded hulls are unchanged. The connected-floor interval uses the same scan with the floor $v_{\rm conn,n}$.

We compared the Fenwick and direct scans on 3,000 random configurations with $2\le n<120$: 1,191 with ties, 943 with zero weights and 1,529 with non-uniform weights, $\tau\in\{.1,.25,.5,.75,.9\}$, three critical values, windows $0\le b<n$ and both floors. Lower and upper endpoints, the weighted-quantile center and the number of acceptance components agreed exactly in every case, including 95 empty and 2,382 unbounded outputs and 72 multi-component acceptance sets. Table~\ref{tab:fenwicktiming} reports single-thread times on Gaussian data with equal weights. At the default power windows the two scans take similar time; at $b=128$ and $512$ the Fenwick scan is $2.3$--$3.4$ and $5.8$--$10.9$ times faster, and its time is nearly constant in $b$.

\begin{table}[H]
\centering\small
\caption{Single-thread scan times in seconds (median of five runs; three at $n=10^6$), equal weights, $\tau=.25$. Both scans return identical intervals.}\label{tab:fenwicktiming}
\begin{tabular}{rrrrr}
\toprule
$n$ & $b$ & Direct scan & Fenwick scan & Ratio\\
\midrule
3,000 & 8 & 0.000163 & 0.000190 & 0.86\\
3,000 & 32 & 0.000295 & 0.000217 & 1.36\\
3,000 & 128 & 0.000755 & 0.000225 & 3.36\\
3,000 & 512 & 0.002379 & 0.000218 & 10.89\\
12,000 & 11 & 0.001108 & 0.000998 & 1.11\\
12,000 & 32 & 0.001281 & 0.001026 & 1.25\\
12,000 & 128 & 0.003198 & 0.001031 & 3.10\\
12,000 & 512 & 0.010388 & 0.001092 & 9.51\\
100,000 & 18 & 0.010631 & 0.010296 & 1.03\\
100,000 & 32 & 0.012959 & 0.010375 & 1.25\\
100,000 & 128 & 0.028227 & 0.010661 & 2.65\\
100,000 & 512 & 0.088507 & 0.010363 & 8.54\\
1,000,000 & 32 & 0.192513 & 0.158894 & 1.21\\
1,000,000 & 128 & 0.381187 & 0.163730 & 2.33\\
1,000,000 & 512 & 1.019840 & 0.174393 & 5.85\\
\bottomrule
\end{tabular}
\end{table}

\subsection{Finite-filter weights}\label{app:hull}
Equal weights give the simplest construction. To retain the finite-filter connection to averaged SGD, use the following deterministic coefficients.
For a positive nominal slope $u_0$ satisfying $\eta_1u_0<1$, define
\begin{equation}\label{eq:filter}
\begin{gathered}
A_{nt}(u_0)=\frac{\eta_t}{\sqrt n}
\sum_{k=t}^n\prod_{j=t+1}^k(1-\eta_j u_0),\\
w_{nt}=\frac{A_{nt}(u_0)}{\norm{A_n(u_0)}_2}.
\end{gathered}
\end{equation}
A backward recurrence evaluates all weights in $O(n)$ operations. These are the exact finite-sample coefficients of a full-average linear recursion. They satisfy the weight conditions of Assumption~\ref{as:score} when $u_0$ is fixed in a compact positive range; see Lemma~\ref{lem:filters}. The coverage theorem does not require $u_0=h$. Equal weights $w_{nt}=n^{-1/2}$ are another admissible choice.

For $p_t=A_{nt}(u_0)/\sum_sA_{ns}(u_0)$, the reversal symmetrization is
\begin{equation}\label{eq:symweights}
p_t^{\rm sym}=\tfrac12(p_t+p_{n+1-t}),\qquad
w_t^{\rm sym}=p_t^{\rm sym}/\|p^{\rm sym}\|_2.
\end{equation}
The symmetric weights make the interval exactly invariant to reversing the observations. At a common sum normalization they also weakly reduce the linearized variance relative to the original filter (Proposition~\ref{prop:symmetry}). This comparison does not order either construction against equal weights.

\subsection{Reversal projection}
\begin{proposition}[Variance and reversal]\label{prop:symmetry}
Let $\Gamma_n$ be the covariance matrix of the stationary scalar target scores, $\mathbf R$ reverse the coordinates, $p^{\rm sym}=(p+\mathbf Rp)/2$, and $a=(p-\mathbf Rp)/2$. Then
\begin{equation}\label{eq:symvariance}
p^\top\Gamma_np-(p^{\rm sym})^\top\Gamma_np^{\rm sym}=a^\top\Gamma_na\ge0.
\end{equation}
Both filters sum to one. With symmetric weights, the score-inversion interval is exactly unchanged when the observations are reversed. These weights satisfy the same diffuse-weight and first-order-equivalence properties as the original finite filter.
\end{proposition}
The identity compares linearized quantile variances at the same population derivative and sum normalization. Interval performance, including equal-weight inversion, is evaluated in Section~\ref{sec:experiments}.

\begin{proof}[Proof of Proposition~\ref{prop:symmetry}]
Let $\mathbf R$ be the matrix that reverses coordinates and let $\Gamma_n=(\Gamma_{|i-j|})_{i,j\le n}$. Stationarity of a real scalar score implies $\mathbf R\Gamma_n\mathbf R=\Gamma_n$. Put $p^{\rm sym}=(p+\mathbf Rp)/2$ and $a=(p-\mathbf Rp)/2$. Then $\mathbf Rp^{\rm sym}=p^{\rm sym}$, $\mathbf Ra=-a$, and
$(p^{\rm sym})^\top\Gamma_na=(p^{\rm sym})^\top \mathbf R\Gamma_n\mathbf Ra=-(p^{\rm sym})^\top\Gamma_na=0$.
Expanding $p=p^{\rm sym}+a$ gives the variance identity in the proposition, and positive semidefiniteness gives its sign. The normalization here is $\sum_t p_t=\sum_t p_t^{\rm sym}=1$, so the population derivative of the quantile score is the same for both filters.

For $p_i^{\rm sym}=p_{n+1-i}^{\rm sym}$, reversing the observations leaves $\sum_i p_i^{\rm sym}\ind(X_i\le q)$ and each lag sum $\sum_{i>k}p_i^{\rm sym}p_{i-k}^{\rm sym}s_i(q)s_{i-k}(q)$ unchanged. The Euclidean normalization is also unchanged. The floor and all acceptance components therefore agree; so do their closed hulls. Equal values can be updated together. The norm-contraction of $(I+\mathbf R)/2$ preserves \eqref{eq:filterl2}. Thus the same weighted-quantile Bahadur argument proves first-order equivalence to the raw averaged SGD.
\end{proof}

\section{Coverage of the acceptance hull}\label{app:sharp}
We record the two probabilistic tools, then prove the coverage bounds for the connected floor and the hull, the data-selected-window corollary, local exactness, Theorem~\ref{cor:sharpcoverage}, and the lattice lower bound. Throughout, $\bar\Delta=2r-1$, $L_b=4b+1$, so that $\ell_{n,b}=L_bw_*$, and for $u>0$
\begin{align}
\beta_n={}&\frac{\bar\Delta w_*}{\sqrt{2\pi V_n}}
+\frac{16\sqrt n\bar\Delta^{3/2}w_*^2}{V_n}
+\frac{10n\bar\Delta^2w_*^3}{V_n^{3/2}},\label{eq:be}\\
\epsilon_n(u)={}&rL_bw_*^2\sqrt{(n+r-1)u/2}.\label{eq:dn}
\end{align}

\subsection{The bounded dependency-graph inequality}
We use Theorem 2.2 of \citet{rinott1994}. If $Y_1,\ldots,Y_n$ have a dependency graph whose maximum degree is strictly smaller than $\bar\Delta$, $|Y_i-\E Y_i|\le \bar M$, and $\sigma^2=\Var(\sum_iY_i)>0$, then
\begin{equation}\label{eq:rinott}
\sup_x\left|\Pp\left\{\frac{\sum_i(Y_i-\E Y_i)}\sigma\le x\right\}-\Phi(x)\right|
\le \frac{\bar\Delta\bar M}{\sqrt{2\pi}\sigma}+\frac{16\sqrt n\bar\Delta^{3/2}\bar M^2}{\sigma^2}
+\frac{10n\bar\Delta^2\bar M^3}{\sigma^3}.
\end{equation}
For $Y_t=w_{nt}\psi_t$ under Assumption~\ref{as:memory}, edges at lags $1,\ldots,r-1$ give degree at most $2r-2$, so $\bar\Delta=2r-1$, $\bar M=w_*$ and $\sigma^2=V_n$. This proves the $\beta_n$ term, including the i.i.d. case $r=1$.

\subsection{Variance concentration from innovation replacement}\label{app:concentration}
At the true threshold, $\E s_t(q^\star)=0$ and $\E s_t(q^\star)s_{t-k}(q^\star)=0$ for $k\ge r$. Hence $\E\widehat V_b(q^\star)=V_n$ whenever $b\ge r-1$.

Changing one indicator $s_i$ by $\pm1$ changes its diagonal square by $|1-2\tau|\le1$, and its at most $2b$ incident off-diagonal terms by at most $2w_{ni}w_{nj}$. Therefore
\begin{equation}\label{eq:singlelip}
|\Delta\widehat V_b|\le w_{ni}^2+2w_{ni}\sum_{0<|j-i|\le b}w_{nj}
\le L_bw_*w_{ni}\le L_bw_*^2.
\end{equation}
The indicator vector is a function of the $n+r-1$ independent innovations indexed from $2-r$ to $n$. Replacing one innovation affects at most $r$ observations. A telescoping replacement of their indicators and \eqref{eq:singlelip} bound its effect by $rL_bw_*^2$. The bounded-differences inequality \citep{mcdiarmid1989} now gives
\[
\Pp(|\widehat V_b(q^\star)-\E\widehat V_b(q^\star)|>t)
\le2\exp\left\{-\frac{2t^2}{(n+r-1)r^2L_b^2w_*^4}\right\}
\]
for every $b<n$. Taking $t=\epsilon_n(u)$ gives failure probability $2e^{-u}$, and the centering equals $V_n$ when $b\ge r-1$.

\subsection{Coverage of the connected-floor interval}
\begin{proof}[Proof of Theorem~\ref{thm:coverage}]
Let $\mathcal G=\{|\widehat V_b(q^\star)-V_n|\le \epsilon_n(u)\}$ and $\delta=\epsilon_n(u)/V_n\le1/2$. On $\mathcal G$, the assumed floor bound ensures that the floor has no effect, and
\[
z\sqrt{V_n}\sqrt{1-\delta}\le z\sqrt{\widehat V_b(q^\star)\vee v_{\rm conn,n}}
\le z\sqrt{V_n}\sqrt{1+\delta}.
\]
Thus deterministic normal intervals bracket the acceptance event, up to $\mathcal G^c$. Applying \eqref{eq:rinott} to both ends of each interval gives an error at most $2\beta_n$. For $|a|\le1/2$,
$|2\Phi(z\sqrt{1+a})-2\Phi(z)|\le C_\alpha |a|$, for example with $C_\alpha=2z$. Add $\Pp(\mathcal G^c)\le2e^{-u}$ to obtain \eqref{eq:coveragerate}.
Finally, $q^\star$ equals none of the finitely many observed values with probability one. Closing the acceptance set therefore does not change its membership event at $q^\star$.
\end{proof}

For fixed $r$, $b\ge r-1$, and $w_*\le C_w/\sqrt n$, $\beta_n=O_r(n^{-1/2})$, $\epsilon_n(\log n)=O_{r,b}(\sqrt{\log n/n})$, and $v_{\rm conn,n}=O_b(n^{-1})$. For $b_n=\lceil\sqrt{\log n}\rceil$, the same formulas yield $\epsilon_n(\log n)=O_r(\log n/\sqrt n)$ and $v_{\rm conn,n}=O(\log n/n)$. Hence \eqref{eq:coveragerate} is $O_r(\log n/\sqrt n)$ for this window once $b_n\ge r-1$.

\subsection{Coverage with an explicit covariance-tail term}
\begin{theorem}[Quantitative hull coverage]\label{thm:hullcoverage}
Under Assumptions~\ref{as:memory} and \ref{as:score}, put $\Delta_n=\epsilon_n(u)+|T_{n,b}|$. If $\Delta_n\le V_n/2$ and $1/n<V_n/2$, then
\begin{equation}\label{eq:hullcoverage}
\begin{split}
\big|\Pp(q^\star\in C^{\rm H}_{n,\alpha})-(1-\alpha)\big|
\le{}&2\beta_n+2e^{-u}\\
&+C_\alpha\left\{\frac{\Delta_n}{V_n}
+\frac{\ell_{n,b}}{\sqrt{V_n}}\right\}.
\end{split}
\end{equation}
No moment or density assumption on the observations is needed beyond Assumptions~\ref{as:memory} and \ref{as:score}.
\end{theorem}

\begin{proof}[Proof of Theorem~\ref{thm:hullcoverage}]
For any $b<n$, stationarity gives the exact identity
\[
\E Q(q^\star)=V_n-T_{n,b},\qquad
T_{n,b}=2\sum_{k=b+1}^{r-1}\Gamma_k\sum_{t=k+1}^nw_tw_{t-k}.
\]
An empty sum is zero. The bounded-differences bound of Appendix~\ref{app:concentration} gives $\Pp\{|Q(q^\star)-\E Q(q^\star)|>\epsilon_n(u)\}\le2e^{-u}$.
Let $\Delta_n=\epsilon_n(u)+|T_{n,b}|$ and $\mathcal G=\{|Q(q^\star)-V_n|\le\Delta_n\}$. Under the theorem's conditions, on $\mathcal G$ we have $Q(q^\star)\ge V_n/2>v_{\min,n}$. Almost surely $q^\star$ is not an observation. Proposition~\ref{prop:hullsandwich} then yields the deterministic event bounds
\begin{align*}
\{|S(q^\star)|\le z\sqrt{V_n-\Delta_n}\}\cap \mathcal G
&\subseteq\{q^\star\in C^{\rm H}_{n,\alpha}\},\\
\{q^\star\in C^{\rm H}_{n,\alpha}\}\cap \mathcal G
&\subseteq\{|S(q^\star)|\le z\sqrt{V_n+\Delta_n}+z^2\ell_{n,b}\}.
\end{align*}
The dependency-graph bound \eqref{eq:be} controls either deterministic interval probability with error at most $2\beta_n$. Normal probabilities change by at most $C_\alpha\Delta_n/V_n$ under the scale perturbation and by at most $C_\alpha\ell_{n,b}/\sqrt{V_n}$ under the additive threshold perturbation. Finally add $\Pp(\mathcal G^c)\le2e^{-u}$. This proves the stated two-sided coverage bound from normal approximation at the target score.

For fixed $r$ and $b\ge r-1$, $T_{n,b}=0$. With $w_*=O(n^{-1/2})$, $\ell_{n,b}=O((b+1)/\sqrt n)$ and $\epsilon_n(\log n)=O_r((b+1)\sqrt{\log n/n})$. Substitution gives the concentration-based bounds for fixed, logarithmic and power windows. The sharper fixed-window rate is proved with Theorem~\ref{cor:sharpcoverage} below. For windows below the true memory, the displayed tail remains in the bound.
\end{proof}

\subsection{Infinite-memory transfer for the hull}
Under the same-marginal indicator coupling in Appendix~\ref{app:geometric}, equality of all $n$ true-quantile indicators makes $S(q^\star)$ and $Q(q^\star)$ agree. The preceding proof uses these quantities and the deterministic score-space sandwich, which holds for either data set. Apply the same lower and upper deterministic event bounds on the coupling event and add its failure probability $nC\rho^r$. This gives the rate stated in Appendix~\ref{app:geometric}.

\subsection{Data-selected windows}\label{app:datawindow}
\begin{proof}[Proof of Corollary~\ref{cor:datawindow}]
Let $\mathcal W=\{r_\psi-1,\ldots,b_{\max}\}$, where $b_{\max}$ may depend on $n$. For each $b\in\mathcal W$, $T_{n,b}=0$ because $\Gamma_k=0$ for $k\ge r_\psi$, so $\E\widehat V_b(q^\star)=V_n$. Replacing one innovation changes $\widehat V_b(q^\star)$ by at most $r(4b+1)w_*^2\le r(4b_{\max}+1)w_*^2$. McDiarmid's inequality with $u=\log n$ and a union bound over the at most $b_{\max}+1$ windows in $\mathcal W$ give $\Pp(\mathcal G^c)\le2(b_{\max}+1)/n$ for
\[
\mathcal G=\bigcap_{b\in\mathcal W}\{|\widehat V_b(q^\star)-V_n|\le\epsilon^{\max}_n\},\qquad
\epsilon^{\max}_n=r(4b_{\max}+1)w_*^2\sqrt{(n+r-1)\log n/2}.
\]
Suppose $\epsilon^{\max}_n\le V_n/2$ and $1/n<V_n/2$; otherwise the asserted bound exceeds one once $C$ is large, because $\epsilon^{\max}_n\le C'(b_{\max}+1)\sqrt{\log n/n}$ and $V_n\ge v_*$. Then on $\mathcal G$ every $\widehat V_b(q^\star)$ with $b\in\mathcal W$ exceeds $v_{\min,n}$, and almost surely $q^\star$ is not an observation. On $\mathcal G\cap\{\hat b\in\mathcal W\}$, Proposition~\ref{prop:hullsandwich} with $b=\hat b$ and $\ell_{n,\hat b}\le\ell_{n,b_{\max}}$ gives
\begin{align*}
\{|S_n(q^\star)|\le z\sqrt{V_n-\epsilon^{\max}_n}\}&\subseteq\{q^\star\in C^{\rm H}_{n,\alpha}(\hat b)\},\\
\{q^\star\in C^{\rm H}_{n,\alpha}(\hat b)\}&\subseteq\{|S_n(q^\star)|\le z\sqrt{V_n+\epsilon^{\max}_n}+z^2\ell_{n,b_{\max}}\}.
\end{align*}
The sandwich holds pointwise for every window, so substituting the data-dependent $\hat b$ is legitimate. Neither bracketing event depends on $\hat b$, and $S_n(q^\star)$ does not depend on the window. The dependency-graph bound \eqref{eq:be} and the perturbation argument of Theorem~\ref{thm:hullcoverage} therefore give
\[
\big|\Pp(q^\star\in C^{\rm H}_{n,\alpha}(\hat b))-(1-\alpha)\big|\le2\beta_n+C_\alpha\Big\{\frac{\epsilon^{\max}_n}{V_n}+\frac{\ell_{n,b_{\max}}}{\sqrt{V_n}}\Big\}+\frac{2(b_{\max}+1)}n+\Pp(\hat b\notin\mathcal W).
\]
Under Assumption~\ref{as:score}, $\beta_n=O(n^{-1/2})$, and the second, third and fourth terms are $O((b_{\max}+1)\sqrt{\log n/n})$ for $n\ge2$, with constants depending on $r,\alpha,C_w,v_*$.
\end{proof}
The argument lets $\hat b$ depend arbitrarily on the data, because both bracketing events are deterministic functions of $S_n(q^\star)$, and it does not require $\hat b$ to converge. The rule $\hat b=\min(2\hat m,\lceil n^{1/3}\rceil)$ of Section~\ref{sec:method} doubles the selected lag, as recommended for flat-top windows \citep{politis2003adaptive}, so that a slightly underestimated memory is still covered. Since $\hat b\le\lceil n^{1/3}\rceil$, the corollary with $b_{\max}=\lceil n^{1/3}\rceil$ bounds its coverage error by $O(n^{-1/6}\sqrt{\log n})+\Pp(\hat b<r_\psi-1)$. For the guarded rule $\max(\hat b,\lceil n^{1/4}\rceil)$ this probability vanishes once $\lceil n^{1/4}\rceil\ge r_\psi-1$, so its coverage error is $O(n^{-1/6}\sqrt{\log n})$ for every fixed memory; Table~\ref{tab:v3auto} reports the selected windows.

\subsection{Local exactness of the convex hull}
This and the next subsection prove that the small-floor interval itself has $O(n^{-1/2})$ coverage error for fixed sufficient windows. Constants may depend on fixed $r,b,\alpha,C_w,v_*$ but not on $n$, and no density or moment of the observations is used.
\begin{proof}[Proof of the local exactness assertion in Proposition~\ref{prop:hullsandwich}]
Abbreviate $S_0=S_n(q^\dagger)$, $Q_0=\widehat V_b(q^\dagger)$ and $\ell=\ell_{n,b}$. Suppose $S_0>z\sqrt{Q_0}$ and $Q_0\ge z^2\ell^2/4$. An accepted point to the right would have score $s\ge S_0$. Acceptance on the floor is impossible since $s>z\sqrt{Q_0}>z\sqrt{v_{\min,n}}$. Otherwise \eqref{eq:scorelipmain} implies $s^2\le z^2\{Q_0+\ell(s-S_0)\}$, whereas
\[
s^2-z^2Q_0-z^2\ell(s-S_0)
=(S_0^2-z^2Q_0)+(s-S_0)(s+S_0-z^2\ell)>0.
\]
Indeed $2S_0>2z\sqrt{Q_0}\ge z^2\ell$. This contradiction excludes hull membership. The case $S_0<-z\sqrt{Q_0}$ follows by choosing an accepted point to the left and reversing score signs. Since $q^\dagger\notin\{X_1,\ldots,X_n\}$, it lies in a constant-score open interval, so closure cannot create an isolated membership exception. Raw acceptance always implies hull membership, proving equality.
\end{proof}

For a quantitative consequence let $Q=\widehat V_b(q^\star)$, $\bar V_{n,b}=\E Q=V_n-T_{n,b}$,
\[
a_n=\max\{n^{-1},z^2\ell_{n,b}^2/4\},\qquad
H_n=(n+r-1)r^2(4b+1)^2w_*^4.
\]
The no-atom assumption gives $q^\star\notin\{X_1,\ldots,X_n\}$ almost surely. Thus, for $\bar V_{n,b}>a_n$,
\begin{equation}\label{eq:hullexponential}
0\le\Pp(q^\star\in C^{\rm H}_{n,\alpha})-\Pp(q^\star\in\mathcal A_n)
\le\Pp(Q\le a_n)\le\exp\{-2(\bar V_{n,b}-a_n)^2/H_n\}.
\end{equation}
The last step is the one-sided innovation bounded-differences inequality. For fixed sufficient $b$, $\bar V_{n,b}\ge v_*$, $a_n=O(n^{-1})$, and $H_n=O(n^{-1})$. Thus the coverage added by taking the hull is exponentially small, even though the raw acceptance set can be disconnected elsewhere.

\subsection{Proof of Theorem~\ref{cor:sharpcoverage}}
\begin{proof}[Proof of Theorem~\ref{cor:sharpcoverage}]
Set $Z_i=w_i\psi_i$, $A_i=\{j\in\{1,\ldots,n\}:|j-i|\le b\}$, $Y_i=\sum_{j\in A_i}Z_j$, $S=\sum_iZ_i$, and $Q=\sum_iZ_iY_i=\widehat V_b(q^\star)$. The centered denominator in \citet[Theorem 2.1 in the cited preprint]{zhang2024self} is
\begin{equation}\label{eq:centereddenominator}
Q_c=Q-\frac Sn\sum_iY_i,\qquad W_c=\frac S{\sqrt{(Q_c)_+}}.
\end{equation}
Here $(x)_+=\max(x,0)$; on $Q_c\le0$ define $W_c=0$. We show below that this exceptional event has exponentially small probability, so the extension convention affects the approximation only at that order.

We verify the source theorem for this triangular array. The $Z_i$ are centered. The neighborhoods $A_i$ satisfy its (LD1) since $b\ge r-1$. With $A_i'=\{j:|j-i|\le2b\}$, the innovation sets generating $\{Z_j:j\in A_i\}$ and $\{Z_j:j\notin A_i'\}$ are disjoint; hence its vector independence condition (LD2) holds. Its neighborhood constant may be chosen as $\varkappa=6b+2$, including $b=0$. Since $|Z_i|\le w_*$ and $\sigma^2=\Var S=V_n\ge v_*$, eventually $|Z_i|\le\sigma/\varkappa$, so the truncated-tail terms $\xi_0,\xi_2$ vanish. The remaining parameters obey
\[
\xi_3\le V_n^{-3/2}\sum_i\E|Z_i|^3\le v_*^{-3/2}w_*,\qquad
\theta\le V_n^{-1}\sum_i\sum_{j\in A_i}\E|Z_iZ_j|\le(2b+1)/v_*.
\]
The last inequality uses $2|Z_iZ_j|\le Z_i^2+Z_j^2$ and $\sum_i\E Z_i^2\le1$. The source bound
$C(1+\theta)\varkappa^2\xi_3+C\varkappa^{1/2}(1+\theta)n^{-1/2}$ is consequently $O(n^{-1/2})$. Thus
\begin{equation}\label{eq:centeredbe}
\sup_x|\Pp(W_c\le x)-\Phi(x)|\le Cn^{-1/2}.
\end{equation}
A related self-normalized bound appears in \citet[Theorem 2.2]{cai2026refined}.

\paragraph{Exact boundary correction.}
Let $M=2b+1$, $\omega_j=M-|A_j|$ and $E_\partial=\sum_j\omega_jZ_j$. For $n>2b$, $\sum_j\omega_j=b(b+1)$ and symmetry of the neighborhoods gives
\begin{equation}\label{eq:boundaryidentity}
Q_c=Q-MS^2/n+SE_\partial/n,\qquad
|E_\partial|\le b(b+1)w_*.
\end{equation}
Also $|Q|\le M\sum_iZ_i^2\le M$ and $|S|\le\sqrt n$. In particular $|SE_\partial|\le b(b+1)C_w$ deterministically.

\paragraph{Nonpositive denominators are exponentially unlikely.}
Let $\mathcal G=\{Q\ge v_*/2\}$. Since $\E Q=V_n\ge v_*$, innovation replacement yields
\[
\Pp(\mathcal G^c)\le\exp\{-v_*^2/(2H_n)\}\le Ce^{-cn}.
\]
On $\mathcal G\cap\{Q_c\le0\}$, \eqref{eq:boundaryidentity} implies $MS^2/n\ge v_*/4$ for all sufficiently large $n$. A replacement changes $S$ by at most $rw_*$. Bounded differences and $\E S=0$ therefore give
\[
\Pp(Q_c\le0)\le\Pp(\mathcal G^c)+
2\exp\left\{-\frac{nv_*}{2M(n+r-1)r^2w_*^2}\right\}\le Ce^{-cn}.
\]
This establishes the convention claim made before \eqref{eq:centeredbe}.

\paragraph{Transfer at a fixed critical value.}
Write $\mathcal H(z)=\{|S|\le z\sqrt{Q\vee n^{-1}}\}$ and $\mathcal H_c(a)=\{Q_c>0,|S|\le a\sqrt{Q_c}\}$. On $\mathcal H(z)$, $|S|\le z\sqrt M$. On $\mathcal H_c(a)$ for $0\le a\le z+1$, the upper bound $Q_c\le M+b(b+1)C_w/n$ likewise bounds $|S|$ by a fixed constant $K$. On either acceptance event, \eqref{eq:boundaryidentity} consequently implies $|Q-Q_c|\le C/n$. Intersecting with $\mathcal G$ gives $Q_c\ge v_*/4$ and, for large $n$, an inactive $1/n$ floor. Hence
\[
\left|S/\sqrt Q-S/\sqrt{Q_c}\right|\le C/n
\]
on the events being compared. With $\delta_n=C/n$ sufficiently large and eventually smaller than $\min(1,z)$, we obtain the deterministic inclusions
\[
\mathcal H_c(z-\delta_n)\cap \mathcal G\subseteq \mathcal H(z),\qquad
\mathcal H(z)\cap \mathcal G\subseteq \mathcal H_c(z+\delta_n).
\]
Equation~\eqref{eq:centeredbe}, Gaussian anti-concentration, and the exponentially small exceptional events imply
\[
|\Pp(\mathcal H(z))-(1-\alpha)|\le Cn^{-1/2}+C\delta_n+Ce^{-cn}.
\]
For discrete pivots, use the CDF left limit at the lower endpoint; the same Kolmogorov bound controls it. Finally $z^2\ell_{n,b}^2/4=O(n^{-1})$, so local hull exactness applies on $\mathcal G$ for large $n$. Equation~\eqref{eq:hullexponential} transfers the last display to the hull. Enlarging the constant covers the finitely many remaining $n$. This proves Theorem~\ref{cor:sharpcoverage}.
\end{proof}

Fixed $b$ is essential to the stated order: both neighborhood and denominator-comparison constants depend on $b$. Growing windows and insufficient windows retain the explicit bound in Theorem~\ref{thm:hullcoverage}.

\subsection{The order cannot be improved}\label{app:lattice}
Let $X_t$ be i.i.d., $b=0$ and $w_t=n^{-1/2}$. For $q$ between consecutive order statistics put $N(q)=\sum_t\ind(X_t\le q)$, so that $S_n(q)=\{N(q)-n\tau\}/\sqrt n$ and $\widehat V_0(q)=\tau^2+(1-2\tau)N(q)/n$. Since $n\tau^2+(1-2\tau)N\ge n\min\{\tau^2,(1-\tau)^2\}$, the floor never binds for $n\ge\max\{\tau^{-2},(1-\tau)^{-2}\}$. Acceptance then means $\{N-n\tau\}^2\le z^2\{n\tau^2+(1-2\tau)N\}$, a quadratic inequality in $N$; because $N(q)$ is nondecreasing in $q$, $\mathcal A_n$ is an interval and the hull adds nothing. With $\sigma_n^2=n\tau(1-\tau)$ and $c_\tau=z^2(1-2\tau)/2$, the inequality holds for $N-n\tau$ between $c_\tau\pm(c_\tau^2+z^2\sigma_n^2)^{1/2}$. With the exact roots $y_n^\pm=n\tau+c_\tau\pm(c_\tau^2+z^2\sigma_n^2)^{1/2}=n\tau+c_\tau\pm z\sigma_n+O(n^{-1/2})$,
\[
\Pp(q^\star\in C^{\rm H}_{n,\alpha})=\Pp(\lceil y_n^-\rceil\le N\le\lfloor y_n^+\rfloor)
\]
for $N\sim\mathrm{Bin}(n,\tau)$. Esseen's expansion for lattice distributions \citep[Section~23]{bhattacharya1976} gives, uniformly over integers $k$,
\[
\Pp(N\le k)=\Phi(x_k)-\frac{1-2\tau}{6\sigma_n}(x_k^2-1)\phi(x_k)+\frac{\phi(x_k)}{2\sigma_n}+O(n^{-1}),\qquad x_k=\frac{k-n\tau}{\sigma_n},
\]
because the periodic term, evaluated at the lattice point $k$, reduces to the continuity correction. The coverage probability is $\Pp(N\le k_+)-\Pp(N\le k_-)$ with $k_+=\lfloor y_n^+\rfloor$ and $k_-=\lceil y_n^-\rceil-1$, and $x_{k_\pm}=\pm z+O(n^{-1/2})$. At these two points the skewness terms agree to $O(n^{-1})$, and $c_\tau$ shifts both endpoints equally. Since $\Phi(x)+\phi(x)/(2\sigma_n)=\Phi\{x+1/(2\sigma_n)\}+O(n^{-1})$, expanding $\Phi$ at $\pm z$ gives
\[
\Pp(q^\star\in C^{\rm H}_{n,\alpha})-(1-\alpha)=\frac{\phi(z)}{\sigma_n}\Lambda_n+O(n^{-1}),\qquad
\Lambda_n=\lfloor y_n^+\rfloor-\lceil y_n^-\rceil+1-(y_n^+-y_n^-),
\]
the number of integers in $[y_n^-,y_n^+]$ minus the length of that interval. Thus $\Lambda_n\in(-1,1]$ and $\Lambda_n\equiv-(y_n^+-y_n^-)$ modulo $1$. Since $y_n^+-y_n^-=2z\sigma_n+O(n^{-1/2})$ increases without bound in increments that tend to zero, its fractional part comes arbitrarily close to $1/2$ infinitely often, and then $|\Lambda_n|$ is close to $1/2$. Hence $\limsup_nn^{1/2}|\Pp(q^\star\in C^{\rm H}_{n,\alpha})-(1-\alpha)|\ge\phi(z)/\{2\sqrt{\tau(1-\tau)}\}>0$, the oscillation documented for binomial confidence intervals by \citet{brown2002}. This lower bound is conservative. Table~\ref{tab:lattice} reports exact binomial sums at 400 sample sizes from $200$ to $2\times10^5$: the range of $n^{1/2}$ times the coverage error does not shrink to zero and approaches the amplitude $\phi(z)/\sqrt{\tau(1-\tau)}$.

\begin{table}[H]
\centering\small
\caption{Range of $n^{1/2}\{\Pp(q^\star\in C^{\rm H}_{n,.05})-.95\}$ for i.i.d.\ data, $b=0$ and equal weights, from exact binomial sums.}\label{tab:lattice}
\begin{tabular}{rrrrr}
\toprule
$\tau$ & $200\le n<2{,}000$ & $2{,}000\le n<20{,}000$ & $20{,}000\le n\le200{,}000$ & $\phi(z)/\sqrt{\tau(1-\tau)}$\\
\midrule
0.10 & -0.324 to 0.133 & -0.188 to 0.153 & -0.169 to 0.159 & 0.195\\
0.25 & -0.155 to 0.111 & -0.131 to 0.113 & -0.124 to 0.121 & 0.135\\
0.50 & -0.119 to 0.114 & -0.117 to 0.116 & -0.116 to 0.116 & 0.117\\
\bottomrule
\end{tabular}
\end{table}

\section{Precision, first-order equivalence and extensions}\label{app:precision}
\subsection{Endpoint expansion and local alternatives}
Let $W_n=\sum_tw_t$ and let $\widehat q_{w,\tau}$ be the weighted empirical quantile defined in the main text. The following general statement implies Corollary~\ref{cor:fixedprecision}, since a fixed sufficient window gives $T_{n,b}=0$, $\ell_{n,b}\to0$ and $\epsilon_n(\log n)\to0$.
\begin{theorem}[Endpoint expansion]\label{thm:endpoints}
Suppose Assumptions~\ref{as:memory} and \ref{as:score} hold for fixed $r$, $F$ is differentiable at $q^\star$ with derivative $h>0$, and
\[
\ell_{n,b}\to0,\qquad \epsilon_n(\log n)+|T_{n,b}|\to0.
\]
Then the hull interval is nonempty and bounded with probability tending to one. Its endpoints satisfy
\begin{equation}\label{eq:endpointsgeneral}
\begin{split}
L_n&=\widehat q_{w,\tau}-\frac{z\sqrt{V_n}}{hW_n}+\op(n^{-1/2}),\\
U_n&=\widehat q_{w,\tau}+\frac{z\sqrt{V_n}}{hW_n}+\op(n^{-1/2}).
\end{split}
\end{equation}
If $V_n\to V>0$, then for each fixed $x\in\R$,
\begin{equation}\label{eq:localpower}
\begin{split}
\Pp(q^\star+x/W_n\in C^{\rm H}_{n,\alpha})\longrightarrow{}&
\Phi(z-hx/\sqrt V)\\
&-\Phi(-z-hx/\sqrt V).
\end{split}
\end{equation}
\end{theorem}

\begin{proof}[Proof of Theorem~\ref{thm:endpoints}]
Write $S_*=S_n(q^\star)$, $Q_*=\widehat V_b(q^\star)$, $W=W_n$, and $\ell=\ell_{n,b}$. From $\sum_tw_t^2=1$, $w_t\ge0$, and $w_*\le C_w/\sqrt n$, we have $\sqrt n/C_w\le W\le\sqrt n$. For fixed $x$, define the local score increment at $q_x=q^\star+x/W$. Its expectation is
\[
W\{F(q_x)-F(q^\star)\}=hx+o(1).
\]
Every indicator increment has marginal second moment $O(|x|/W)$ on a fixed local window. Finite dependence and $2|w_tw_s|\le w_t^2+w_s^2$ bound the centered increment's variance by $C_r|x|/W=o(1)$. Monotonicity of the score and a finite grid, refined after $n\to\infty$, yield
\begin{equation}\label{eq:localscoreuniform}
\sup_{|x|\le K}|S_n(q^\star+x/W)-S_*-hx|\longrightarrow0
\quad\text{in probability}
\end{equation}
for every fixed $K$. The variance theorem gives $Q_*-V_n=\op(1)$. By \eqref{eq:scorelipmain}, the supremum of $|\widehat V_b(q^\star+x/W)-V_n|$ on $|x|\le K$ is also $\op(1)$.

We next localize the entire acceptance set. If $q\in\mathcal A_n$ and its variance exceeds the floor, then
\[
|S_n(q)|^2\le z^2\{Q_*+\ell |S_n(q)-S_*|\}
\le z^2\{|Q_*|+\ell(|S_n(q)|+|S_*|)\}.
\]
The positive root gives an $O_{\Pp}(1)$ upper bound on $|S_n(q)|$, simultaneously for all accepted $q$, because $S_*=O_{\Pp}(1)$, $Q_*=O_{\Pp}(1)$, and $\ell=o(1)$. If the floor is active, $|S_n(q)|\le z/\sqrt n$ directly. Equation~\eqref{eq:localscoreuniform} and monotonicity now imply that every accepted point lies within $O_{\Pp}(W^{-1})$ of $q^\star$: on an event where the preceding bounds and $|S_*|$ are at most a fixed constant, choose $K$ large enough that the scores at $q^\star\pm K/W$ exceed these bounds with the appropriate signs. The same localization holds for the closed hull.

On this root-$n$ neighborhood, the critical radius is uniformly $z\sqrt{V_n}+\op(1)$. Its lower and upper crossings in score space must therefore equal $-z\sqrt{V_n}+\op(1)$ and $z\sqrt{V_n}+\op(1)$. Existence follows by testing local scores a fixed small distance inside and outside these thresholds and then letting that distance decrease. To include possible ties, observe that \eqref{eq:localscoreuniform} forces the largest local jump to be $\op(1)$, since its limiting local function is continuous; no assumption that two different observations are almost surely distinct is needed. These arguments show that the acceptance set, and hence the hull, is nonempty and bounded with probability tending to one.

The same expansion gives $hW(\widehat q_{w,\tau}-q^\star)=-S_*+\op(1)$; a tied jump at the weighted quantile is again $\op(1)$. Consequently
\[
hW(L_n-q^\star)=-S_*-z\sqrt{V_n}+\op(1),\qquad
hW(U_n-q^\star)=-S_*+z\sqrt{V_n}+\op(1),
\]
which proves \eqref{eq:endpointsgeneral}. The argument also establishes the weighted-quantile expansion when observations have ties.

If $V_n\to V$, the dependency-graph bound gives $S_*\Rightarrow N(0,V)$. Use the joint endpoint expansions just proved: membership of $q^\star+x/W$ is the event $hW(L_n-q^\star)\le hx\le hW(U_n-q^\star)$. Its limiting inequalities are $-S_*-z\sqrt V\le hx\le -S_*+z\sqrt V$. The two Gaussian boundary probabilities are zero, so joint convergence gives \eqref{eq:localpower}. This argument also applies when the moving candidate is an atom of the marginal law; it does not require that candidate to be unobserved.
\end{proof}

For the normalized finite SGD filters, Lemma~\ref{lem:filters} gives $W_n/\sqrt n\to1$; symmetrization preserves this. For each fixed $k$, $\sum_{t>k}w_tw_{t-k}\to1$. Thus finite dependence yields $V_n\to\Gamma_0+2\sum_{k<r}\Gamma_k=\Omega$. The deterministic width in Theorem~\ref{thm:endpoints} is then the usual marginal-quantile width. The width expansion holds in probability; convergence of expected length additionally requires uniform integrability.
\subsection{Weighted-quantile center and first-order equivalence}\label{app:equivalence}
Let $p_{nt}=w_{nt}/\sum_sw_{ns}$ and $F_{n,p}(q)=\sum_tp_{nt}\ind(X_t\le q)$. Write $\widehat q_{w,\tau}=\inf\{q:F_{n,p}(q)\ge\tau\}$. Suppose $F$ is differentiable at $q^\star$, with $h>0$, $\max_tp_{nt}\le C/n$, and $\sum_tp_{nt}^2=O(n^{-1})$.
For fixed $x$, the centered local increment
\[
Z_n(x)=\sum_tp_{nt}\{\ind(X_t\le q^\star+x/\sqrt n)-\ind(X_t\le q^\star)
-[F(q^\star+x/\sqrt n)-F(q^\star)]\}
\]
has variance at most $C_r n^{-3/2}|x|$. This follows from finite dependence, Cauchy--Schwarz for the incident covariances, and the marginal probability of the local interval. Hence $\sqrt nZ_n(x)\to0$ in probability. Monotonicity and a finite grid imply uniform convergence over every compact set of $x$: adjacent grid increments have expected size at most $C\Delta/\sqrt n$, and the grid can be refined after taking the limit.

The variance of $F_{n,p}(q^\star)-\tau$ is $O_r(n^{-1})$. Comparing the CDF at $q^\star\pm K/\sqrt n$ proves $\sqrt n$-tightness of $\widehat q_{w,\tau}-q^\star$ as $K\to\infty$. Uniform local convergence to a continuous linear function makes the largest local empirical jump $\op(n^{-1/2})$, including tied observations. Substitution in the local increment expansion yields
\begin{equation}\label{eq:weightedbahadur}
\sqrt n(\widehat q_{w,\tau}-q^\star)
=-h^{-1}\sqrt n\sum_tp_{nt}\psi_t+\op(1).
\end{equation}
For the finite-filter weights, Lemma~\ref{lem:filters} and finite dependence give
\[
\Var\left(\sqrt n\sum_t(p_{nt}-n^{-1})\psi_t\right)
\le C_r n\sum_t(p_{nt}-n^{-1})^2\to0.
\]
The raw SGD expansion in Proposition~\ref{prop:bahadur} has a remainder tending to zero. Applying Lemma~\ref{lem:filters} to its true-slope weights gives the same first-order linear term, $-(h\sqrt n)^{-1}\sum_t\psi_t$. This proves Proposition~\ref{prop:equivalence} for equal and original finite-filter weights, consistent with dependent-quantile representations \citep{bahadur1966,wu2005bahadur}; Proposition~\ref{prop:symmetry} covers symmetric weights.

\subsection{Fixed collections of quantiles}
For each fixed $\tau_j$, suppose Assumptions~\ref{as:memory} and \ref{as:score} hold at that level, and construct the small-floor hull interval $[L_j,U_j]$ with $z=z_{1-\alpha/(2K)}$. Theorem~\ref{thm:hullcoverage} gives a marginal error $e_{n,j}$ from \eqref{eq:hullcoverage}. The union bound proves
\[
\Pp\{q^\star(\tau_j)\in[L_j,U_j]\text{ for every }j\}\ge1-\alpha-\sum_{j=1}^K e_{n,j}.
\]
For fixed $K$, if every level satisfies Theorem~\ref{cor:sharpcoverage} with fixed sufficient windows, the displayed lower bound is $1-\alpha-C_Kn^{-1/2}$. No independence across quantile levels is assumed. The same argument applies to connected-floor intervals using Theorem~\ref{thm:coverage} and its corresponding errors. The tightened endpoints
$L_j^\uparrow=\max_{k\le j}L_k$ and $U_j^\uparrow=\min_{k\ge j}U_k$
preserve the simultaneous event because the true quantile curve is nondecreasing. The simultaneous-coverage lower bound may be conservative. Extending it to growing $K$ would require controlling the sum of marginal errors and the normal critical values.

\subsection{Geometric indicator coupling}\label{app:geometric}
Assume that for every $r$ one can construct, jointly with $X_t$, a stationary process
$X_t^{(r)}=G_r(\zeta_{t-r+1},\ldots,\zeta_t)$ from i.i.d. innovations, such that $X_t^{(r)}$ has the same marginal distribution as $X_t$ and
\begin{equation}\label{eq:coupling}
\Pp\{\ind(X_t\le q^\star)\ne\ind(X_t^{(r)}\le q^\star)\}\le C\rho^r,\qquad 0<\rho<1.
\end{equation} The augmented innovation $\zeta_t$ may include a fresh independent copy of a remote past used only at time $t$. Assume the approximate weighted score variances are uniformly bounded below. Put $b=r-1$.
On the event that all $n$ threshold indicators agree, the score and candidate variance at $q^\star$ agree exactly for the original and approximate data. That event fails with probability at most $nC\rho^r$. The coverage argument depends only on acceptance at $q^\star$, so the random endpoints do not need to coincide elsewhere. Apply Theorem~\ref{thm:coverage} to the approximating data and add $nC\rho^r$.
Choose $r=\lceil c_\rho\log n\rceil$ with $c_\rho>3/(2|\log\rho|)$. Then $n\rho^r=o(n^{-1/2})$,
$\beta_n=O((\log n)^2/\sqrt n)$,
$\epsilon_n(\log n)=O((\log n)^{5/2}/\sqrt n)$,
and $v_{\rm conn,n}=O((\log n)^2/n)$. Thus $b=r-1\asymp\log n$ gives coverage error $O((\log n)^{5/2}/\sqrt n)$; the window constant is set by the coupling envelope.

For a primitive sufficient crossing condition, suppose $X_t$ has density bounded near $q^\star$ and $\norm{X_t-X_t^{(r)}}_{p_0}\le C\varrho^r$ for some $p_0>0$. For any $a>0$,
\[
\Pp\{\ind(X_t\le q^\star)\ne\ind(X_t^{(r)}\le q^\star)\}
\le Ca+C^{p_0}\varrho^{p_0r}a^{-p_0}.
\]
Choosing $a\asymp\varrho^{p_0r/(p_0+1)}$ gives a geometric indicator bound. Alternatively, an increasing heavy-tailed transformation preserves indicator mismatch from a latent process, without requiring that transformed observations have a moment. The resulting assumption concerns indicator coupling directly. For a causal process $X_t=G(\ldots,\varepsilon_{t-1},\varepsilon_t)$, replacing the innovations before time $t-r+1$ by an independent copy gives a same-marginal coupling, and the moment condition above holds when the physical dependence measures of \citet{wu2005physical} decay geometrically.

\section{Experimental design}\label{app:experiments}
\subsection{Guide to interval configurations}
Table~\ref{tab:configindex} identifies the weights, window, and floor for each interval study. Counts refer to independent streams per cell; methods and marginal transformations are paired within streams. Three later studies were prespecified in the same way, with protocols hashed before sampling: the bias-scale study (Appendix~\ref{app:finite}), the interval comparison of Appendix~\ref{app:v3intervals}, and extrapolation at $\gamma=.85$ (Appendix~\ref{app:rr085}); they use new seeds and leave all earlier samples unchanged. Write $b_{\log}=\lceil\sqrt{\log n}\rceil$, $b_{\rm pow}=\lceil n^{1/4}\rceil$, and $v_\lambda=(\lambda z(4b+1)w_*)^2$ for normalized weights.
\begin{table}[H]
\centering\small
\caption{Index of interval experiments. ``Filter'' denotes the original finite filter; ``Sym'' its symmetric version; ``Equal'' equal weights. A reused sample is indicated explicitly.}\label{tab:configindex}
{\setlength{\tabcolsep}{4pt}
\begin{tabular}{p{.22\textwidth}p{.13\textwidth}p{.12\textwidth}p{.11\textwidth}p{.12\textwidth}p{.19\textwidth}}
\toprule
Study & Weights & Window & Floor & Streams/cell & Results\\
\midrule
CBB comparison & Sym, Equal & $b_{\rm pow}$ & $1/n$ & 600 & Table~\ref{tab:block600}\\
SN, $\hat b$, online pivot & Equal & $b_{\rm pow},\hat b,r-1$ & $1/n$ & 1,600 independent & Table~\ref{tab:block}; Appendix~\ref{app:v3intervals}\\
Hull validation & Sym, Equal & $b_{\rm pow}$ & $1/n$ & 1,600 & Table~\ref{tab:hullfull}; floor controls also use $v_{1/2}$\\
Fixed-window control & Sym, Equal & $11,b_{\rm pow},b_{\log}$ & $1/n$ & same 1,600 & Tables~\ref{tab:fixedwindow}, \ref{tab:fixedfull}\\
AR window sensitivity & Equal & $8,15,30,60$ & $1/n$ & 1,600 independent & Table~\ref{tab:arwindow}\\
Negative/scale checks & Sym, Equal, Filter & $b_{\rm pow}$ & $1/n$ & 1,600 independent & Table~\ref{tab:edgefull}; S-floor uses $v_{1/2}$\\
Known-window & Filter, Equal & $r-1$ & $v_1$ & 1,600 & Appendix~\ref{app:tables}; with Wald controls\\
Log-window & Filter, Equal & $b_{\log}$ & $v_{1/2}$ & 1,600 independent & Appendix~\ref{app:tables}; with Wald controls\\
Simultaneous & Filter, Equal & $b_{\log}$ & $v_{1/2}$ & 1,600 independent & Table~\ref{tab:simultaneous}\\
\bottomrule
\end{tabular}}
\end{table}

\subsection{Observation models, quantiles and sampling budgets}
The Wald comparisons use stationary finite-innovation-memory models. The hull study also includes an infinite-memory AR process (Appendix~\ref{app:revisionexperiments}). Let $U_t$ be i.i.d. standard normal. The five latent models are
\begin{align*}
\text{Independent: }&Z_t=U_t,\\
\text{Positive MA(1): }&Z_t=(U_t+U_{t-1})/\sqrt2,\\
\text{Negative MA(1): }&Z_t=(U_t-.5U_{t-1})/\sqrt{1.25},\\
\text{Memory four: }&Z_t=(U_t+U_{t-1}+U_{t-2}+U_{t-3})/2,\\
\text{Scale mixture: }&Z_t=(.5+.5\Phi(U_{t-1}))U_t.
\end{align*}
The innovation memory lengths are $1,2,2,4,2$, respectively. For the first four, the two observation margins are $X_t=Z_t$ and
$X_t=\tan\{\pi(\Phi(Z_t)-1/2)\}$. The latter is standard Cauchy. Their true quantiles are $\Phi^{-1}(\tau)$ and $\tan\{\pi(\tau-1/2)\}$.
The scale-mixture model is used without transformation. Since $.5+.5\Phi(U_{t-1})$ is uniform on $[.5,1]$ and independent of $U_t$, its marginal CDF is
$F(q)=2\int_{.5}^1\Phi(q/s)\,\dd s$. Its quantile is obtained by numerical integration and scalar root finding. The target is the marginal quantile of this conditionally heteroskedastic process.

For the Wald comparisons, each replication supplies a common chronological budget: $M=\lceil n^{3/4}\rceil$ pilot observations, $r-1$ guard observations, and $n$ analysis observations. A common underlying innovation stream generates all three segments. The guard separates the analysis observations from the pilot innovations. The pilot sizes are $406$ and $1{,}147$ at $n=3{,}000$ and $12{,}000$. Inversion uses the analysis observations, with the same total observation budget charged to each method.

Normal and Cauchy margins share the same latent streams. Inversion endpoints transform exactly under the increasing marginal map, so coverage indicators coincide. Direct inversion of transformed samples verifies this identity numerically. Wald intervals are evaluated on the transformed observation values.

\subsection{Independent samples and pairing}
\begin{figure*}[t]
\centering\includegraphics[width=.96\textwidth]{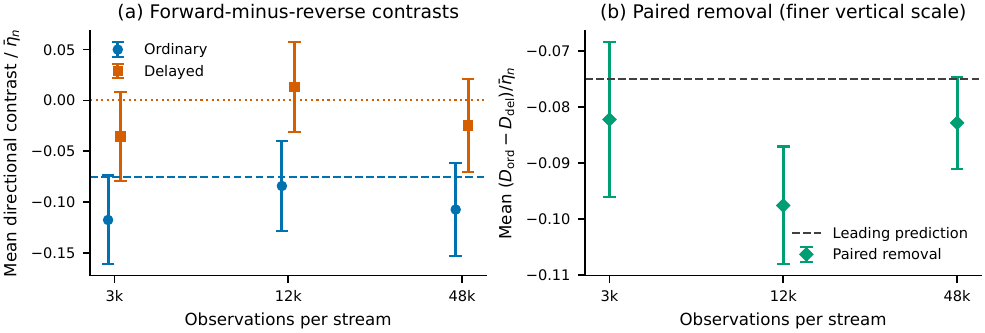}
\caption{Delaying the threshold removes the directional component. Both directions and algorithms share each of 2,500 independent streams per sample size. (a) Normalized means of $D_{\rm ord}$ and $D_{\rm del}$; reference lines give their leading predictions. (b) Normalized mean of $D_{\rm ord}-D_{\rm del}$ on a finer vertical scale; the dashed line is $B_\tau^{\to}-B_\tau^{\leftarrow}$. Bars are mean $\pm1.96$ MCSE. }\label{fig:delay}
\end{figure*}

The two-delay experiment uses 2,500 independent streams at each of three sample sizes, with directions and algorithms paired within streams. These streams are independent of the 4,000-stream reversal experiment. In Figure~\ref{fig:delay}, $D_{\rm ord}=D_1$ and $D_{\rm del}=D_2$; both panels divide the displayed contrasts by $\bar\eta_n$.

A 1,200-stream-per-cell diagnostic study informed the hull design. Performance estimates in Table~\ref{tab:hullfull} use an independent sample of 1,600 streams in each of eight cells. The five-cell bootstrap comparison uses a further independent sample of 600 streams per cell and 399 resamples per stream. All three block lengths share those 600 streams, with separate random-number generators for data and bootstrap draws. The diagnostic data and sampling protocols are included in the artifact.

For moving averages, generate $r$ independent innovations per moving window and normalize the coefficient vector to Euclidean norm one. Thus $r=12$ gives an MA(11). The stationary AR control is initialized from its exact $N(0,1)$ stationary distribution and uses innovation standard deviation $.6$. The Cauchy transform is applied afterward. No raw observation is clipped or winsorized. The paired delayed experiment projects states to $[-6,6]$ and starts at the target; Appendix~\ref{app:readability} repeats it from $q_0=0$.

The circular bootstrap uses uniform starting indices, wraps blocks modulo $n$, and truncates only the final block so that every resample has exactly $n$ observations. Each resample returns the inverse empirical CDF at probability $.25$. The outer $.025$ and $.975$ percentiles use linear interpolation between ordered bootstrap estimates (NumPy \texttt{method="linear"}). Sorting the original observations once and accumulating resampling counts is algebraically identical to explicitly resampling and sorting; this identity is tested. Block lengths are $j\lceil n^{1/3}\rceil$, $j\in\{1,2,4\}$. The entire block-length grid is reported, with the score window fixed independently of its outcomes.

\subsection{Implementation}
The implementation uses NumPy, SciPy and Numba. Direct quadratic-form calculations verify the scan with tied observations, signed lag contributions and increasing transformations; direct candidate enumeration checks ties, disconnected acceptance sets and empty or unbounded hulls; and direct unrolling of the linear recursion verifies the finite-filter weights. At the full window $b=n-1$ the code uses the identity $Q=S^2$: all candidates are accepted for $z\ge1$, and otherwise acceptance reduces to $|S|\le z\sqrt{\mathrm{floor}}$, which avoids rounding differences between algebraically equal expressions. Saved per-stream endpoints support reaggregation of coverage, directional misses, length and Monte Carlo uncertainty; the artifact contains these checks and the full protocols.

\section{Bias experiments}\label{app:biasexp}
This appendix gives the designs and complete results of the bias experiments summarized in Section~\ref{sec:experiments}; the finite-$n$ predictions come from Appendix~\ref{app:finite}.

\subsection{Cauchy, reversal and two-delay experiments}
For the Cauchy bias experiment, use $Z_t=r^{-1/2}\sum_{j=0}^{r-1}U_{t-j}$ followed by the Cauchy transformation, $r\in\{1,2,4\}$, $\tau=.25$, and true initial value $q_0=-1$. For each $r$ and $n\in\{3{,}000,12{,}000,48{,}000\}$, we run 4,000 independent paired trajectories with gains $3$ and $1.5$. The extrapolated estimate is $2\bar q_n(1.5)-\bar q_n(3)$, using the same data for both recursions. There are 36,000 paired replications and 72,000 SGD trajectories. Model constants are calculated independently from the latent Gaussian law.

Appendix~\ref{app:cauchyexample} gives the model constants and the coefficients $B_\tau=-.2945$, $-.3775$ and $-.5659$ for memories $1,2,4$. Table~\ref{tab:bias} gives leading and finite-$n$ predictions, mean biases and last-iterate second moments. The leading term understates the observed biases, which the finite-$n$ recursion matches within 1.6 MCSE.

For the reversal example \eqref{eq:nonreversible}, the same 4,000 samples are processed forward and backward at each of the three sample sizes, producing 12,000 paired replications. Both runs start at the numerically calculated true quantile. The same compact projection and learning rates are used. The constants are those of Appendix~\ref{app:biasscale}, computed with 160 Gauss--Hermite nodes and checked against 220 nodes (largest discrepancy below $2.1\times10^{-12}$); in particular $B^{\to}-B^{\leftarrow}=-.075018187$. Table~\ref{tab:reversal} gives the paired contrast and the two single-direction biases with leading and finite-$n$ predictions; Appendix~\ref{app:finite} discusses the single-direction biases at $n=48{,}000$.

\begin{table}[ht]
\centering\small
\caption{Cauchy signed-bias experiment, 4,000 paired replications per row. Biases, RR biases and MCSEs are multiplied by $10^3$. ``Leading'' is $B_\tau\bar\eta_n$ and ``Finite-$n$'' the recursion of Appendix~\ref{app:finite}. The last three columns compare the uncentered last-iterate second moment divided by the step size with its limit $\Omega/(2h)$ and with the finite-$n$ recursion.}\label{tab:bias}
{\setlength{\tabcolsep}{3.5pt}
\begin{tabular}{rrrrrrrrrr}
\toprule
$r$ & $n$ & Leading & Finite-$n$ & Mean bias (MCSE) & RR bias (MCSE) & RR finite-$n$ & $\E e_n^2/\eta_n$ (MCSE) & Limit & Finite-$n$\\
\midrule
1 & 3k & -6.227 & -9.403 & -8.045 (0.862) & -1.499 (1.067) & -3.093 & 0.682 (0.015) & 0.589 & 0.662\\
1 & 12k & -2.457 & -3.594 & -3.765 (0.416) & -1.861 (0.515) & -1.630 & 0.670 (0.015) & 0.589 & 0.638\\
1 & 48k & -0.933 & -1.293 & -1.350 (0.207) & -0.642 (0.238) & -0.554 & 0.617 (0.015) & 0.589 & 0.623\\
2 & 3k & -7.982 & -12.418 & -12.351 (1.094) & -3.502 (1.348) & -4.276 & 1.108 (0.025) & 0.952 & 1.070\\
2 & 12k & -3.150 & -4.741 & -4.559 (0.530) & -1.846 (0.649) & -2.288 & 1.045 (0.024) & 0.952 & 1.032\\
2 & 48k & -1.196 & -1.700 & -1.668 (0.265) & -0.814 (0.305) & -0.780 & 1.026 (0.022) & 0.952 & 1.006\\
4 & 3k & -11.964 & -19.155 & -18.073 (1.482) & -6.169 (1.809) & -6.849 & 1.953 (0.043) & 1.729 & 1.944\\
4 & 12k & -4.721 & -7.306 & -6.654 (0.696) & -3.251 (0.866) & -3.726 & 1.872 (0.043) & 1.729 & 1.873\\
4 & 48k & -1.793 & -2.612 & -2.244 (0.355) & -0.889 (0.407) & -1.274 & 1.875 (0.043) & 1.729 & 1.828\\
\bottomrule
\end{tabular}}
\end{table}
\begin{table}[ht]
\centering\small
\caption{Time reversal, 4,000 paired replications per row. Biases, contrasts and MCSEs are multiplied by $10^3$. Predictions use the leading term and the finite-$n$ recursion of Appendix~\ref{app:finite}.}\label{tab:reversal}
{\setlength{\tabcolsep}{4pt}
\begin{tabular}{rrrrrrrr}
\toprule
 & \multicolumn{3}{c}{Contrast} & \multicolumn{2}{c}{Forward bias} & \multicolumn{2}{c}{Reverse bias}\\
$n$ & Leading & Finite-$n$ & Observed (MCSE) & Observed (MCSE) & Finite-$n$ & Observed (MCSE) & Finite-$n$\\
\midrule
3k & -1.586 & -1.928 & -1.656 (0.369) & -2.211 (0.536) & -3.101 & -0.555 (0.544) & -1.173\\
12k & -0.626 & -0.745 & -0.797 (0.148) & -1.330 (0.262) & -1.192 & -0.532 (0.263) & -0.446\\
48k & -0.238 & -0.276 & -0.217 (0.058) & -0.787 (0.129) & -0.437 & -0.570 (0.129) & -0.161\\
\bottomrule
\end{tabular}}
\end{table}

\subsection{The delayed recursion from an ordinary initial value}\label{app:readability}
The ordinary-initialization experiment uses $q_0=0$ in both directions and delays $d\in\{1,2\}$. It retains $c=3$, $\gamma=.75$, $t_0=20$, projection to $[-6,6]$, and 2,500 streams per sample size, with seeds $73092100+n$.

\begin{table}[H]
\centering\small
\caption{Ordinary initialization, 2,500 streams per sample size. Bias, finite-$n$ prediction (Appendix~\ref{app:finite}), unbiased sample variance, MSE and MCSEs are multiplied by $10^3$; projection counts are unscaled.}\label{tab:nonoraclefull}
\begin{tabular}{rrlrrrrrrr}
\toprule
$n$ & $d$ & Direction & Bias & MCSE & Finite-$n$ & Variance & MSE & MSE MCSE & Projections\\
\midrule
3,000 & 1 & forward & -0.8454 & 0.6713 & -0.9396 & 1.12675 & 1.12701 & 0.03125 & 0\\
3,000 & 1 & reverse & 1.7184 & 0.6702 & 0.9883 & 1.12300 & 1.12550 & 0.03153 & 0\\
3,000 & 2 & forward & -0.1541 & 0.6700 & -0.2251 & 1.12218 & 1.12176 & 0.03124 & 0\\
3,000 & 2 & reverse & 0.5374 & 0.6676 & -0.2251 & 1.11430 & 1.11415 & 0.03120 & 0\\
12,000 & 1 & forward & -0.6319 & 0.3365 & -0.6512 & 0.28315 & 0.28344 & 0.00781 & 0\\
12,000 & 1 & reverse & 0.0529 & 0.3357 & 0.0940 & 0.28169 & 0.28158 & 0.00811 & 0\\
12,000 & 2 & forward & -0.3277 & 0.3361 & -0.3515 & 0.28237 & 0.28236 & 0.00778 & 0\\
12,000 & 2 & reverse & -0.4271 & 0.3353 & -0.3515 & 0.28113 & 0.28120 & 0.00806 & 0\\
48,000 & 1 & forward & -0.1753 & 0.1671 & -0.3018 & 0.06979 & 0.06979 & 0.00197 & 0\\
48,000 & 1 & reverse & 0.1589 & 0.1677 & -0.0263 & 0.07031 & 0.07031 & 0.00198 & 0\\
48,000 & 2 & forward & -0.0650 & 0.1669 & -0.1858 & 0.06964 & 0.06962 & 0.00197 & 0\\
48,000 & 2 & reverse & 0.0038 & 0.1679 & -0.1858 & 0.07048 & 0.07045 & 0.00198 & 0\\
\bottomrule
\end{tabular}
\end{table}
\begin{table}[H]
\centering\small
\caption{Paired directional contrasts from $q_0=0$. Predictions for the normalized ordinary and removed contrasts are $B_\tau^{\to}-B_\tau^{\leftarrow}=-.075018$; the delayed contrast has leading prediction zero.}\label{tab:nonoraclecontrasts}
\begin{tabular}{rlrrrrr}
\toprule
$n$ & Contrast & Mean ($10^{-3}$) & MCSE ($10^{-3}$) & Mean / $\bar\eta_n$ & MCSE / $\bar\eta_n$ & Finite-$n$ / $\bar\eta_n$\\
\midrule
3,000 & $D_{\rm ord}$ & -2.5638 & 0.4733 & -0.12127 & 0.02239 & -0.09119\\
3,000 & $D_{\rm del}$ & -0.6915 & 0.4698 & -0.03271 & 0.02222 & 0.00000\\
3,000 & $D_{\rm ord}-D_{\rm del}$ & -1.8723 & 0.1491 & -0.08856 & 0.00705 & -0.09119\\
12,000 & $D_{\rm ord}$ & -0.6848 & 0.1899 & -0.08208 & 0.02276 & -0.08932\\
12,000 & $D_{\rm del}$ & 0.0994 & 0.1883 & 0.01192 & 0.02257 & 0.00000\\
12,000 & $D_{\rm ord}-D_{\rm del}$ & -0.7842 & 0.0451 & -0.09399 & 0.00541 & -0.08932\\
48,000 & $D_{\rm ord}$ & -0.3342 & 0.0740 & -0.10545 & 0.02336 & -0.08695\\
48,000 & $D_{\rm del}$ & -0.0688 & 0.0743 & -0.02171 & 0.02346 & 0.00000\\
48,000 & $D_{\rm ord}-D_{\rm del}$ & -0.2654 & 0.0135 & -0.08375 & 0.00426 & -0.08695\\
\bottomrule
\end{tabular}
\end{table}

At $n=48{,}000$, paired removal is $-.0002654$ (MCSE $.0000135$), close to the oracle-initialized $-.0002624$ (MCSE $.0000133$). The finite-$n$ recursion includes the initialization transient and matches every bias and contrast in Tables~\ref{tab:nonoraclefull}--\ref{tab:nonoraclecontrasts} within 1.5 MCSE. Variance and MSE changes are small, projection is inactive, and the interaction-removal effect persists from $q_0=0$.
\raggedbottom
\subsection{Four-innovation delay-profile experiment}\label{app:profileexperiment}
To test the partial-delay part of Theorem~\ref{thm:delay}, let
\begin{equation}\label{eq:profilemodel}
X_t=V_t+0.6V_{t-1}^2+0.4V_{t-2}^2+0.2V_{t-3}^2+0.7Z_t,
\end{equation}
where $V_t$ are i.i.d. uniform on $\{-1,0,1\}$, $Z_t$ are i.i.d. standard normal, and the two sequences are independent. The innovation memory is $r=4$. We fix $\tau=.25$, $c=3$, $\gamma=.75$, $t_0=20$, projection to $[-6,6]$, and $q_0=q^\star$ to isolate the mechanism. Every stream is processed in both directions with all delays $d=1,\ldots,5$.

\paragraph{Constants independent of the SGD outcomes.}
The marginal CDF has an 81-term Gaussian-mixture representation, one equally weighted term for each innovation tuple. For lag $k=1,2,3$, summing over $3^{4+k}$ innovation patterns gives $J_k$ and its two boundary derivatives. The independent Gaussian noises contribute products of univariate Gaussian CDFs or densities. Differentiated sums agree with central finite differences within $1.1\times10^{-12}$. We obtain
\[
q^\star=-.016206525,\quad h=.274918969,\quad f'(q^\star)=.118899317,\quad \Omega=.180635585.
\]
For $d=1,\ldots,5$, the predicted directional coefficients in \eqref{eq:profilereversal} are
$(-.023646,-.011595,-.003932,0,0)$.
The common curvature coefficient is $-.071042$, and the second-moment limit is $\Omega/(2h)=.328525$.

The model satisfies the stated assumptions in both directions. In the reverse direction a causal representation is
$V_{t-3}+.6V_{t-2}^2+.4V_{t-1}^2+.2V_t^2+.7Z_t$, after reindexing independent innovations in distribution. Conditional on the past, each observation is a finite mixture of Gaussians with variance $.49$. Its density and derivatives are uniformly bounded, and the marginal density is positive on the compact projection interval. All constants were computed before the SGD experiments.

\paragraph{Monte Carlo precision and pairing.}
Each sample size uses 40,000 independent streams, with delays and directions paired within streams. An independent 400-stream pilot at $n=48{,}000$ informed the repeat count through the variances of normalized adjacent directional contrasts, subject to a prespecified cap. The pilot is excluded from reported performance. Separate seeds for $n=3{,}000,12{,}000,48{,}000$ give 120,000 independent streams and 1.2 million recursion trajectories. At $n=48{,}000$, the four primary 95\% Monte Carlo halfwidths are $.00185$--$.00186$, below the $.002$ precision target. The reproducibility protocol records the complete planning calculation.

Write $D_d=\bar q_n^{(d),\to}-\bar q_n^{(d),\leftarrow}$. Figure~\ref{fig:profile} displays $(D_d-D_4)/\bar\eta_n$ and $(D_d-D_{d+1})/\bar\eta_n$. Subtracting $D_4$ retains the leading directional coefficient, because its coefficient is zero, and reduces Monte Carlo variability. The $d=4$ entry in the first panel is identically zero by construction; the $d=5$ entry is a separate sufficient-delay comparison. Error bars quantify pointwise Monte Carlo uncertainty across independent streams.

At $n=48{,}000$, the adjacent directional effects for $d=1,2,3$ are $-.01381$, $-.00812$ and $-.00401$, each with MCSE about $.00095$. Their predictions are $-.01205$, $-.00766$ and $-.00393$. The sufficient-delay contrast $d=4$ versus $d=5$ is $.00007$ (MCSE $.00095$). These comparisons confirm the lag-specific mechanism. The cumulative $d=1$ versus $d=4$ effect, $-.02594$ (MCSE $.00113$), differs from the leading value $-.02365$ but agrees with the finite-$n$ value $-.02673$. Table~\ref{tab:profilecontrasts} retains every sample size and contrast.

\begin{table}[H]
\centering\small
\caption{Paired directional contrasts divided by $\bar\eta_n$, with Monte Carlo standard errors and leading predictions. All rows use 40,000 independent streams. The $d=4$ versus $4$ entry is identically zero.}\label{tab:profilecontrasts}
\begin{tabular}{rrrrrrrrrr}
\toprule
$n$ & $d$ & $d$ vs. $4$ & MCSE & Leading & Finite-$n$ & $d$ vs. $d+1$ & MCSE & Leading & Finite-$n$\\
\midrule
3,000 & 1 & -0.02722 & 0.00185 & -0.02365 & -0.02784 & -0.01316 & 0.00155 & -0.01205 & -0.01418\\
3,000 & 2 & -0.01407 & 0.00175 & -0.01160 & -0.01365 & -0.00932 & 0.00157 & -0.00766 & -0.00902\\
3,000 & 3 & -0.00475 & 0.00156 & -0.00393 & -0.00463 & -0.00475 & 0.00156 & -0.00393 & -0.00463\\
3,000 & 4 & 0.00000 & 0.00000 & 0.00000 & 0.00000 & 0.00082 & 0.00156 & 0.00000 & 0.00000\\
3,000 & 5 & -0.00082 & 0.00156 & 0.00000 & 0.00000 & -- & -- & -- & --\\
\midrule
12,000 & 1 & -0.02774 & 0.00144 & -0.02365 & -0.02734 & -0.01396 & 0.00120 & -0.01205 & -0.01393\\
12,000 & 2 & -0.01378 & 0.00135 & -0.01160 & -0.01341 & -0.00825 & 0.00119 & -0.00766 & -0.00886\\
12,000 & 3 & -0.00553 & 0.00120 & -0.00393 & -0.00455 & -0.00553 & 0.00120 & -0.00393 & -0.00455\\
12,000 & 4 & 0.00000 & 0.00000 & 0.00000 & 0.00000 & 0.00063 & 0.00120 & 0.00000 & 0.00000\\
12,000 & 5 & -0.00063 & 0.00120 & 0.00000 & 0.00000 & -- & -- & -- & --\\
\midrule
48,000 & 1 & -0.02594 & 0.00113 & -0.02365 & -0.02673 & -0.01381 & 0.00094 & -0.01205 & -0.01362\\
48,000 & 2 & -0.01213 & 0.00106 & -0.01160 & -0.01311 & -0.00812 & 0.00095 & -0.00766 & -0.00866\\
48,000 & 3 & -0.00401 & 0.00095 & -0.00393 & -0.00445 & -0.00401 & 0.00095 & -0.00393 & -0.00445\\
48,000 & 4 & 0.00000 & 0.00000 & 0.00000 & 0.00000 & 0.00007 & 0.00095 & 0.00000 & 0.00000\\
48,000 & 5 & -0.00007 & 0.00095 & 0.00000 & 0.00000 & -- & -- & -- & --\\
\bottomrule
\end{tabular}
\end{table}

\paragraph{Single-direction lag contributions.}
Table~\ref{tab:profilesingle} tests \eqref{eq:adjacentdelay} separately in both directions using the same paired records. At $n=48{,}000$, the reverse $d=1$ versus $d=2$ contrast is $.01467$ (MCSE $.00067$), against a leading value $.01251$ and a finite-$n$ value $.01414$. The forward first-lag prediction is positive, $.00046$, whereas its next two predictions are negative. Keeping the two directions visible distinguishes removal of the complete interaction from removal of its directional difference. 
\begin{table}[H]
\centering\small
\caption{Single-direction paired adjacent-delay effects $(\bar q_n^{(d)}-\bar q_n^{(d+1)})/\bar\eta_n$. MCSEs are computed from within-stream differences; all entries use 40,000 streams. The $d=4$ row compares two sufficient delays.}\label{tab:profilesingle}
\begin{tabular}{rrrrrrrrrr}
\toprule
$n$ & $d$ & Forward & MCSE & Leading & Finite-$n$ & Reverse & MCSE & Leading & Finite-$n$\\
\midrule
3,000 & 1 & 0.00539 & 0.00109 & 0.00046 & 0.00054 & 0.01854 & 0.00110 & 0.01251 & 0.01472\\
3,000 & 2 & -0.00204 & 0.00110 & -0.00214 & -0.00252 & 0.00728 & 0.00111 & 0.00553 & 0.00650\\
3,000 & 3 & -0.00175 & 0.00110 & -0.00276 & -0.00326 & 0.00300 & 0.00111 & 0.00117 & 0.00137\\
3,000 & 4 & 0.00247 & 0.00111 & 0.00000 & -0.00001 & 0.00165 & 0.00111 & 0.00000 & -0.00001\\
\midrule
12,000 & 1 & 0.00244 & 0.00084 & 0.00046 & 0.00053 & 0.01640 & 0.00086 & 0.01251 & 0.01446\\
12,000 & 2 & -0.00098 & 0.00084 & -0.00214 & -0.00248 & 0.00727 & 0.00085 & 0.00553 & 0.00639\\
12,000 & 3 & -0.00318 & 0.00085 & -0.00276 & -0.00320 & 0.00235 & 0.00085 & 0.00117 & 0.00135\\
12,000 & 4 & 0.00226 & 0.00085 & 0.00000 & -0.00001 & 0.00163 & 0.00085 & 0.00000 & -0.00001\\
\midrule
48,000 & 1 & 0.00086 & 0.00066 & 0.00046 & 0.00052 & 0.01467 & 0.00067 & 0.01251 & 0.01414\\
48,000 & 2 & -0.00082 & 0.00067 & -0.00214 & -0.00242 & 0.00730 & 0.00067 & 0.00553 & 0.00625\\
48,000 & 3 & -0.00235 & 0.00067 & -0.00276 & -0.00312 & 0.00166 & 0.00067 & 0.00117 & 0.00132\\
48,000 & 4 & 0.00080 & 0.00067 & 0.00000 & -0.00000 & 0.00073 & 0.00067 & 0.00000 & -0.00000\\
\bottomrule
\end{tabular}
\end{table}

\paragraph{Second moments and estimation loss.}
Table~\ref{tab:profilemoments} reports both directions' signed means, second moments, variance and MSE for every delay. At $n=48{,}000$, terminal second moments divided by $\eta_n$ range from $.3363$ to $.3409$, close to the finite-$n$ value $.3390$ and above the limit $.3285$; their small variation across delays reflects the retained full $\Omega$. Absolute bias and MSE vary non-monotonically with delay. The predicted forward coefficient also becomes slightly more negative from $d=1$ to $d=2$, reflecting the signed lag contributions.

The supplementary material contains the pilot records, the experimental protocol, the mixture constants and summaries of all 600,000 paired-delay records, which the supplied code regenerates from the recorded seeds. It also supplies unadjusted directional contrasts and single-direction adjacent contrasts. Independent full-history recursions verify the delayed updates, including tied observations and active projection boundaries.

\begin{table}[H]
\centering\small
\caption{All delay-profile estimation results. Means are normalized by $\bar\eta_n$ and terminal second moments by $\eta_n$; MCSEs use the same normalizations. The final two columns concern the averaged estimate and are multiplied by $10^5$. The common terminal second-moment limit is $.328525$.}\label{tab:profilemoments}
\begin{tabular}{rrlrrrrrrrr}
\toprule
$n$ & $d$ & Direction & Bias / $\bar\eta_n$ & MCSE & Finite-$n$ & $\E e_n^2/\eta_n$ & MCSE & Finite-$n$ & Variance & MSE\\
\midrule
3,000 & 1 & forward & -0.1012 & 0.0070 & -0.0957 & 0.3521 & 0.0025 & 0.3504 & 86.8382 & 87.2935\\
3,000 & 1 & reverse & -0.0741 & 0.0070 & -0.0678 & 0.3533 & 0.0025 & 0.3504 & 87.1068 & 87.3502\\
3,000 & 2 & forward & -0.1066 & 0.0070 & -0.0962 & 0.3530 & 0.0025 & 0.3504 & 87.1666 & 87.6720\\
3,000 & 2 & reverse & -0.0927 & 0.0070 & -0.0826 & 0.3542 & 0.0025 & 0.3504 & 86.9796 & 87.3612\\
3,000 & 3 & forward & -0.1045 & 0.0070 & -0.0937 & 0.3539 & 0.0025 & 0.3504 & 87.3839 & 87.8700\\
3,000 & 3 & reverse & -0.0999 & 0.0070 & -0.0891 & 0.3546 & 0.0025 & 0.3504 & 86.9592 & 87.4035\\
3,000 & 4 & forward & -0.1028 & 0.0070 & -0.0904 & 0.3548 & 0.0025 & 0.3504 & 87.6240 & 88.0939\\
3,000 & 4 & reverse & -0.1029 & 0.0070 & -0.0904 & 0.3554 & 0.0025 & 0.3504 & 87.0531 & 87.5246\\
3,000 & 5 & forward & -0.1052 & 0.0070 & -0.0904 & 0.3553 & 0.0025 & 0.3504 & 87.4156 & 87.9084\\
3,000 & 5 & reverse & -0.1046 & 0.0070 & -0.0904 & 0.3561 & 0.0025 & 0.3504 & 87.4196 & 87.9064\\
12,000 & 1 & forward & -0.0761 & 0.0087 & -0.0933 & 0.3457 & 0.0025 & 0.3437 & 21.2760 & 21.3157\\
12,000 & 1 & reverse & -0.0524 & 0.0088 & -0.0659 & 0.3430 & 0.0024 & 0.3437 & 21.3200 & 21.3386\\
12,000 & 2 & forward & -0.0785 & 0.0087 & -0.0938 & 0.3459 & 0.0025 & 0.3437 & 21.2307 & 21.2731\\
12,000 & 2 & reverse & -0.0688 & 0.0087 & -0.0804 & 0.3432 & 0.0024 & 0.3437 & 21.3131 & 21.3455\\
12,000 & 3 & forward & -0.0775 & 0.0087 & -0.0913 & 0.3464 & 0.0025 & 0.3437 & 21.2291 & 21.2704\\
12,000 & 3 & reverse & -0.0760 & 0.0087 & -0.0868 & 0.3433 & 0.0024 & 0.3437 & 21.3140 & 21.3537\\
12,000 & 4 & forward & -0.0744 & 0.0087 & -0.0881 & 0.3466 & 0.0025 & 0.3437 & 21.2420 & 21.2800\\
12,000 & 4 & reverse & -0.0784 & 0.0087 & -0.0881 & 0.3436 & 0.0024 & 0.3437 & 21.2886 & 21.3308\\
12,000 & 5 & forward & -0.0766 & 0.0087 & -0.0881 & 0.3466 & 0.0025 & 0.3437 & 21.2942 & 21.3345\\
12,000 & 5 & reverse & -0.0800 & 0.0087 & -0.0881 & 0.3436 & 0.0024 & 0.3437 & 21.3117 & 21.3558\\
48,000 & 1 & forward & -0.0879 & 0.0114 & -0.0903 & 0.3363 & 0.0024 & 0.3390 & 5.1836 & 5.1912\\
48,000 & 1 & reverse & -0.0561 & 0.0113 & -0.0636 & 0.3405 & 0.0024 & 0.3390 & 5.1562 & 5.1592\\
48,000 & 2 & forward & -0.0888 & 0.0114 & -0.0908 & 0.3364 & 0.0024 & 0.3390 & 5.1807 & 5.1885\\
48,000 & 2 & reverse & -0.0708 & 0.0113 & -0.0777 & 0.3407 & 0.0024 & 0.3390 & 5.1579 & 5.1628\\
48,000 & 3 & forward & -0.0880 & 0.0114 & -0.0884 & 0.3365 & 0.0024 & 0.3390 & 5.1811 & 5.1887\\
48,000 & 3 & reverse & -0.0781 & 0.0113 & -0.0840 & 0.3407 & 0.0024 & 0.3390 & 5.1609 & 5.1669\\
48,000 & 4 & forward & -0.0856 & 0.0114 & -0.0853 & 0.3365 & 0.0024 & 0.3390 & 5.1884 & 5.1956\\
48,000 & 4 & reverse & -0.0798 & 0.0113 & -0.0853 & 0.3407 & 0.0024 & 0.3390 & 5.1608 & 5.1671\\
48,000 & 5 & forward & -0.0864 & 0.0114 & -0.0853 & 0.3367 & 0.0024 & 0.3390 & 5.1844 & 5.1918\\
48,000 & 5 & reverse & -0.0805 & 0.0113 & -0.0853 & 0.3409 & 0.0024 & 0.3390 & 5.1622 & 5.1686\\
\bottomrule
\end{tabular}
\end{table}
\section{Interval experiments}\label{app:revisionexperiments}
This appendix reports the interval studies behind Section~\ref{sec:experiments}: block lengths, the variance floor, complete hull validation, window sensitivity, covariance tails, negative dependence and conditional scale, the self-normalized and online comparisons, and the JPY/USD example.

\subsection{The 600-stream block-bootstrap comparison}\label{app:block600}
\begin{table}[H]
\centering\small
\caption{Coverage (\%) / mean length of nominal 95\% intervals on 600 common streams per row. CBB$_j$ uses circular blocks of length $j\lceil n^{1/3}\rceil$ and 399 resamples. Coverage MCSE is about .89 percentage point near 95\%.}\label{tab:block600}
\begin{tabular}{llccccc}
\toprule
$n$ & Process & E-hull & S-hull & CBB$_1$ & CBB$_2$ & CBB$_4$\\
\midrule
3k & i.i.d. & 95.67 / 0.1956 & 95.00 / 0.1964 & 94.67 / 0.1916 & 94.67 / 0.1903 & 94.00 / 0.1898\\
3k & MA(1) & 95.67 / 0.2491 & 95.33 / 0.2503 & 94.17 / 0.2411 & 94.33 / 0.2411 & 94.33 / 0.2408\\
3k & MA(11) & 94.50 / 0.5562 & 94.83 / 0.5587 & 91.83 / 0.4965 & 93.17 / 0.5330 & 93.83 / 0.5453\\
12k & MA(11) & 96.83 / 0.2829 & 97.17 / 0.2840 & 94.17 / 0.2566 & 95.50 / 0.2671 & 95.83 / 0.2722\\
3k & AR(.8) & 91.67 / 0.4458 & 91.50 / 0.4480 & 87.33 / 0.4037 & 89.67 / 0.4353 & 91.00 / 0.4489\\
\bottomrule
\end{tabular}
\end{table}

Table~\ref{tab:block600} reports the full prespecified block-length grid for circular block percentile bootstrap (CBB). At MA(11), $n=3{,}000$, longer blocks raise CBB coverage from 91.83\% to 93.83\%, compared with 94.50\% for E-hull. The E-hull--CBB$_4$ difference is .67 percentage point (paired MCSE .58), with mean length larger by .01084 (MCSE .00177). At $n=12{,}000$, E-hull covers 96.83\% in this sample, 94.44\% in the independent 1,600-stream sample of Table~\ref{tab:hullfull} and 96.12\% in that of Table~\ref{tab:block}. E-hull and S-hull have similar coverage, with E-hull slightly shorter throughout the matched comparison.

The AR(.8) cell has lower coverage for both E-hull (91.67\%) and CBB$_4$ (91.00\%); their paired difference is .67 point (MCSE .75). On 1,600 independent streams, widening the E-hull window from 8 to 15 raises coverage from 93.69\% to 94.81\% and increases length (Table~\ref{tab:arwindow}).

\subsection{Variance-floor control}\label{app:floor}
\begin{figure}[H]
\centering\includegraphics[width=.62\textwidth]{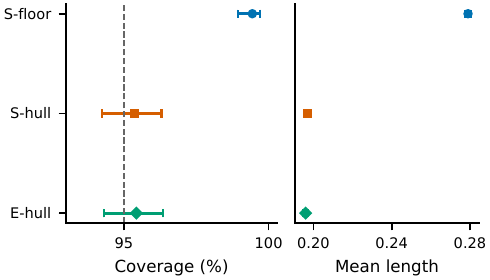}
\caption{Variance-floor control on 1,600 i.i.d. Cauchy streams, $n=3{,}000$, with the power window. S-floor and S-hull differ in the floor and acceptance rule; E-hull also uses equal weights. Coverage bars are Wilson 95\% Monte Carlo intervals; length bars are mean $\pm1.96$ MCSE.}\label{fig:floorremoval}
\end{figure}

The connectedness floor is active at the target on all 1,600 i.i.d. streams at $n=3{,}000$, giving S-floor coverage of 99.44\%. With the same symmetric weights, S-hull covers 95.38\% with 29.4\% shorter intervals (Figure~\ref{fig:floorremoval}); E-hull covers 95.44\%. All 12,800 S-hull acceptance sets in the validation sample are connected. Negative MA(1) and conditional-scale experiments show the same floor effect; E-hull covers 94.75\% in both cells (Table~\ref{tab:edgefull}). Across six finite-memory cells, median S-hull length is .977--1.003 times the first-order width in \eqref{eq:endpointsexpansion}.

\subsection{Complete hull validation}
S-hull and E-hull use symmetric and equal weights with the power window and $1/n$ floor. S-floor uses the same symmetric power-window configuration with the larger connectedness floor. Original-filter and logarithmic-window variants are in the artifact; Table~\ref{tab:fixedfull} isolates the window.
Table~\ref{tab:hullfull} reports the independent 1,600-stream study by process and sample size. ``Lower'' is the event that the interval is below the target ($U<q^\star$); ``Upper'' means that it is above the target ($L>q^\star$). ``Floor'' records whether the candidate variance at the truth is below the procedural floor. ``Multi'' is the proportion of raw acceptance sets with multiple components. Every interval in this study is finite; coverage denominators include all replications.
{\renewcommand{\arraystretch}{.92}%
\begingroup\setlength{\tabcolsep}{3pt}
\begin{longtable}{llrrrrrrrr}
\caption{Independent acceptance-hull confirmation. Each setting uses 1,600 streams; MA($r-1$) is the equal-coefficient moving average with $r$ innovations. MCSE is for total coverage. Coverage, misses, floor activation and multiple-component frequencies are percentages.}\label{tab:hullfull}\\
\toprule
Process, $n$ & Method & Cover & MCSE & Lower & Upper & Length & Floor & Multi & Max comp.\\
\midrule\endfirsthead
\toprule Process, $n$ & Method & Cover & MCSE & Lower & Upper & Length & Floor & Multi & Max comp.\\\midrule\endhead
i.i.d., 3k & E-hull & 95.44 & 0.52 & 2.63 & 1.94 & 0.1960 & 0.0 & 0.0 & 1\\
i.i.d., 3k & S-floor & 99.44 & 0.19 & 0.44 & 0.13 & 0.2790 & 100.0 & 0.0 & 1\\
i.i.d., 3k & S-hull & 95.38 & 0.53 & 2.44 & 2.19 & 0.1968 & 0.0 & 0.0 & 1\\
MA(1), 3k & E-hull & 95.44 & 0.52 & 2.25 & 2.31 & 0.2496 & 0.0 & 0.0 & 1\\
MA(1), 3k & S-floor & 97.56 & 0.39 & 1.19 & 1.25 & 0.2793 & 98.9 & 0.0 & 1\\
MA(1), 3k & S-hull & 95.31 & 0.53 & 2.13 & 2.56 & 0.2508 & 0.0 & 0.0 & 1\\
MA(3), 3k & E-hull & 95.19 & 0.54 & 1.81 & 3.00 & 0.3369 & 0.0 & 0.0 & 1\\
MA(3), 3k & S-floor & 95.31 & 0.53 & 1.81 & 2.88 & 0.3383 & 0.1 & 0.0 & 1\\
MA(3), 3k & S-hull & 95.31 & 0.53 & 1.81 & 2.88 & 0.3383 & 0.0 & 0.0 & 1\\
MA(11), 3k & E-hull & 94.44 & 0.57 & 2.06 & 3.50 & 0.5577 & 0.0 & 0.0 & 1\\
MA(11), 3k & S-floor & 94.13 & 0.59 & 1.81 & 4.06 & 0.5608 & 0.0 & 0.0 & 1\\
MA(11), 3k & S-hull & 94.13 & 0.59 & 1.81 & 4.06 & 0.5608 & 0.0 & 0.0 & 1\\
MA(1), 12k & E-hull & 94.44 & 0.57 & 2.25 & 3.31 & 0.1240 & 0.0 & 0.0 & 1\\
MA(1), 12k & S-floor & 94.38 & 0.58 & 2.44 & 3.19 & 0.1244 & 0.0 & 0.0 & 1\\
MA(1), 12k & S-hull & 94.38 & 0.58 & 2.44 & 3.19 & 0.1244 & 0.0 & 0.0 & 1\\
MA(11), 12k & E-hull & 94.44 & 0.57 & 2.63 & 2.94 & 0.2837 & 0.0 & 0.0 & 1\\
MA(11), 12k & S-floor & 94.31 & 0.58 & 2.50 & 3.19 & 0.2846 & 0.0 & 0.0 & 1\\
MA(11), 12k & S-hull & 94.31 & 0.58 & 2.50 & 3.19 & 0.2846 & 0.0 & 0.0 & 1\\
AR(.8), 3k & E-hull & 92.94 & 0.64 & 2.94 & 4.13 & 0.4496 & 0.0 & 0.0 & 1\\
AR(.8), 3k & S-floor & 93.44 & 0.62 & 2.88 & 3.69 & 0.4515 & 0.0 & 0.0 & 1\\
AR(.8), 3k & S-hull & 93.44 & 0.62 & 2.88 & 3.69 & 0.4515 & 0.0 & 0.0 & 1\\
AR(.8), 12k & E-hull & 94.31 & 0.58 & 2.81 & 2.88 & 0.2298 & 0.0 & 0.0 & 1\\
AR(.8), 12k & S-floor & 94.19 & 0.59 & 2.81 & 3.00 & 0.2306 & 0.0 & 0.0 & 1\\
AR(.8), 12k & S-hull & 94.19 & 0.59 & 2.81 & 3.00 & 0.2306 & 0.0 & 0.0 & 1\\
\bottomrule\end{longtable}\endgroup
}

\subsection{A fixed sufficient window for MA(11)}
We use the same equal-coefficient twelve-innovation Gaussian process, Cauchy marginal transform, and $\tau=.25$ as in the hull-validation cells at $n=3{,}000$ and $12{,}000$. Each sample size has 1,600 streams. The weights are symmetric finite-filter or equal weights; the windows are $\lceil\sqrt{\log n}\rceil$, $\lceil n^{1/4}\rceil$, and the fixed sufficient $b=11$. Every configuration uses the $1/n$ floor. The common floor isolates the effect of window choice.

The paired streams use seeds $103092100+100003j$, with $j=3,5$ for the two sample sizes. At $n=12{,}000$, the power and fixed windows both equal 11. Covariance tails use the quadrature of Appendix~\ref{app:tails}.

\begin{table}[H]
\centering\small
\caption{Fixed-window control: coverage, directional misses and coverage MCSE in percentage points; length and its MCSE on the observation scale. Each row uses 1,600 streams, paired across configurations at the same $n$. All intervals are finite, and no true-candidate nonpositive variance or small-floor activation occurs.}\label{tab:fixedfull}
\begin{tabular}{rlrrrrrrr}
\toprule
$n$ & Weights / rule & $b$ & Cover & MCSE & Length & MCSE & Below & Above\\
\midrule
3,000 & Equal / Fixed & 11 & 95.19 & 0.535 & 0.57949 & 0.00226 & 1.56 & 3.25\\
3,000 & Equal / Log & 3 & 85.31 & 0.885 & 0.42890 & 0.00161 & 6.38 & 8.31\\
3,000 & Equal / Power & 8 & 94.44 & 0.573 & 0.55772 & 0.00213 & 2.06 & 3.50\\
3,000 & Sym / Fixed & 11 & 95.13 & 0.539 & 0.58293 & 0.00230 & 1.50 & 3.38\\
3,000 & Sym / Log & 3 & 85.19 & 0.888 & 0.43103 & 0.00163 & 6.56 & 8.25\\
3,000 & Sym / Power & 8 & 94.13 & 0.588 & 0.56077 & 0.00216 & 1.81 & 4.06\\
12,000 & Equal / Fixed & 11 & 94.44 & 0.573 & 0.28374 & 0.00057 & 2.63 & 2.94\\
12,000 & Equal / Log & 4 & 89.56 & 0.765 & 0.23075 & 0.00047 & 4.94 & 5.50\\
12,000 & Equal / Power & 11 & 94.44 & 0.573 & 0.28374 & 0.00057 & 2.63 & 2.94\\
12,000 & Sym / Fixed & 11 & 94.31 & 0.579 & 0.28458 & 0.00058 & 2.50 & 3.19\\
12,000 & Sym / Log & 4 & 89.75 & 0.758 & 0.23151 & 0.00047 & 4.56 & 5.69\\
12,000 & Sym / Power & 11 & 94.31 & 0.579 & 0.28458 & 0.00058 & 2.50 & 3.19\\
\bottomrule
\end{tabular}
\end{table}

At $n=3{,}000$, completing the symmetric window from $b=8$ to $b=11$ increases coverage by 1.00 percentage point (paired MCSE .249) and average length by .02216 (paired MCSE .00029). Equal weights show the same tradeoff, with a .75-point increase (MCSE .216) and a .02178 length increase. The severely truncated logarithmic windows under-cover at both sample sizes. The controls quantify the coverage and length changes from including the omitted covariance.

\subsection{Sensitivity to the AR lag window}\label{app:arwindow}
To examine the effect of dependence beyond the default window, we use 1,600 independent stationary AR(.8) streams of length $n=3{,}000$. The latent process is $Y_t=.8Y_{t-1}+.6\varepsilon_t$, with independent standard normal innovations and $Y_0\sim N(0,1)$. We set $\tau=.25$ and use E-hull with floor $1/n$ at $b\in\{8,15,30,60\}$. These four configurations share streams, generated with seed 27271031, independently of the samples in Tables~\ref{tab:block600}, \ref{tab:hullfull} and \ref{tab:block}. Lengths use the Cauchy transformation; coverage is unchanged by this monotone map.

\begin{table}[H]
\centering\small
\caption{AR(.8) window sensitivity on 1,600 independent streams, $n=3{,}000$ and $\tau=.25$. E-hull uses floor $1/n$ and Cauchy margins. Parentheses report MCSEs; coverage differences from $b=8$ are paired and measured in percentage points. Omit is the percentage of finite-sample score variance excluded by the window.}\label{tab:arwindow}
\begin{tabular}{rrrrr}
\toprule
$b$ & Coverage (\%) & Mean length & Coverage change (pp) & Omit (\%)\\
\midrule
8 & 93.69 (0.61) & 0.4516 (0.0015) & --- & 12.35\\
15 & 94.81 (0.55) & 0.4782 (0.0017) & 1.125 (0.264) & 2.55\\
30 & 95.19 (0.54) & 0.4902 (0.0019) & 1.500 (0.373) & 0.0889\\
60 & 95.56 (0.51) & 0.5032 (0.0024) & 1.875 (0.421) & 0.0001\\
\bottomrule
\end{tabular}
\end{table}

Increasing $b$ from 8 to 15 raises coverage by 1.125 percentage points (paired MCSE .264), with mean length increasing from .4516 to .4782. All intervals are finite. Larger windows further increase coverage and length in this experiment.

The omitted-covariance calculation gives a corresponding explanation. For $z_\tau=\Phi^{-1}(\tau)$, the latent AR target-score covariance at lag $k$ is
\[
\Gamma_k=\int_0^{.8^k}\frac{\exp\{-z_\tau^2/(1+u)\}}{2\pi\sqrt{1-u^2}}\,\dd u.
\]
At $n=3{,}000$, $V_n=\Gamma_0+2\sum_{k=1}^{n-1}(1-k/n)\Gamma_k\simeq1.11205$. The fraction omitted by the window decreases from 12.35\% at $b=8$ to 2.55\% at $b=15$. Numerical integration through lag 249 makes the remaining geometric tail negligible at the displayed precision. These fixed-window comparisons quantify the sensitivity of coverage to covariance truncation; each window is evaluated on the same streams.

\subsection{Covariance tails and weight gains}\label{app:tails}
For a Gaussian moving average, $\Gamma_k=\Phi_2(z_\tau,z_\tau;\rho_k)-\tau^2$, with $\rho_k=(r-k)/r$ and $\Gamma_0=\tau(1-\tau)$. One-dimensional conditional-normal quadrature computes $\Phi_2$ to absolute tolerance $10^{-12}$. These probabilities and the exact finite filter determine both the omitted covariance \eqref{eq:tail} and the reversal-projection variance reduction without Monte Carlo error. For the twelve-innovation example, the relative variance reduction is .013919 at $n=3{,}000$ and .009674 at $n=12{,}000$. These reductions compare symmetric weights with the original finite filter at a common sum normalization.

\subsection{Negative dependence and conditional scale}\label{app:finalvalidation}
The negative-dependence and conditional-scale study uses two prespecified cells with $n=3{,}000$ and 1,600 independent streams each. Negative MA(1) uses $Y_t=(U_t-.5U_{t-1})/\sqrt{1.25}$, Cauchy transform $X_t=\tan\{\pi(\Phi(Y_t)-1/2)\}$, and $\tau=.5$. The scale mixture uses $X_t=(.5+.5\Phi(U_{t-1}))U_t$ and $\tau=.75$. Here $U_t$ are i.i.d. standard normal. Both have innovation memory two. The scale-mixture quantile solves $2\int_{.5}^1\Phi(q/s)\,\dd s=\tau$, using quadrature and a scalar root solver. Seeds are 1726092201 and 1726092202, respectively. All method settings were specified before sampling.

All four methods receive the same stream and use $b=8=\lceil n^{1/4}\rceil$. The filter parameters are $u_0=1,c=3,\gamma=.75,t_0=20$. S-hull, E-hull and Filter-hull use the $1/n$ floor; S-floor uses the connectedness floor with factor $.5$. There is no pilot or true-quantile initialization for these intervals. Table~\ref{tab:edgefull} reports every method. Below means $U_n<q^\star$ and above means $L_n>q^\star$. Floor and $Q\le0$ refer to the candidate variance evaluated at the truth. All 12,800 intervals are nonempty and finite, and every raw acceptance set in these two cells is connected. Small-floor activation and nonpositive true-candidate variance occur zero times. 

\begin{table}[H]
\centering\small
\caption{Negative-dependence and conditional-scale results. Coverage, directional misses, floor activation, and nonpositive variance are percentages. MCSE is the sample standard deviation across 1,600 streams divided by $\sqrt{1600}$. Length is on the observation scale.}\label{tab:edgefull}
\begin{tabular}{llrrrrrrrr}
\toprule
Model & Method & Cover & MCSE & Below & Above & Length & MCSE & Floor & $Q\le0$\\
\midrule
Negative MA(1) & E-hull & 94.750 & 0.558 & 2.562 & 2.688 & 0.07828 & 0.00024 & 0.0 & 0.0\\
Negative MA(1) & Filter-hull & 95.062 & 0.542 & 2.438 & 2.500 & 0.07912 & 0.00025 & 0.0 & 0.0\\
Negative MA(1) & S-floor & 99.938 & 0.062 & 0.062 & 0.000 & 0.13882 & 0.00030 & 100.0 & 0.0\\
Negative MA(1) & S-hull & 94.875 & 0.551 & 2.375 & 2.750 & 0.07854 & 0.00025 & 0.0 & 0.0\\
Scale mixture & E-hull & 94.750 & 0.558 & 3.000 & 2.250 & 0.07725 & 0.00021 & 0.0 & 0.0\\
Scale mixture & Filter-hull & 94.875 & 0.551 & 2.875 & 2.250 & 0.07818 & 0.00022 & 0.0 & 0.0\\
Scale mixture & S-floor & 98.875 & 0.264 & 0.625 & 0.500 & 0.10287 & 0.00021 & 100.0 & 0.0\\
Scale mixture & S-hull & 94.812 & 0.555 & 3.125 & 2.062 & 0.07755 & 0.00021 & 0.0 & 0.0\\
\bottomrule
\end{tabular}
\end{table}

\subsection{Self-normalization, data-selected windows and an online pivot}\label{app:v3intervals}
\subsubsection{Design}
This prespecified study uses 1,600 new streams in each of six cells: i.i.d., MA(1) ($r=2$) and MA(11) ($r=12$) at $n=3{,}000$, MA(11) at $n=12{,}000$, and AR(.8) at both sample sizes, all with Cauchy margins, $\tau=.25$ and nominal level 95\%. Every method uses the same stream. E-hull uses equal weights and the $1/n$ floor with three windows: $\lceil n^{1/4}\rceil$, the data-selected $\hat b=\min(2\hat m,\lceil n^{1/3}\rceil)$ of Section~\ref{sec:method}, and the oracle $b=r-1$ for finite-memory cells. Here $\hat m$ is the smallest $m\ge0$ such that the next $\max(5,\lceil(\log_{10}n)^{1/2}\rceil)$ sample autocorrelations of $\ind(X_t\le\widehat q_\tau)$, at the sample $\tau$-quantile $\widehat q_\tau$, are below $2(\log_{10}n/n)^{1/2}$ in absolute value \citep{politis2003adaptive}. E-hull with the deterministic cap $\lceil n^{1/3}\rceil$ is computed on the same streams, regenerated from their seeds. SN is the interval $\{\theta:n(\widehat q_n-\theta)^2\le45.4\,\mathcal D_n\}$ of \citet{shao2010self}, with recursive sample quantiles $\widehat q_t$ (the $\lceil t\tau\rceil$th order statistic of $X_1,\ldots,X_t$), $\mathcal D_n=n^{-2}\sum_{t\le n}t^2(\widehat q_t-\widehat q_n)^2$, and the 95\% critical value of \citet{lobato2001}. CBB$_4$ is the circular block percentile bootstrap of Appendix~\ref{app:experiments} with blocks of length $4\lceil n^{1/3}\rceil$. RS-SGD is the random-scaling interval of \citet{lee2022random} for averaged quantile SGD with $\eta_t=3t^{-.505}$, $q_0=0$, 300 discarded warm-up steps and critical value 6.747. Its configuration was chosen on 500 separate pilot streams per model among six candidates, before this study. Seeds are $2026100431+100003j+i$ for cell $j$ and stream $i$; CBB resampling uses an independent generator. All intervals were finite.

\subsubsection{Results}
Table~\ref{tab:v3full} reports all methods, Table~\ref{tab:v3paired} paired differences from E-hull with the power window, and Table~\ref{tab:v3auto} the selected windows together with the deterministic cap. The data-selected window raises coverage by $.81$--$1.75$ points in the three cells where the power window omits covariance (paired MCSEs $.22$--$.33$), at $3$--$6\%$ longer median length. In the MA(11) cell at $n=3{,}000$, $\hat b$ reaches its cap of 15 in every stream, whereas $\hat m$ has median 10: the correlogram rule slightly underestimates the memory 11, and doubling covers it. On the same streams, E-hull with the deterministic cap $\lceil n^{1/3}\rceil$ has the coverage of $\hat b$ in the four cells with strong dependence, whereas for i.i.d.\ and MA(1) data $\hat b$ stays at $0$ and $2$ and gives median lengths $1.3\%$ and $0.7\%$ shorter than the cap. Because $\widehat q_1=X_1$ enters $\mathcal D_n$, the SN length is at least $2(45.4)^{1/2}n^{-3/2}|X_1-\widehat q_n|$ and has infinite expectation under Cauchy margins; Tables~\ref{tab:v3full} and \ref{tab:v3paired} therefore summarize lengths by quantiles. RS-SGD undercovers in every cell, mostly through intervals lying below the target ($U<q^\star$).

\paragraph{Bias of the SGD-based pivot.}
The RS-SGD stepsizes $3t^{-.505}$ have $\gamma$ close to $1/2$, so the bias of the averaged iterate, of order $n^{-\gamma}$, is comparable to its standard deviation. Table~\ref{tab:rsbias} applies the finite-$n$ recursion of Appendix~\ref{app:finite} to this schedule, with $q_0=0$ and the 300 warm-up iterates excluded. The leading term predicts the observed bias to within $6\%$ and the recursion to within $11\%$ in every cell, including the infinite-memory AR(.8); the bias amounts to $0.47$--$0.82$ standard deviations. A Gaussian pivot with this bias would miss below the target in $6.8$--$13.4\%$ of streams and above it in at most $0.8\%$; the observed misses, $6.4$--$8.4\%$ below and $0.3$--$2.0\%$ above, show the same asymmetry, moderated by the heavier-tailed random-scaling critical value. The second-order bias thus determines the direction and much of the size of the coverage loss of this SGD-based pivot, whereas inversion at fixed candidates does not involve it.

{\renewcommand{\arraystretch}{.92}%
\begingroup\setlength{\tabcolsep}{3.5pt}
\begin{longtable}{llrrrrrr}
\caption{Interval comparison, 1,600 streams per cell. Coverage, its MCSE and directional misses in percent; ``Below'' means $U<q^\star$ and ``Above'' means $L>q^\star$. Lengths are summarized by their median and interquartile range.}\label{tab:v3full}\\
\toprule
Cell & Method & Cover & MCSE & Below & Above & Median length & IQR of length\\
\midrule\endfirsthead
\toprule Cell & Method & Cover & MCSE & Below & Above & Median length & IQR of length\\\midrule\endhead
i.i.d., 3k & E-hull, $\lceil n^{1/4}\rceil$ & 95.00 & 0.54 & 1.75 & 3.25 & 0.1962 & 0.1799--0.2129\\
i.i.d., 3k & E-hull, $\hat b$ & 95.38 & 0.53 & 1.56 & 3.06 & 0.1947 & 0.1795--0.2098\\
i.i.d., 3k & E-hull, $b=r-1$ & 95.38 & 0.53 & 1.56 & 3.06 & 0.1947 & 0.1795--0.2099\\
i.i.d., 3k & SN & 95.50 & 0.52 & 2.12 & 2.38 & 0.2402 & 0.1796--0.3172\\
i.i.d., 3k & CBB$_4$ & 93.75 & 0.61 & 2.69 & 3.56 & 0.1906 & 0.1710--0.2081\\
i.i.d., 3k & RS-SGD & 91.44 & 0.70 & 6.69 & 1.88 & 0.2318 & 0.1690--0.3163\\
MA(1), 3k & E-hull, $\lceil n^{1/4}\rceil$ & 95.06 & 0.54 & 1.94 & 3.00 & 0.2495 & 0.2296--0.2693\\
MA(1), 3k & E-hull, $\hat b$ & 95.12 & 0.54 & 1.94 & 2.94 & 0.2489 & 0.2301--0.2670\\
MA(1), 3k & E-hull, $b=r-1$ & 95.25 & 0.53 & 1.88 & 2.88 & 0.2485 & 0.2309--0.2663\\
MA(1), 3k & SN & 94.94 & 0.55 & 2.19 & 2.88 & 0.2999 & 0.2271--0.3930\\
MA(1), 3k & CBB$_4$ & 94.19 & 0.58 & 2.50 & 3.31 & 0.2401 & 0.2176--0.2648\\
MA(1), 3k & RS-SGD & 91.62 & 0.69 & 6.38 & 2.00 & 0.2923 & 0.2152--0.3887\\
MA(11), 3k & E-hull, $\lceil n^{1/4}\rceil$ & 94.25 & 0.58 & 1.81 & 3.94 & 0.5455 & 0.4959--0.6049\\
MA(11), 3k & E-hull, $\hat b$ & 95.06 & 0.54 & 1.50 & 3.44 & 0.5690 & 0.5141--0.6332\\
MA(11), 3k & E-hull, $b=r-1$ & 95.06 & 0.54 & 1.50 & 3.44 & 0.5681 & 0.5140--0.6282\\
MA(11), 3k & SN & 95.38 & 0.53 & 1.25 & 3.38 & 0.6800 & 0.5166--0.9305\\
MA(11), 3k & CBB$_4$ & 92.94 & 0.64 & 2.50 & 4.56 & 0.5310 & 0.4769--0.6012\\
MA(11), 3k & RS-SGD & 90.25 & 0.74 & 8.44 & 1.31 & 0.7135 & 0.5098--0.9821\\
MA(11), 12k & E-hull, $\lceil n^{1/4}\rceil$ & 96.12 & 0.48 & 1.56 & 2.31 & 0.2826 & 0.2680--0.2975\\
MA(11), 12k & E-hull, $\hat b$ & 96.06 & 0.49 & 1.62 & 2.31 & 0.2834 & 0.2670--0.2993\\
MA(11), 12k & E-hull, $b=r-1$ & 96.12 & 0.48 & 1.56 & 2.31 & 0.2826 & 0.2680--0.2975\\
MA(11), 12k & SN & 95.44 & 0.52 & 1.75 & 2.81 & 0.3431 & 0.2594--0.4587\\
MA(11), 12k & CBB$_4$ & 95.00 & 0.54 & 2.44 & 2.56 & 0.2722 & 0.2538--0.2918\\
MA(11), 12k & RS-SGD & 91.44 & 0.70 & 8.25 & 0.31 & 0.3628 & 0.2703--0.4885\\
AR(.8), 3k & E-hull, $\lceil n^{1/4}\rceil$ & 94.38 & 0.58 & 2.31 & 3.31 & 0.4474 & 0.4085--0.4905\\
AR(.8), 3k & E-hull, $\hat b$ & 96.12 & 0.48 & 1.56 & 2.31 & 0.4734 & 0.4291--0.5221\\
AR(.8), 3k & SN & 95.12 & 0.54 & 2.00 & 2.88 & 0.5586 & 0.4157--0.7535\\
AR(.8), 3k & CBB$_4$ & 93.56 & 0.61 & 2.81 & 3.62 & 0.4479 & 0.4014--0.4969\\
AR(.8), 3k & RS-SGD & 91.19 & 0.71 & 7.19 & 1.62 & 0.5709 & 0.4167--0.7909\\
AR(.8), 12k & E-hull, $\lceil n^{1/4}\rceil$ & 94.19 & 0.58 & 3.00 & 2.81 & 0.2304 & 0.2197--0.2426\\
AR(.8), 12k & E-hull, $\hat b$ & 95.00 & 0.54 & 2.44 & 2.56 & 0.2383 & 0.2254--0.2512\\
AR(.8), 12k & SN & 94.38 & 0.58 & 2.62 & 3.00 & 0.2887 & 0.2257--0.3836\\
AR(.8), 12k & CBB$_4$ & 93.19 & 0.63 & 3.25 & 3.56 & 0.2288 & 0.2134--0.2448\\
AR(.8), 12k & RS-SGD & 90.94 & 0.72 & 8.12 & 0.94 & 0.3014 & 0.2276--0.4033\\
\bottomrule\end{longtable}\endgroup
}

\begin{table}[H]
\centering\small
\caption{Paired comparison with E-hull for $b=\lceil n^{1/4}\rceil$ on common streams: difference in coverage (percentage points, MCSE in parentheses), median of the paired length differences, percentage of streams with a longer interval, and ratio of median lengths.}\label{tab:v3paired}
\begin{tabular}{llrrrr}
\toprule
Cell & Method & $\Delta$ coverage & Median $\Delta$ length & Longer (\%) & Median-length ratio\\
\midrule
i.i.d., 3k & E-hull, $\hat b$ & 0.38 (0.22) & -0.0008 & 41.4 & 0.993\\
i.i.d., 3k & SN & 0.50 (0.60) & 0.0432 & 68.3 & 1.224\\
i.i.d., 3k & CBB$_4$ & -1.25 (0.37) & -0.0060 & 34.8 & 0.971\\
i.i.d., 3k & RS-SGD & -3.56 (0.77) & 0.0364 & 65.0 & 1.181\\
MA(1), 3k & E-hull, $\hat b$ & 0.06 (0.19) & -0.0007 & 43.1 & 0.998\\
MA(1), 3k & SN & -0.12 (0.57) & 0.0485 & 66.8 & 1.202\\
MA(1), 3k & CBB$_4$ & -0.88 (0.36) & -0.0090 & 33.2 & 0.963\\
MA(1), 3k & RS-SGD & -3.44 (0.72) & 0.0409 & 63.7 & 1.172\\
MA(11), 3k & E-hull, $\hat b$ & 0.81 (0.22) & 0.0219 & 89.4 & 1.043\\
MA(11), 3k & SN & 1.12 (0.61) & 0.1272 & 70.1 & 1.247\\
MA(11), 3k & CBB$_4$ & -1.31 (0.36) & -0.0138 & 37.9 & 0.973\\
MA(11), 3k & RS-SGD & -4.00 (0.87) & 0.1544 & 70.2 & 1.308\\
MA(11), 12k & E-hull, $\hat b$ & -0.06 (0.06) & 0.0004 & 53.2 & 1.003\\
MA(11), 12k & SN & -0.69 (0.51) & 0.0578 & 67.5 & 1.214\\
MA(11), 12k & CBB$_4$ & -1.12 (0.32) & -0.0100 & 26.1 & 0.963\\
MA(11), 12k & RS-SGD & -4.69 (0.77) & 0.0786 & 71.9 & 1.284\\
AR(.8), 3k & E-hull, $\hat b$ & 1.75 (0.33) & 0.0247 & 97.0 & 1.058\\
AR(.8), 3k & SN & 0.75 (0.57) & 0.1025 & 70.1 & 1.249\\
AR(.8), 3k & CBB$_4$ & -0.81 (0.39) & 0.0001 & 50.2 & 1.001\\
AR(.8), 3k & RS-SGD & -3.19 (0.79) & 0.1273 & 71.4 & 1.276\\
AR(.8), 12k & E-hull, $\hat b$ & 0.81 (0.22) & 0.0070 & 91.9 & 1.034\\
AR(.8), 12k & SN & 0.19 (0.57) & 0.0574 & 73.1 & 1.253\\
AR(.8), 12k & CBB$_4$ & -1.00 (0.35) & -0.0021 & 44.1 & 0.993\\
AR(.8), 12k & RS-SGD & -3.25 (0.79) & 0.0733 & 74.8 & 1.308\\
\bottomrule
\end{tabular}
\end{table}

\begin{table}[H]
\centering\small
\caption{Data-selected windows: medians of $\hat m$ and $\hat b$, interquartile range of $\hat b$, the cap $\lceil n^{1/3}\rceil$ and the percentage of streams with $\hat b$ at the cap; coverage (\%) / median length of E-hull with $\hat b$ and with the deterministic cap on the same streams.}\label{tab:v3auto}
{\setlength{\tabcolsep}{4pt}
\begin{tabular}{lrrrrrcc}
\toprule
Cell & $\hat m$ & $\hat b$ & IQR of $\hat b$ & Cap & At cap (\%) & E-hull, $\hat b$ & E-hull, cap\\
\midrule
i.i.d., 3k & 0 & 0 & 0--0 & 15 & 0.0 & 95.38 / 0.1947 & 94.75 / 0.1972\\
MA(1), 3k & 1 & 2 & 2--2 & 15 & 0.0 & 95.12 / 0.2489 & 95.50 / 0.2506\\
MA(11), 3k & 10 & 15 & 15--15 & 15 & 100.0 & 95.06 / 0.5690 & 95.06 / 0.5690\\
MA(11), 12k & 11 & 22 & 20--22 & 23 & 10.6 & 96.06 / 0.2834 & 96.06 / 0.2835\\
AR(.8), 3k & 9 & 15 & 15--15 & 15 & 79.5 & 96.12 / 0.4734 & 96.12 / 0.4734\\
AR(.8), 12k & 12 & 23 & 20--23 & 23 & 52.3 & 95.00 / 0.2383 & 95.00 / 0.2385\\
\bottomrule
\end{tabular}}
\end{table}

\begin{table}[H]
\centering\small
\caption{Bias of the RS-SGD averaged iterate ($\times10^3$, MCSE in parentheses) against the finite-$n$ recursion and the leading term for its schedule $\eta_t=3t^{-.505}$, $q_0=0$, with 300 warm-up steps excluded; the bias in standard deviations of the averaged iterate; miss rates (\%) below and above the target predicted by a Gaussian pivot with the finite-$n$ bias, and observed.}\label{tab:rsbias}
\begin{tabular}{lrrrrcc}
\toprule
Cell & Bias & Finite-$n$ & Leading & Bias / SD & Gaussian below / above & Observed below / above\\
\midrule
i.i.d., 3k & -24.6 (1.3) & -25.0 & -23.7 & -0.47 & 6.8 / 0.8 & 6.7 / 1.9\\
MA(1), 3k & -32.3 (1.7) & -32.1 & -30.3 & -0.48 & 6.9 / 0.7 & 6.4 / 2.0\\
MA(11), 3k & -106.0 (4.3) & -117.1 & -109.2 & -0.62 & 10.1 / 0.4 & 8.4 / 1.3\\
MA(11), 12k & -62.4 (1.9) & -64.6 & -61.7 & -0.82 & 13.4 / 0.2 & 8.2 / 0.3\\
AR(.8), 3k & -77.4 (3.4) & -81.3 & -75.7 & -0.58 & 8.8 / 0.5 & 7.2 / 1.6\\
AR(.8), 12k & -44.0 (1.6) & -44.8 & -42.8 & -0.68 & 10.3 / 0.4 & 8.1 / 0.9\\
\bottomrule
\end{tabular}
\end{table}

\subsubsection{Computing time}
Table~\ref{tab:v3timing} reports end-to-end single-thread medians over nine runs after two warm-ups on one MA(1) Cauchy array per sample size, on Windows 11 with an Intel Family 6 Model 183 CPU, Python 3.12.14, NumPy 2.3.5, SciPy 1.17.0 and Numba 0.65.1, with all thread settings equal to one. Timing excludes data generation and compilation. CBB$_4$ is $3$--$10$ times slower than E-hull with the direct scan and $5$--$10$ times slower than with the Fenwick scan; SN and RS-SGD are faster, since each needs one pass.

\begin{table}[H]
\centering\small
\caption{Median single-thread time in milliseconds.}\label{tab:v3timing}
\begin{tabular}{lrrrr}
\toprule
Method & $n=3{,}000$ & $12{,}000$ & $10^5$ & $10^6$\\
\midrule
E-hull, power window, direct scan & 0.24 & 1.26 & 14.06 & 430.47\\
E-hull, power window, Fenwick scan & 0.23 & 1.27 & 12.99 & 248.82\\
E-hull, $\hat b$ (selection and scan) & 0.27 & 1.50 & 17.93 & 361.64\\
SN (two-heap recursion) & 0.09 & 0.51 & 5.09 & 58.40\\
CBB$_4$ (399 resamples) & 2.35 & 9.66 & 87.04 & 1317.46\\
RS-SGD & 0.08 & 0.35 & 3.04 & 33.44\\
\bottomrule
\end{tabular}
\end{table}

\subsection{Daily JPY/USD returns}\label{app:data}
We use the daily noon buying rate of the Japanese yen against the U.S. dollar \citep[series DEXJPUS of the Federal Reserve H.10 release, a public-domain series;][]{fred_dexjpus} from 1980-01-02 to 2026-09-25, drop missing quotes, and take percent log returns between consecutive quotes ($n=11{,}727$; 288 tied values, including 135 zero returns). The data enter in chronological order, and the example measures how dependence changes interval width. Table~\ref{tab:data} compares E-hull with $b=0$, which calibrates as if the returns were independent, with the power and data-selected windows, the fixed windows $\lceil n^{1/3}\rceil$ and $30$, CBB$_4$ and SN. At $\tau=.05$ the sum of the first 30 autocorrelations of the centered exceedance indicator is 0.72, reflecting volatility clustering. Accounting for dependence widens the interval at $\tau=.05$ by $31\%$ with $\hat b=6$ and $43\%$ with the power window, against $88\%$ for CBB$_4$. The wider windows $\lceil n^{1/3}\rceil=23$ and $b=30$ widen it by $62\%$ and $72\%$, comparable to the factor $(1+2\times0.72)^{1/2}=1.56$ implied by these autocorrelations at the sample quantile; at $\tau=.95$ they widen it by $34\%$ and $41\%$. Length ratios need not equal this factor, because the hull endpoints are determined by the candidate-dependent variance $\widehat V_b(q)$ at the two endpoints rather than at the sample quantile. The correlogram rule compares each autocorrelation with $2(\log_{10}n/n)^{1/2}=.037$ and stops at $\hat m=3$ for $\tau=.05$, whereas volatility clustering spreads small positive autocorrelations over many lags. At the median, indicators are nearly uncorrelated and the windows agree. SN intervals are $3.1$--$6.3$ times wider than E-hull with $b=0$ and at least $1.9$ times wider than every other interval; over 46 years the recursive quantiles drift with the volatility regime, which inflates the self-normalizer.

\begin{table}[H]
\centering\small
\caption{Intervals for quantiles of daily JPY/USD percent log returns, $n=11{,}727$. Length ratio is relative to E-hull with $b=0$.}\label{tab:data}
\begin{tabular}{rlrrrrr}
\toprule
$\tau$ & Method & $b$ & Lower & Upper & Length & Length ratio\\
\midrule
0.05 & E-hull, $b=0$ & 0 & -1.0804 & -1.0133 & 0.0672 & 1.00\\
0.05 & E-hull, $\lceil n^{1/4}\rceil$ & 11 & -1.0941 & -0.9978 & 0.0964 & 1.43\\
0.05 & E-hull, $\hat b$ & 6 & -1.0904 & -1.0023 & 0.0881 & 1.31\\
0.05 & E-hull, $\lceil n^{1/3}\rceil$ & 23 & -1.1040 & -0.9950 & 0.1089 & 1.62\\
0.05 & E-hull, $b=30$ & 30 & -1.1087 & -0.9932 & 0.1155 & 1.72\\
0.05 & CBB$_4$ &  & -1.1189 & -0.9922 & 0.1266 & 1.88\\
0.05 & SN &  & -1.2588 & -0.8335 & 0.4253 & 6.33\\
0.50 & E-hull, $b=0$ & 0 & 0.0000 & 0.0236 & 0.0236 & 1.00\\
0.50 & E-hull, $\lceil n^{1/4}\rceil$ & 11 & 0.0000 & 0.0240 & 0.0240 & 1.02\\
0.50 & E-hull, $\hat b$ & 0 & 0.0000 & 0.0236 & 0.0236 & 1.00\\
0.50 & E-hull, $\lceil n^{1/3}\rceil$ & 23 & 0.0000 & 0.0247 & 0.0247 & 1.04\\
0.50 & E-hull, $b=30$ & 30 & 0.0000 & 0.0246 & 0.0246 & 1.04\\
0.50 & CBB$_4$ &  & 0.0000 & 0.0236 & 0.0236 & 1.00\\
0.50 & SN &  & -0.0259 & 0.0496 & 0.0755 & 3.20\\
0.95 & E-hull, $b=0$ & 0 & 0.9574 & 1.0255 & 0.0681 & 1.00\\
0.95 & E-hull, $\lceil n^{1/4}\rceil$ & 11 & 0.9505 & 1.0310 & 0.0805 & 1.18\\
0.95 & E-hull, $\hat b$ & 23 & 0.9433 & 1.0348 & 0.0915 & 1.34\\
0.95 & E-hull, $\lceil n^{1/3}\rceil$ & 23 & 0.9433 & 1.0348 & 0.0915 & 1.34\\
0.95 & E-hull, $b=30$ & 30 & 0.9422 & 1.0379 & 0.0957 & 1.41\\
0.95 & CBB$_4$ &  & 0.9414 & 1.0513 & 0.1100 & 1.62\\
0.95 & SN &  & 0.8859 & 1.0961 & 0.2102 & 3.09\\
\bottomrule
\end{tabular}
\end{table}

\section{Wald, online-bootstrap and simultaneous comparisons}\label{app:wald}
This appendix collects comparisons based on the connected-floor interval: density-based Wald intervals around averaged SGD, a smoothed online multiplier bootstrap, and Bonferroni intervals for several quantiles. In the connected-floor, known-window study, finite-filter inversion for positive-MA(1) Cauchy input at $n=3{,}000$ covers 95.13\% with mean length .2511. Frozen-score Wald covers 91.63\% with length .2828; equal-weight inversion covers 95.00\% with length .2484. Table~\ref{tab:fullknown} gives all models and sample sizes.

For independent Normal input, the CKW pointwise procedure covers 94.63--96.50\% across the two quantiles; positive MA(1) gives 86.00--87.38\%. The latter setting is outside its independent-input guarantee. Its inversion comparators use equal or finite-filter weights, the logarithmic window, and the connectedness floor with $\lambda=.5$. Table~\ref{tab:simultaneous} gives joint coverage for the connected-floor Bonferroni construction: finite-filter intervals cover 96.88\%, 95.00\%, and 95.63\% under positive MA(1), negative MA(1), and scale-mixture input.

\subsection{Wald and inversion procedures}
All intervals are nominally 95\%, with $z=\Phi^{-1}(.975)$. Define $\mathcal Q_L(v)=v^\top K_Lv$, where $(K_L)_{st}=(1-|s-t|/L)_+$. This is computed exactly by the moving-window identity $\mathcal Q_L(v)=L^{-1}\sum_j(\sum_{t=j-L+1}^{j}v_t)^2$, taking observations outside $1{:}n$ to be zero. Let
$\ell_n^{\rm Wald}=\lceil 8n^{3/16}\rceil$ throughout the three Wald comparisons. The taper, density bandwidths and learning rates are fixed independently of coverage results.

\paragraph{Frozen-score finite-filter Wald.}
Split the pilot at $m=\lfloor M/2\rfloor$. On its first part, run \eqref{eq:sgd} with $(c,\gamma,t_0)=(3,.75,20)$, $q_0=0$, and projection to $[-6,6]$, then average the last two thirds to get $q_P$. Estimate the density from the second pilot part using a box kernel of width $(M-m)^{-1/4}$, clipping the estimate to $[.025,2.5]$. Denote it by $h_P$. On the $n$ analysis observations, run the same projected SGD initialized at $q_P$, and take the full average $\bar q_n$. With $u_t=\ind(X_t\le q_P)-n^{-1}\sum_s\ind(X_s\le q_P)$, the reported interval is
\[
\bar q_n\ \pm\ z\sqrt{\mathcal Q_{\ell_n^{\rm Wald}}(A_n(h_P)\odot u)/n}.
\]
A numerical variance lower bound $10^{-12}$ is used before taking the square root.

\paragraph{Second-pass finite-filter Wald.}
The SGD point estimate and initialization are unchanged. Estimate $h$ by the box kernel at $\bar q_n$ using the analysis sample and width $n^{-1/4}$, clipped to $[.02,4]$. Recompute and empirically center the threshold indicators at $\bar q_n$. Use the same finite-filter Bartlett expression. The second pass updates calibration around the same SGD point estimate.

\paragraph{One-step Wald.}
Starting from the second-pass point estimate and density $\widehat h$, form
$q_1=\bar q_n-\{n^{-1}\sum_t\ind(X_t\le\bar q_n)-\tau\}/\widehat h$.
Reestimate the density at $q_1$ with the same analysis-sample rule. Let $u_t=\ind(X_t\le q_1)-n^{-1}\sum_s\ind(X_s\le q_1)$. The standard error is
$\sqrt{\mathcal Q_{\ell_n^{\rm Wald}}(u/\sqrt n)/n}/\widehat h_1$.
This score correction updates both the center and the density estimate.

\paragraph{Score inversion.}
Filtered inversion uses \eqref{eq:filter} with $u_0=1$ and the same $(c,\gamma,t_0)$ as above, normalized to unit squared norm. Equal-weight inversion uses $w_t=n^{-1/2}$. The $b=0$ control uses the original finite-filter weights and omits all lag products. Its comparison therefore assesses the effect of lag correction for that weighting. Covariances are centered at the hypothesized $\tau$, as in \eqref{eq:candvar}; the floor is specified in Table~\ref{tab:configindex}.

\subsection{Design and seeds of the connected-floor studies}
An independent 400-replication positive-MA(1) Cauchy pilot was used to develop the comparison protocol and is excluded from reported performance. The known-window study fixes $b=r-1$, $\lambda=1$, and the method parameters. At each $n\in\{3{,}000,12{,}000\}$, the eight settings are independent/$\tau=.25$, positive MA(1)/$\tau=.25,.5,.75$, negative MA(1)/$\tau=.25$, memory four/$\tau=.25$, and scale mixture/$\tau=.25,.75$. Each cell contains 1,600 independent streams. This gives 25,600 latent streams, 44,800 marginal-model evaluations, and 268,800 endpoint pairs across six methods.

The log-window study uses independent seeds, $b=\lceil\sqrt{\log n}\rceil$, and $\lambda=1/2$. It includes independent, positive MA(1), negative MA(1), memory-four, and scale-mixture data at $\tau=.25$ and both sample sizes. Its 16,000 latent streams give 28,800 marginal-model evaluations and 172,800 endpoint pairs. The windows are $b=3$ and $4$, sufficient for these models. All method settings were fixed before this independently seeded study; the protocol was informed by the known-window results.

For a coverage or tail indicator $Y$, the reported Monte Carlo standard error is $\operatorname{sd}(Y_1,\ldots,Y_R)/\sqrt R$. Mean lengths and biases use the same formula with their respective observations. The figures use Wilson 95\% Monte Carlo intervals. Comparisons within a replication are paired; the full package includes paired coverage differences. The directional misses are defined as follows: ``below'' means the upper endpoint is less than the true quantile, and ``above'' means the lower endpoint exceeds it. All empty, unbounded and disconnected outputs are included in the records. None occurred in these two connected-floor studies.

The first-stage seeds are $92127000+100003\,j$, where $j=0,\ldots,15$ is the cell index in the order specified above. The second-stage seed adds $10{,}000{,}000$ and uses $j\in\{0,1,4,5,6,8,9,12,13,14\}$. The two independent studies are reported separately.

\subsection{Coverage tables}\label{app:tables}
The abbreviations are: Filter, finite-filter score inversion; Equal, equal-weight inversion; Frozen, frozen-score finite-filter Wald; Two-pass, second-pass finite-filter Wald; One-step, one-step Wald; and $b=0$, finite-filter inversion omitting all lag corrections. Each row summarizes 1,600 paired method evaluations. Monte Carlo standard errors, both tails and every individual endpoint are retained in the accompanying CSV files. These tables use connected-floor intervals. The normalized floor is $(\lambda z(4b+1)w_*)^2$. Known-window and independently seeded log-window results are kept separate.

{\footnotesize
\begin{longtable}{lrrrrrrrrrrr}
\caption{Connected-floor, known-window validation: coverage percentages, $b=r-1$, $\lambda=1$. Inversion coverage events are identical for Gaussian and Cauchy margins by transformation equivariance and are listed once; Wald methods are listed for both margins. Scale-mixture observations are untransformed and appear under the Gaussian column.}\label{tab:fullknown}\\
\toprule
Process & $n$ & $\tau$ & Filter & Equal & $b=0$ & \multicolumn{2}{c}{Frozen} & \multicolumn{2}{c}{Two-pass} & \multicolumn{2}{c}{One-step}\\
 & & & & & & Gauss. & Cauchy & Gauss. & Cauchy & Gauss. & Cauchy\\
\midrule
\endfirsthead
\toprule
Process & $n$ & $\tau$ & Filter & Equal & $b=0$ & \multicolumn{2}{c}{Frozen} & \multicolumn{2}{c}{Two-pass} & \multicolumn{2}{c}{One-step}\\
 & & & & & & Gauss. & Cauchy & Gauss. & Cauchy & Gauss. & Cauchy\\
\midrule
\endhead
Scale mixture & 3k & 0.25 & 95.25 & 95.19 & 96.63 & 94.63 & -- & 94.63 & -- & 94.38 & --\\
Scale mixture & 3k & 0.75 & 95.50 & 95.00 & 94.38 & 94.88 & -- & 94.88 & -- & 95.06 & --\\
Scale mixture & 12k & 0.25 & 95.31 & 95.13 & 96.50 & 94.63 & -- & 95.13 & -- & 94.94 & --\\
Scale mixture & 12k & 0.75 & 95.69 & 95.56 & 93.88 & 95.63 & -- & 95.19 & -- & 95.56 & --\\
Independent & 3k & 0.25 & 94.38 & 94.38 & 94.38 & 93.94 & 92.06 & 94.06 & 94.13 & 94.44 & 93.38\\
Independent & 12k & 0.25 & 95.56 & 95.56 & 95.56 & 95.25 & 95.00 & 95.25 & 95.13 & 96.00 & 95.63\\
Memory four & 3k & 0.25 & 94.50 & 94.31 & 74.25 & 92.31 & 90.19 & 93.25 & 93.44 & 93.38 & 92.69\\
Memory four & 12k & 0.25 & 95.19 & 95.38 & 76.19 & 94.63 & 93.06 & 94.75 & 95.00 & 94.56 & 94.19\\
Negative MA(1) & 3k & 0.25 & 95.19 & 94.94 & 98.75 & 94.50 & 91.44 & 94.75 & 93.94 & 95.38 & 94.63\\
Negative MA(1) & 12k & 0.25 & 94.69 & 95.00 & 98.44 & 94.94 & 93.75 & 94.44 & 94.75 & 94.44 & 94.19\\
Positive MA(1) & 3k & 0.25 & 95.13 & 95.00 & 87.56 & 94.94 & 91.63 & 94.56 & 94.56 & 94.63 & 94.25\\
Positive MA(1) & 3k & 0.5 & 95.13 & 95.63 & 88.31 & 94.75 & 94.44 & 94.38 & 94.19 & 94.25 & 94.56\\
Positive MA(1) & 3k & 0.75 & 95.06 & 95.00 & 87.81 & 94.31 & 92.13 & 94.50 & 94.31 & 94.81 & 94.50\\
Positive MA(1) & 12k & 0.25 & 95.75 & 95.63 & 88.38 & 94.69 & 94.06 & 95.00 & 95.25 & 95.50 & 95.63\\
Positive MA(1) & 12k & 0.5 & 94.19 & 94.25 & 85.44 & 94.56 & 94.50 & 94.25 & 94.25 & 93.94 & 94.06\\
Positive MA(1) & 12k & 0.75 & 94.69 & 94.88 & 87.75 & 93.81 & 94.00 & 94.56 & 94.63 & 94.38 & 94.19\\
\bottomrule
\end{longtable}
}

{\footnotesize
\begin{longtable}{lrrrrrrrrrrr}
\caption{Connected-floor, log-window validation: coverage percentages, $b=\lceil\sqrt{\log n}\rceil$, $\lambda=1/2$, arranged as in Table~\ref{tab:fullknown}.}\\
\toprule
Process & $n$ & $\tau$ & Filter & Equal & $b=0$ & \multicolumn{2}{c}{Frozen} & \multicolumn{2}{c}{Two-pass} & \multicolumn{2}{c}{One-step}\\
 & & & & & & Gauss. & Cauchy & Gauss. & Cauchy & Gauss. & Cauchy\\
\midrule
\endfirsthead
\toprule
Process & $n$ & $\tau$ & Filter & Equal & $b=0$ & \multicolumn{2}{c}{Frozen} & \multicolumn{2}{c}{Two-pass} & \multicolumn{2}{c}{One-step}\\
 & & & & & & Gauss. & Cauchy & Gauss. & Cauchy & Gauss. & Cauchy\\
\midrule
\endhead
Scale mixture & 3k & 0.25 & 94.44 & 94.38 & 95.94 & 94.13 & -- & 94.69 & -- & 95.06 & --\\
Scale mixture & 12k & 0.25 & 93.88 & 93.88 & 95.63 & 93.31 & -- & 93.31 & -- & 93.50 & --\\
Independent & 3k & 0.25 & 94.75 & 94.88 & 94.88 & 94.13 & 90.69 & 94.38 & 93.88 & 94.31 & 93.50\\
Independent & 12k & 0.25 & 95.25 & 95.00 & 95.31 & 94.81 & 94.63 & 95.06 & 95.00 & 94.88 & 94.44\\
Memory four & 3k & 0.25 & 94.63 & 94.56 & 74.31 & 93.38 & 90.56 & 94.13 & 93.19 & 94.38 & 94.06\\
Memory four & 12k & 0.25 & 95.44 & 95.38 & 75.63 & 95.00 & 93.75 & 95.25 & 94.56 & 94.69 & 94.69\\
Negative MA(1) & 3k & 0.25 & 95.38 & 95.38 & 98.94 & 95.44 & 92.63 & 95.00 & 95.88 & 95.63 & 95.38\\
Negative MA(1) & 12k & 0.25 & 95.13 & 94.94 & 98.88 & 94.81 & 94.13 & 95.06 & 94.94 & 94.38 & 94.81\\
Positive MA(1) & 3k & 0.25 & 94.81 & 95.06 & 87.44 & 93.81 & 91.50 & 94.38 & 93.88 & 93.63 & 93.56\\
Positive MA(1) & 12k & 0.25 & 95.31 & 95.31 & 88.38 & 94.81 & 94.19 & 95.06 & 94.81 & 95.00 & 94.88\\
\bottomrule
\end{longtable}
}

Mean lengths for every cell are in the accompanying CSV files. Relative to equal-weight inversion, frozen-score Wald intervals are 3--21\% longer, second-pass Wald intervals 1--8\% longer, and one-step Wald intervals range from 3.3\% shorter to 0.5\% longer; finite-filter inversion is 0.6--1.3\% longer than equal-weight inversion. In the known-window study, the $b=0$ control is 21--42\% shorter than finite-filter inversion under positive indicator autocorrelation (positive MA(1) and memory four), 28\% longer under negative MA(1), identical for independent data, and 6\% shorter to 9\% longer for the scale mixture, matching the direction of its coverage errors.

\subsection{Smoothed online multiplier bootstrap}
The Chen--Keilbar--Wu (CKW) procedure uses the smoothed original and multiplier recursions (1) and (14) of \citet{chen2026smooth}. The step size is $2t^{-2/3}$, the smoothing half-width is four times the step size, and each bootstrap multiplier is $2\operatorname{Bernoulli}(1/2)$. The original and 199 perturbed trajectories start at zero. The pointwise interval radius is the empirical 95th percentile of $|\bar q_n^*-\bar q_n|$; the order-statistic index is $\lceil .95(R_{\rm b}+1)\rceil$, capped at $R_{\rm b}=199$.
We use the pointwise restriction of the source procedure, with 199 bootstrap trajectories; the source's simultaneous-band experiments use 1,000. Each method receives $n=3{,}000$ observations. We evaluate 800 independent streams for each combination of independent or positive-MA(1) Normal input and $\tau=.25,.5$. MA(1) supplies a stress test outside the bootstrap's independent-input guarantee. Table~\ref{tab:ckw} reports coverage and length. The bootstrap maintains 199 extra trajectories, while inversion stores the observations.

\begin{table}[ht]
\centering\small
\caption{Smoothed online multiplier bootstrap (CKW) versus score inversion. Coverage and tail probabilities are in percent. Every entry uses 800 replications and $n=3{,}000$; the pointwise CKW restriction uses 199 bootstrap trajectories. Parentheses give Monte Carlo standard errors.}\label{tab:ckw}
\begin{tabular}{llrrrrr}
\toprule
Process & Method & $\tau$ & Coverage (MCSE) & Below & Above & Length \\
\midrule
Independent & CKW pointwise & 0.25 & 96.50 (0.65) & 3.38 & 0.13 & 0.1025 \\
Independent & Equal inversion & 0.25 & 95.75 (0.71) & 1.50 & 2.75 & 0.0980 \\
Independent & Filtered inversion & 0.25 & 95.50 (0.73) & 2.00 & 2.50 & 0.0992 \\
Independent & CKW pointwise & 0.5 & 94.63 (0.80) & 2.00 & 3.38 & 0.0929 \\
Independent & Equal inversion & 0.5 & 94.50 (0.81) & 2.63 & 2.88 & 0.0901 \\
Independent & Filtered inversion & 0.5 & 94.38 (0.82) & 2.50 & 3.13 & 0.0911 \\
Positive MA(1) & CKW pointwise & 0.25 & 87.38 (1.17) & 9.00 & 3.62 & 0.1021 \\
Positive MA(1) & Equal inversion & 0.25 & 95.00 (0.77) & 2.25 & 2.75 & 0.1241 \\
Positive MA(1) & Filtered inversion & 0.25 & 95.25 (0.75) & 2.25 & 2.50 & 0.1259 \\
Positive MA(1) & CKW pointwise & 0.5 & 86.00 (1.23) & 6.50 & 7.50 & 0.0925 \\
Positive MA(1) & Equal inversion & 0.5 & 93.63 (0.86) & 3.00 & 3.38 & 0.1156 \\
Positive MA(1) & Filtered inversion & 0.5 & 94.25 (0.82) & 2.75 & 3.00 & 0.1169 \\
\bottomrule
\end{tabular}
\end{table}

\subsection{Simultaneous intervals on a fixed quantile grid}
The simultaneous-interval study evaluates $\tau\in\{.25,.5,.75\}$ on common samples, with $n=3{,}000$ and 1,600 independent streams per process. It uses the log window, the connectedness floor with $\lambda=1/2$, and no pilot. Positive and negative MA(1) have Cauchy margins; the scale-mixture observations are unchanged. Finite-filter and equal-weight intervals use per-level error $.05/3$ followed by monotone tightening. Unadjusted finite-filter intervals use per-level error $.05$. The theoretical target is a lower bound on simultaneous coverage; Bonferroni intervals can be conservative.
\begin{table}[ht]
\centering\small
\caption{Connected-floor, log-window three-quantile validation. Joint coverage is in percent; Monte Carlo standard errors are in parentheses. Length is averaged over the three tightened intervals.}\label{tab:simultaneous}
\begin{tabular}{llrrr}
\toprule
Process & Method & Joint coverage (MCSE) & Mean length & Empty rate \\
\midrule
Scale mixture & Equal-weight Bonferroni & 95.38 (0.53) & 0.0851 & 0.00 \\
Scale mixture & Finite-filter Bonferroni & 95.63 (0.51) & 0.0861 & 0.00 \\
Scale mixture & Unadjusted finite-filter & 86.25 (0.86) & 0.0704 & 0.00 \\
Negative MA(1) & Equal-weight Bonferroni & 95.13 (0.54) & 0.1564 & 0.00 \\
Negative MA(1) & Finite-filter Bonferroni & 95.00 (0.55) & 0.1584 & 0.00 \\
Negative MA(1) & Unadjusted finite-filter & 86.06 (0.87) & 0.1292 & 0.00 \\
Positive MA(1) & Equal-weight Bonferroni & 96.75 (0.44) & 0.2627 & 0.00 \\
Positive MA(1) & Finite-filter Bonferroni & 96.88 (0.44) & 0.2658 & 0.00 \\
Positive MA(1) & Unadjusted finite-filter & 90.38 (0.74) & 0.2169 & 0.00 \\
\bottomrule
\end{tabular}
\end{table}

\section{Detailed comparison with related work}\label{app:related}
This appendix places the paper in a broader context and, in Table~\ref{tab:closestwork}, lists for the closest work the statistical object, the assumptions and the relation to our results.

\paragraph{Stochastic approximation and delays.}
The ODE method and Poisson-equation techniques give convergence and central limit theorems for SA with Markovian or state-dependent noise \citep{benveniste1990,kushner2003,borkar2008,fort2015}, and Polyak--Ruppert averaging attains the efficient covariance \citep{ruppert1988,polyak1992}. Non-asymptotic bounds for linear and nonlinear SA include \citet{bach2011,srikant2019,mou2020,durmus2021}, and \citet{duchi2012ergodic} quantify how mixing slows stochastic optimization on ergodic data. Constant-step bias expansions and Richardson--Romberg extrapolation are developed for independent noise by \citet{dieuleveut2020,sheshukova2025rr} and for Markov noise by \citet{huo2023linear,allmeier2024,huo2024nonlinear,levin2026}. Delayed gradients arising from asynchrony cost little under stochastic noise \citep{agarwal2011delay,arjevani2020delay,mishchenko2022}; likewise, the deliberate delays of Theorem~\ref{thm:delay} leave the leading second moment unchanged.

\paragraph{Recursive quantiles, conformal calibration and sketches.}
Recursive estimation also covers the geometric median in Hilbert spaces \citep{cardot2017}. Beyond adaptive conformal inference, PID-type rules adapt the calibration level to trends \citep{angelopoulos2023pid}, and time-series conformal methods rely on ensembles under mixing or on reweighting \citep{xu2021enbpi,barber2023}; these methods target the coverage of prediction sets rather than inference for a population quantile. For approximate rank queries, \citet{karnin2016} give a streaming sketch with optimal space.

\paragraph{Online inference for SGD.}
Hierarchical incremental gradient descent provides another online uncertainty quantification \citep{su2023higrad}, divide-and-conquer estimators treat quantile regression under memory constraints \citep{chen2019qrmemory}, Gaussian approximation and multiplier-bootstrap validity hold for independently sampled SGD \citep{sheshukova2026}, and statistical inference extends to Markovian linear SA \citep{samsonov2025markov}. These methods calibrate the covariance of the averaged iterate; Appendix~\ref{app:v3intervals} shows that at stepsizes close to $t^{-1/2}$ its mean term is comparable to the standard error.

\paragraph{Quantile inference.}
Rank-test inversion extends interval construction to quantile regression \citep{koenker2005}, and Wilks-type regions cover empirical weighted quantiles \citep{allouche2026}. Resampling alternatives for dependent data include the stationary bootstrap \citep{politis1994stationary}, subsampling \citep{politis1999subsampling}, automatic block-length selection \citep{politis2004block} and a smooth block bootstrap for quantile regression \citep{gregory2018}. Online covariance estimation and time-series bootstrap calibrate dependence in streaming problems \citep{leung2026,palm2024}. Normal approximation through dependency graphs goes back to \citet{baldi1989}, and \citet{delapena2009} survey self-normalized limit theory.
\begin{table}[htbp]
\centering\small
\caption{Scope of related work and the corresponding result here. Entries identify the statistical object and assumptions relevant to comparison.}\label{tab:closestwork}
\begin{tabular}{p{.22\textwidth}p{.31\textwidth}p{.39\textwidth}}
\toprule
Work & Object and setting & Relation to this paper\\
\midrule
\citet{huo2023linear,allmeier2024,huo2024nonlinear} & Constant-step linear or smooth nonlinear SA with Markov noise; stationary bias and extrapolation & The raw indicator is discontinuous. Our decreasing-step boundary coefficient separates two directions with identical score covariance; a delayed threshold removes it.\\
\citet{lauand2025,levin2026,sheshukova2025rr} & Decreasing-step linear-SA bias and covariance; constant-step shared-stream RR for linear or smooth SGD & We derive the polynomial-schedule cross-gain coefficient and transfer it to raw dependent quantile SGD for $\gamma>4/5$.\\
\citet{sheshukova2026} & Gaussian approximation and multiplier bootstrap for independently sampled SGD & We separate finite covariance from an adaptive boundary mean effect, and calibrate a different pivot for the same target.\\
\citet{chen2026smooth} & Smoothed quantile recursion with independent inputs; single-pass multiplier and simultaneous inference & Our bias theory treats the raw indicator under finite memory. Exact target inversion stores the chronological observations and avoids density estimation.\\
\citet{leung2026,palm2024} & Online dependence calibration through covariance estimation or time-series bootstrap & Their streaming objectives complement the stored-data exact scan. Our coverage result tracks deterministic weights, lag truncation and acceptance-hull geometry.\\
\citet{lahiri2009quantile,dey2025quantile} & Quantitative normal approximation for sample quantiles under weak or local dependence; the latter also treats joint quantiles & Our bias result concerns a recursive adaptive threshold. Our interval result controls candidatewise studentization and acceptance-hull coverage.\\
\citet{zhang2024self,cai2026refined} & Normal approximation for self-normalized sums under local dependence & Their approximation order is transferred to the implemented interval through the denominator identity, local hull exactness and the exact scan.\\
\citet{woodruff1952,sitter2001,gregory2018} & Distribution-function inversion and dependent quantile-regression resampling & Inversion and resampling are established principles. We add quantitative hull control, the transfer of the finite-window rate to the implemented interval, data-selected windows and an $O(n\log n)$ exact scan.\\
\citet{shao2010self,kitamura1997,chen2009sbel,sun2006quantile} & Self-normalization, blockwise empirical likelihood and block bootstrap for dependent quantiles & Our pivot keeps Gaussian critical values and an explicit window, is invariant under increasing transformations, and has coverage guarantees for the implemented hull; Table~\ref{tab:block} compares it with SN and CBB.\\
\citet{gibbs2021aci,angelopoulos2024decaying,rowland2024qtd} & Online conformal calibration and quantile TD learning built on quantile-SGD updates & Their inputs are serially dependent; for the decreasing-step, fixed-target recursion of the decaying-stepsize variant, Theorem~\ref{thm:bias} identifies the resulting mean term.\\
\citet{nagaraj2020markov,kowshik2021rer} & SGD on Markovian data; data dropping, experience replay and reverse replay & We identify the boundary functional through which processing direction changes quantile-SGD bias, and delay thresholds without discarding data.\\
\bottomrule
\end{tabular}
\end{table}

\paragraph{Self-normalization.}
\citet{shao2010self} constructs a pivotal normalization from recursive estimates, leading to non-Gaussian critical values under a functional limit \citep{kiefer2000,lobato2001,shao2015review}. Our candidatewise lag estimate retains Gaussian critical values and an explicit window. In Table~\ref{tab:block}, the two have comparable coverage and SN intervals are 20--25\% longer. Blockwise empirical likelihood also avoids variance estimation \citep{kitamura1997,chen2009sbel}, at the cost of a block length and, for quantiles, a smoothing bandwidth.

\paragraph{Dependent sample quantiles.}
\citet{lahiri2009quantile} establish an $O(n^{-1/2})$ Berry--Esseen rate for the sample quantile under their conditions (C.1)--(C.5). These include density and variance nondegeneracy, polynomial mixing and additional approximation conditions. \citet[Theorems 1.2 and 1.3]{dey2025quantile} give quantitative univariate and joint-quantile approximation under local-dependence neighborhoods. Converting a quantile event into an indicator sum is already central to that analysis. Our fixed-window theorem calibrates a candidate-dependent empirical denominator and controls the closed hull of the accepted candidates.

\citet{wei2026constant} study constant-step quantile SGD with independent inputs and small-step limits of its stationary law. Theorem~\ref{thm:bias} uses decreasing stepsizes and finite-memory inputs, with a lag-resolved mean correction at an adaptive indicator boundary.
\end{document}